\documentclass{article} % For LaTeX2e
\usepackage{iclr2026_conference,times}

\usepackage{amsmath,amsfonts,bm}

\def\eqref#1{equation~\ref{#1}}
\def\1{\bm{1}}

\DeclareMathAlphabet{\mathsfit}{\encodingdefault}{\sfdefault}{m}{sl}
\SetMathAlphabet{\mathsfit}{bold}{\encodingdefault}{\sfdefault}{bx}{n}

\usepackage{url}
\usepackage{bm}
\usepackage{algorithm}
\usepackage{algorithmicx}
\usepackage{algpseudocode}
\usepackage{comment}
\usepackage{graphicx}
\usepackage{enumitem}
\usepackage{amssymb}
\usepackage{multirow}
\usepackage{siunitx}
\usepackage{wrapfig}
\usepackage{graphicx} 
\usepackage{appendix}
\usepackage{adjustbox}
\usepackage{float}
\usepackage{lscape}
\usepackage{titletoc}
\usepackage[nottoc]{tocbibind}
\usepackage{minitoc}
\usepackage{subcaption}
\usepackage{booktabs}
\usepackage[table,xcdraw]{xcolor}
\usepackage[colorlinks = true,
            linkcolor = blue,
            urlcolor  = blue,
            citecolor = blue,
            anchorcolor = blue]{hyperref}
\hypersetup{citecolor=[rgb]{0,0.2,0.75}}
\hypersetup{linkcolor=[rgb]{0,0.2,0.75}}
\hypersetup{urlcolor=[rgb]{0,0.2,0.75}}
\usepackage[capitalize]{cleveref}
\usepackage[symbol]{footmisc}

\DeclareMathOperator{\cS}{\mathcal{S}}
\DeclareMathOperator{\cA}{\mathcal{A}}

\DeclareMathOperator{\cP}{\mathcal{P}}

\DeclareMathOperator{\cZ}{\mathcal{Z}}

\DeclareMathOperator{\HV}{\text{HV}}
\newcommand{\blambda}{\bm{\lambda}}
\newcommand{\bz}{\bm{z}}

\newcommand{\bbB}{\mathbf{B}}
\newcommand{\bbF}{\mathbf{F}}

\newcommand{\br}{\boldsymbol{r}}
\newcommand{\bmR}{\bm{R}}
\newcommand{\ie}{\textit{i.e., }}
\newcommand{\eg}{\textit{e.g., }}

\newcommand{\pchreb}[1]{\textcolor{black}{#1}}
\newcommand{\wh}[1]{\textcolor{black}{#1}}

\newcommand{\bswreb}[1]{\textcolor{black}{#1}}
\newcommand{\rev}[1]{\textcolor{black}{#1}}
\newcommand{\iclrwh}[1]{\textcolor{black}{#1}}

\title{A Reward-Free Viewpoint on Multi-Objective Reinforcement Learning}
\author{%
  \hspace{-3pt}\textbf{Ying-Tu Chen\textsuperscript{\normalfont{1},}\footnotemark[2]}\ \ \ \textbf{ Wei Hung\textsuperscript{\normalfont{1},}\footnotemark[2]} \ \  \textbf{ Bing-Shu Wu\textsuperscript{\normalfont{1},}\footnotemark[2]} \ \ \textbf{ Zhang-Wei Hong\textsuperscript{\normalfont{2}}} \ \ \textbf{ Ping-Chun Hsieh\textsuperscript{\normalfont{1}}}\\
  \hspace{-3pt}\textsuperscript{1}National Yang Ming Chiao Tung University, Hsinchu, Taiwan\\
  \hspace{-3pt}\textsuperscript{2}Massachusetts Institute of Technology\\
  \hspace{-3pt}\texttt{\{shu0924.cs13,pinghsieh\}@nycu.edu.tw},\quad \texttt{zwhong@mit.edu} 
}
\iclrfinalcopy % Uncomment for camera-ready version, but NOT for submission.
\begin{document}

\maketitle

\begin{abstract}
Many sequential decision-making tasks involve optimizing multiple conflicting objectives, requiring policies that adapt to different user preferences. \iclrwh{In multi-objective reinforcement learning (MORL), one widely studied approach} addresses this by training a single policy network conditioned on preference-weighted rewards. \pchreb{In this paper, we explore a novel algorithmic perspective: leveraging reward-free reinforcement learning (RFRL) for MORL.} While RFRL has historically been studied independently of MORL, it learns optimal policies for any possible reward function, making it a natural fit for MORL's challenge of handling unknown user preferences. We propose using the RFRL's training objective as an auxiliary task to enhance MORL, enabling more effective knowledge sharing beyond the multi-objective reward function given at training time. To this end, we adapt a state-of-the-art RFRL algorithm to the MORL setting and introduce a preference-guided exploration strategy that focuses learning on relevant parts of the environment. \pchreb{Through extensive experiments and ablation studies, we demonstrate that our approach significantly outperforms the state-of-the-art MORL methods across diverse MO-Gymnasium tasks, achieving superior performance and data efficiency. This work provides the first systematic adaptation of RFRL to MORL, demonstrating its potential as a scalable and empirically effective solution to multi-objective policy learning.\footnotemark[1]}\footnotetext[2]{Equal contribution.}\footnotetext[1]{\texttt{https://rl-bandits-lab.github.io/MORL-FB/}}

%, especially in settings with limited preference samples. 
%Many sequential decision-making tasks involve optimizing multiple conflicting objectives, requiring policies that adapt to different user preferences. Multi-objective reinforcement learning (MORL) typically addresses this by training a single policy conditioned on preference-weighted rewards. In this paper, we explore a novel perspective: leveraging reward-free reinforcement learning (RFRL) for MORL. While RFRL has historically been studied independently of MORL, it learns optimal policies for any possible reward function, making it a natural fit for MORL's challenge of handling unknown user preferences. We propose using RFRL's training objective as an auxiliary task to enhance MORL, enabling more effective knowledge sharing beyond the multi-objective reward function given at training time. To this end, we adapt a state-of-the-art RFRL algorithm to the MORL setting and introduce a preference-guided exploration strategy that focuses learning on relevant part of the environment. Our approach significantly outperforms state-of-the-art MORL methods across diverse MO-Gymnasium tasks, achieving superior performance and data efficiency, especially in settings with limited preference samples. This work is the first to explicitly adapt RFRL for MORL, demonstrating its potential as a scalable and effective solution.
\end{abstract}

\section{Introduction}
\label{sec:intro}

Many sequential decision-making tasks require optimizing multiple, often conflicting objectives. For example, in robot control, there is a trade-off between minimizing energy consumption and maximizing speed. \iclrwh{One common approach to find a Pareto optimal policy is to maximize a weighted sum of the objectives}, where the weights represent \textit{user preferences}. User preferences depend on context—for instance, prioritizing speed in emergencies and energy efficiency in routine operations. Since user preferences are unknown in advance, solving multi-objective decision-making requires learning a set of policies for different preferences before testing.

Reinforcement learning (RL) \citep{sutton2018reinforcement} has achieved strong performance in sequential decision-making, making multi-objective RL (MORL) a widely studied approach for learning policies for different user preferences~\citep{Hayes_2022}. A naive but inefficient solution is to train a separate policy for each preference. Another more scalable approach is to train a single policy network \citep{yang2019general,basaklar2023pd,weihung2023qp} conditioned on preferences, enabling parameter sharing and generalization across preferences. During training, the policy is optimized over a range of sampled preferences, each defining a reward function weighted by the preference. At test time, users can specify a preference to obtain the corresponding policy.

Another approach to handling unknown user preferences at test time is reward-free reinforcement learning (RFRL) \citep{jin2020reward,ahmed2023zero}, which has historically been developed independently of MORL despite addressing a similar problem. In RFRL, the agent explores the environment without receiving reward signals during training and instead learns a set of optimal policies for any possible reward function in the environment. MORL can be seen as a special case of RFRL \citep{alegre2022optimistic}, as RFRL does not restrict the reward function to be a weighted sum of predefined reward functions. However, despite their similarities, no prior work has explicitly \iclrwh{adapted} RFRL methods to solve MORL problems.

In this paper, we ask: \textit{Can RFRL inform MORL?} We hypothesize that the objective of RFRL to learn optimal policies for any reward function could serve as a useful \textit{auxiliary task} for MORL \citep{jaderberg2016reinforcement,rafiee2022makes,veeriah2019discovery}. \pchreb{Although MORL under linear scalarization only needs optimal policies for linear combinations of known objectives,} learning beyond these combinations could accelerate MORL via effective knowledge sharing. To investigate this question, we adapt a state-of-the-art RFRL algorithm \citep{ahmed2023zero} to the MORL setting, treating the preference-weighted reward as the test-time reward function given to RFRL. However, this naive approach performs poorly compared to existing MORL methods, likely because purely reward-free exploration does not prioritize states that are important for optimizing the preference-weighted reward in the given MORL task. As a result, the policies learned by RFRL for these reward functions could be suboptimal. To address this, we propose \textit{guiding exploration using sampled preferences and mini-batch sampling}, directing the agent to visit states that maximize the corresponding preference-weighted reward function. This ensures that learning is focused on policies most relevant to MORL.

We highlight the main technical novelty of this paper: (1) \textbf{A new perspective of solving MORL}: We identify the close connection between MORL and RFRL, which have evolved independently despite tackling similar challenges of unknown user preferences at test time. This insight motivates new MORL algorithms by rethinking policy learning with multiple objectives from the perspective of RFRL. (2) \textbf{Algorithmic enhancements for adapting RFRL to MORL}: Even though RFRL and MORL are closely related, vanilla RFRL can perform poorly in the MORL setting (see Section \ref{sec:exp}). To address this, we introduce three key enhancements: (i) \textit{Preference-guided exploration}: We propose to use the preference vector to sample latent vectors aligned with the target rewards to facilitate exploration in the latent space; (ii) \textit{Training on latent vectors computed by mini-batch sampling from replay buffer as auxiliary tasks}: Our approach trains the policy network on latent vectors computed from mini-batch transitions sampled from the replay buffer. This design learns a broader range of policies than required for MORL and can be beneficial by providing auxiliary tasks; (iii) \textit{Auxiliary Q loss}: To better adapt RFRL to MORL, we further facilitate the learning of representations from the observed reward vectors (instead of pseudo rewards as in RFRL) via an auxiliary Q loss as an additional learning signal.

Our experimental results demonstrate that our approach is both simple and effective. First, our method significantly outperforms the state-of-the-art MORL algorithms \iclrwh{across various tasks in the MO-Gymnasium benchmark suite~\citep{felten2023toolkit}}, including discrete and continuous control. Second, when trained with a limited number of preference samples, our method achieves substantially higher performance than other MORL approaches. This highlights that decoupling environment knowledge from reward information enhances generalization, particularly in scenarios with limited preference samples. To the best of our knowledge, this is the first work to adapt RFRL for MORL and present a practical algorithm that performs well across diverse deep RL tasks.

\vspace{-0.6em}
\section{Preliminaries}
\label{sec:prelim}
\vspace{-0.6em}

{This section provides a brief review of MORL, along with the notation used throughout the paper. We use boldface symbols for vectors and matrices. For any $n\in\mathbb{N}$, we use $[n]$ as a shorthand for $\{1,\cdots,n\}$. For a set \(\cZ\), we let \(\Delta(\cZ)\) denote the set of all probability distributions over \(\cZ\).}

% Flow:
% 1. MOMDP
% 2. Pareto Front
% 3. To facilitate the discovery of PF, we use linear scalrization
% 4. Recently in (Lu et al., 2023), it has been shown that PF is convex.

{We formulate the MORL problem as an Multi-Objective Markov Decision Process (MOMDP) defined by the tuple \((\cS, \cA, \cP, \bmR, \gamma,\mu)\), where \(\cS\) and \(\cA\) are the state and action spaces, \(\cP:\cS\times \cA \rightarrow \Delta(\cS)\) is the transition function, \(\mathbf{R}:\cS\times \cA \rightarrow \mathbb{R}^d\) is a vector-valued reward function of \(d\) objectives, \(\gamma\in [0,1)\) is the discount factor, and $\mu$ is the initial state distribution.
%$f_{\bm{\lambda}}:\mathbb{R}^d\rightarrow \mathbb{R}$ is the scalarization utility function under a user preference vector $\bm{\lambda}\in\mathbb{R}^d$, and $\Lambda$ denotes the preference space.
Let $\Pi$ denote the set of all stationary randomized policies. Let $s_t$, $a_t$, $\boldsymbol{r}_t$ be the state, action, and reward received at time $t$.
For a policy $\pi\in\Pi$, define $\mathbf{V}^{\pi}:=\mathbb{E}_{\pi,s_0\sim\mu}[\sum_{t=0}^{\infty}\gamma^t \mathbf{R}(s_t,a_t)]$ as the expected total discounted return vector achieved by $\pi$. Let $\mathbf{V}^{\pi}_i$ denote the $i$-th entry of  $\mathbf{V}^{\pi}$.
For a pair of policies $\pi$ and $\pi'$, we say that \textit{$\pi$ Pareto-dominates $\pi'$} (denoted by $\pi \succ \pi'$) if $\mathbf{V}^{\pi}_i\geq \mathbf{V}^{\pi'}_i$ for all $i\in [d]$ and there exists some $j\in [d]$ such that $\mathbf{V}^{\pi}_j> \mathbf{V}^{\pi'}_j$.
}

{The general goal of MORL is to discover the \textit{Pareto front}, which is defined as the set of non-dominated policies. That is, for each policy $\pi$ in the Pareto Front, there exists no other policy $\pi'\in\Pi$ such that $\pi'\succ \pi$.
To search for the Pareto front, one common approach is to leverage a scalarization utility function $f_{\bm{\lambda}}:\mathbb{R}^d\rightarrow \mathbb{R}$ under a user preference vector $\lambda\in\Lambda$, where $\Lambda$ denotes the preference set. 
In this paper, we focus on the linear scalarization setting where $f_{\bm{\lambda}}(\bm{r})=\bm{\lambda}^\top \bm{r}$, as commonly adopted in the MORL literature~\citep{abels2019dynamic,yang2019general,basaklar2023pd,weihung2023qp,lu2023multi}.
Without loss of generality, we presume that $\Lambda$ is the $d$-dimensional unit simplex.
%Under such scalarization, we define the convex coverage set (CCS)~\citep{roijers2013survey,yang2019general}, a refinement of the Pareto front~\citep{censor1977pareto}, as the set that contains those $\mathbf{V}^{\pi}$ where the linearly scalarized utility is maximized under some preference vector, \textit{i.e.}, $\text{CCS}:=\{\mathbf{V}^{\pi}\rvert \exists \bm{\lambda}\in \Lambda \text{ s.t. }\bm{\lambda}^\top \mathbf{V}^{\pi}\geq \bm{\lambda}^\top \mathbf{V}^{\pi'}, \forall {\pi}'\in \Pi\}$, where $\Pi$ denotes the set of all stationary policies.
Notably, it has recently been shown by~\citet{lu2023multi} that any point on the Pareto front can be achieved by training a policy using linear scalarization due to the convexity of the policy-induced value function's range.
%Accordingly, we say that $\pi$ is a CCS policy if $\mathbf{V}^\pi \in \text{CCS}$.
Since the preference $\blambda_{\text{test}}$ at test time is unknown during training, our goal is to learn a preference-conditioned policy $\pi:\cS\times \Lambda\rightarrow \Delta(\cA)$ that can maximize the total discounted scalarized reward $\mathbb{E}\left[\sum^{\infty}_{t=1} \gamma^t \blambda^\top \br_t\right]$, for \textit{any} $\bm{\lambda}\in\Lambda$.}
%where \(\blambda\in \mathbb{R}^d\) represents the user preference vector. %Without loss of generality, we presume that $\blambda$ is in the $d$-dimensional probability simplex. 
%Since \(\blambda\) is unknown during training, MORL algorithms typically learn a set of policies to cover the entire preference space.
% \begin{equation}
%     \sum^{\infty}_{t=1} \gamma^t \blambda^\top \br_t,
% \end{equation}
%{\color{blue}
%    \begin{equation}
%        \mathbb{E}\left[\sum^{\infty}_{t=1} \gamma^t \blambda^\top \br_t\right],
%    \end{equation}
%}

\section{Reward-Free RL for Multi-Objective RL}
\label{sec:alg}
\vspace{-0.5em}

This section explains why MORL can be seen as a special case of RFRL. We then discuss how this perspective enhances MORL by improving generalization and sample efficiency.

{\textbf{MORL as a special case of RFRL.}
The goal of RFRL is to compute an optimal policy for \textit{any} scalar reward function $R:\mathcal{S}\times\mathcal{A}\rightarrow \mathbb{R}$ provided at test time, \textit{without} observing any reward signal during training (\ie ``reward-free") nor requiring additional environment interaction at test time. Formally, RFRL solves the following optimization problem at test time: $\text{arg}\max_\pi\mathbb{E}_{\pi,s_0\sim \mu}\big[ \sum^\infty_{t=0} \gamma^{t} r_t \big],$
where $r_t$ is the reward realization that corresponds to the test-time reward function. 

%\iclrwh{In the linear scalarization setting which is the focus of this paper,} 
MORL under linear scalarization addresses a similar problem, but presumes the vector-valued reward signal from $\mathbf{R}(s,a)$ can be observed during training, and assigns $R(s,a) = \blambda^\top \mathbf{R}(s,a)$, where \( \blambda \) (a user-specified preference vector) defines a linear combination of multiple reward components in \( \mathbf{R} \). Both RFRL and MORL aim to retrieve an optimal policy for a given reward function at test time, but their approaches differ. While MORL typically focuses on finding the Pareto front by learning a set of optimal policies for various preferences \( \blambda \), RFRL learns policies for \textit{all} possible reward functions, potentially including optimal policies for scalarized MORL rewards. RFRL achieves this by training a conditional policy network \citep{touati2021learning} or leveraging a pre-collected dataset through planning or batch RL \citep{jin2020reward}, providing a broader policy set than traditional MORL.}

\textbf{Key idea: RFRL as a source of auxiliary tasks.}
RFRL learns policies for a broader class of reward functions than required for MORL, but this can be beneficial by providing \textit{auxiliary tasks}. 
%Training only for CCS policies, as MORL aims to do, is challenging, especially for difficult preference vectors. Here, a preference vector is considered more difficult if it requires more samples for the RL agent to learn the corresponding optimal policy, often due to greater exploration needs or greater environmental stochasticity. In contrast, learning non-CCS policies is often easier and can serve as auxiliary tasks to guide training. 
Prior research has shown that incorporating auxiliary tasks improves sample efficiency and generalization in RL~\citep{jaderberg2016reinforcement,veeriah2019discovery,rafiee2022makes}. Since RFRL naturally trains policies across a spectrum of reward functions, it provides a structured way to design these auxiliary tasks.
%RFRL learns a broader range of policies than required for MORL, but this can be beneficial by providing auxiliary tasks. Training only for CCS policies, as MORL aims to do, is challenging, especially for difficult preference vectors. Here, a preference vector is considered more difficult if it requires more samples for the RL agent to learn the corresponding optimal policy, often due to greater exploration needs or greater environmental stochasticity. In contrast, learning non-CCS policies is often easier and can serve as auxiliary tasks to guide training. Prior research has shown that incorporating auxiliary tasks improves sample efficiency and generalization in RL~\citep{jaderberg2016reinforcement,rafiee2022makes,veeriah2019discovery}. Since RFRL naturally trains policies across a spectrum of reward functions, it provides a structured way to design these auxiliary tasks.}
{However, directly applying RFRL to MORL can be data-inefficient since reward-free exploration may not prioritize states that are crucial for learning the Pareto front in MORL. The key challenge is: \textit{How to utilize the auxiliary tasks offered by RFRL effectively to accelerate the learning of optimal policies in MORL?} In the sequel, we describe how RFRL can be adapted to effectively improve training in MORL.}

\vspace{-0.5em}
\subsection{Forward-Backward MORL (MORL-FB)}
\label{subsec:fbmorl_impl}
%\vspace{-1em}

In this section, we formally present MORL-FB by describing the implementation of RFRL and the key components that adapt RFRL to MORL by enhancing its learning efficiency.

\textbf{RFRL Implementation.} We implement RFRL for MORL using the state-of-the-art Forward-Backward (FB) RL algorithm \citep{touati2021learning}. The FB method learns a set of policies optimized for different reward functions by decomposing the Q-value of an optimal policy for a scalar reward function $R$ into two neural networks: $\bbF_{\theta}$ (forward representation) and $\bbB_{\omega}$ (backward representation), where $\theta$ and $\omega$ denote their parameters. This decomposition allows the Q-function for a given reward function \( R \) to be expressed as:

\vspace{-1em}
\begin{equation}
    Q(s, a, \bz_R) = \bbF_{\theta}(s,a,\bz_R)^\top \bz_R,
\end{equation}
\vspace{-1.2em}

where \( \bz_R \in \mathbb{R}^{d_z} \) is an \( d_z\)-dimensional latent vector, and both $\bbF_{\theta}$ and $\bbB_{\omega}$ are neural networks producing \( d_z \)-dimensional outputs. Intuitively, \( \bz_R \) is meant to encode the optimal policy that corresponds to the current reward function. Once a reward function $R$ is revealed, \( \bz_R \) is defined as:
% \begin{align}
% \label{eq:opt_z}
%     \bz_r = \mathbb{E}_{(s_t, a_t, s_{t+1})\sim \mathcal{D}}\left[\bbB(s_{t+1})r(s_{t+1})\right],
% \end{align}

\vspace{-0.6em}
\begin{equation}
\label{eq:opt_z}
    \bz_R = \mathbb{E}_{(s, a)\sim \mathcal{D}}\left[\bbB_{\omega}(s,a)R(s,a)\right],
\end{equation}
\vspace{-0.3em}

where \( \mathcal{D} \) represents an arbitrary state-action distribution. In our implementation, we use \( \mathcal{D} \) as the distribution induced by the replay buffer collected by the agent during training.
Using this formulation, the greedy policy for a given reward function \( R \) is defined as:
\begin{equation}
\label{eq:opt_policy}
    \pi(s, \bz_R) = \text{arg}\max_a\  \bbF_{\theta}(s,a,\bz_R)^\top \bz_R.
\end{equation}
\vspace{-0.8em}

% Here we introduce first the implementation of RFRL, and then present a key component that enables RFRL to learn more efficiently.

% \textbf{RFRL implementation.} We implemented our idea of RFRL for MORL using the state-of-the-art RFRL algorithm: Forward-backward (FB) RL \citep{touati2021learning}. How does FBRL learn a set of policies for possible reward functions? The central idea of FB is to decompose the Q-value of the optimal policy for a given reward function via two neural networks $\bbF$ (forward representation) and $\bbB$ (backward representation) such that the Q-function of the optimal policy for a reward function $r$ is given as:
% \begin{align}
%     Q(s, a, \bz_r) = \bbF(s,a,\bz_r)^\top \bz_r,
% \end{align}
% where $\bz_r \in \mathbb{R}^d$ represents a $d$-dimensional vector, $\bbF$ and $\bbB$ are networks outputting $d$-dimensional vectors. Intuitively, $\bz_r$ indicates what reward function we are facing now and is given as:
% \begin{align}
% \label{eq:opt_z}
%     \bz_r = \mathbb{E}_{s,a,s^\prime \sim \mathcal{D}}\left[\bbB(s^\prime)r(s,a)\right],
% \end{align}
% where $\mathcal{D}$ denotes an arbitrary state-action distribution. In our paper, we 
% refer $\mathcal{D}$ to as a replay buffer collected by an agent during training time. The greedy policy for a reward function $r$ is therefore defined as:
% \begin{align}
% \label{eq:opt_policy}
%     \pi(s, \bz_r) = \text{arg}\max_a \bbF(s,a,\bz_r)^\top \bz_r
% \end{align}
% We outline the algorithm for training in~\Cref{alg:MORL-FB} and elaborate the details below.

Now we are ready to present the design of MORL-FB at both training time and test time.

\begin{figure*}[ht]
    \centering
    \begin{minipage}[t]{0.6\textwidth}
        \vspace{0pt}
        \centering
\includegraphics[width=1.0\linewidth]{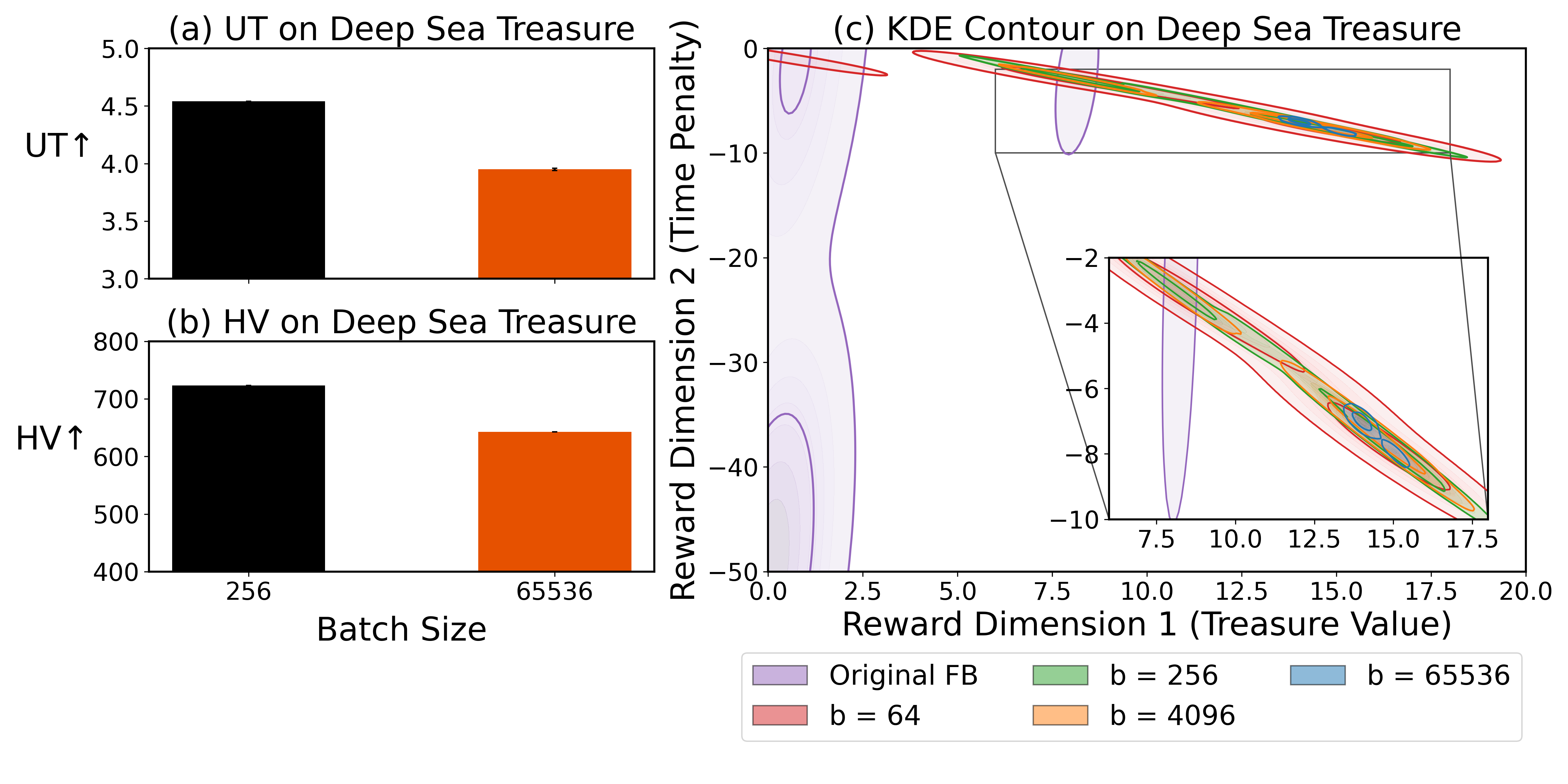}
    \vspace{-7mm}
        \caption{A motivating experiment on Deep Sea Treasure. (a)(b) Training performance (UT and HV defined in the sequel) of MORL-FB under different batch sizes for $\hat{\bm{z}}_{\blambda}$. (c) KDE contour of return vector distributions of $\pi(\cdot,\bm{z})$ induced by $\hat{\bm{z}}_{\blambda}$ \pchreb{(with various batch sizes $b$)} and $\hat{\bm{z}}\sim\mathcal{N}(0, \mathbb{I}^{d_z})$. This shows that $\hat{\bm{z}}_{\bm{\lambda}}$ corresponds to learning for more diverse and relevant behavior in MORL than $\boldsymbol{z}_{\blambda}$ and the $\bm{z}$ sampling strategy of the original FB. The detailed configuration is provided in Appendix~\ref{app:experiment_config}.}
        \label{fig:dst_ablation}
    \end{minipage}%
    \hfill
    \begin{minipage}[t]{0.38\textwidth}
        \vspace{-7pt}
        \centering
        \begin{algorithm}[H]
            \caption{MORL-FB}
            \footnotesize
            \label{alg:Pseudo_MORL-FB}
            \begin{algorithmic}[1]
                \State \textbf{Input:} $\bz$ dimension $d_z$, sample number $n_s$
                \State Initialize replay buffer $\mathcal{M} \leftarrow \varnothing$
                \For{$\text{each iteration } i$}
                    \State Sample preference $\boldsymbol{\lambda}$ uniformly from $\Lambda$
                    \State $\bz \leftarrow \textsc{PG-Explore}(\boldsymbol{\lambda})$
                    \State Generate rollouts using $\bz$
                    \State Sample $n_s$ transitions $\mathcal{D} \sim \mathcal{M}$
                    \State Update FB networks $\bbF_\theta$, $\bbB_\omega$ and policy $\pi$
                \EndFor

                \Function{PG-Explore}{$\boldsymbol{\lambda}$}
                    \State Sample $n_s$ transitions $\mathcal{D} \sim \mathcal{M}$
                    \State $\bz \leftarrow \sum_{(s,a,\mathbf{r},s')\in \mathcal{D}} \frac{\mathbf{B}_{\omega}(s,a)\mathbf{r}^\top\boldsymbol{\lambda}}{n_s}$
                    \State Normalize $\bz$ to $\bz \leftarrow \sqrt{d_z}\frac{\bz}{\Vert \bz \Vert_{2}}$
                    \State \Return $\bz$
                \EndFunction
            \end{algorithmic}
        \end{algorithm}
    \end{minipage}
    
\end{figure*}

\textbf{Test Time:} At test time, we can easily adapt Equation~(\ref{eq:opt_z}) by replacing \( R(s,a) \) with a user-specified scalarized multi-objective reward based on a preference vector \( \blambda \) as $R(s,a) = \blambda^\top \mathbf{R}(s,a)$.
%\begin{align}
%    r(s) = \blambda^\top \mathbf{R}(s).
%\end{align}
Next, given the learned $\bbF_{\theta}$ and $\bbB_{\omega}$, we compute the corresponding \( \bz_R \) and use it in the policy defined by Equation~(\ref{eq:opt_policy}). This effectively retrieves an optimal policy for the given preference \( \blambda \).

\textbf{Training with Preference-Guided Exploration:}
During training, \( \bbF_{\theta} \), \( \bbB_{\omega} \), and \( \pi \) must be trained by sampling \( \bz \) and conditioning the networks on these sampled values. Since the test-time user preference $\blambda_\text{test}$ is unknown at this stage, we cannot directly compute \( \bz \) using Equation~(\ref{eq:opt_z}). At a high level, training on a diverse set of \( \bz \) samples is equivalent to training the agent on a variety of reward functions, since \( \bz \) is inherently linked to rewards through~\Cref{eq:opt_z}.

In principle, \( \bz \) can be sampled from any distribution without restriction. In~\citep{touati2021learning}, \( \bz \) is drawn from a standard normal distribution \( \mathcal{N}(0, \mathbb{I}^{d_z}) \) in a \( d_z \)-dimensional space. However, we found that this approach leads to poor sample efficiency when testing the agent on MORL tasks.
We hypothesize that sampling \( \bz \) from a normal distribution produces representations that differ significantly from the actual \( \bz_R \) obtained from a preference-weighted multi-objective reward function (see~\Cref{fig:z_distribution} for a visualization of empirical $\bm{z}$ distributions).

To address this issue, we propose \textit{Preference-Guided Exploration} (PG-Explore), which constructs a more relevant \( \bz \) distribution via sampling guided by preference-weighted rewards. The design of PG-Explore builds on the following insights:

\begin{itemize}[leftmargin=*]
    \item \textbf{Using $\{\boldsymbol{z}_{\blambda}\}_{\blambda\in\Lambda}$ only leads to limited exploration of ${\bm{z}}$}: Recall that in MORL, we can observe multi-objective rewards \( \mathbf{R}(s,a) \) (or its noisy version) during training. One direct approach is to compute \( \bz \) as:
\vspace{-0.6em}
\begin{equation}
\label{eq:opt_z_pref}
    \boldsymbol{z}_{\blambda}=\mathbb{E}[\mathbf{B}_{\omega}(s,a)\blambda^{\top} \mathbf{R}(s,a)]\stackrel{(a)}{=}\mathbb{E}[\mathbf{B}_{\omega}(s,a) \mathbf{R}(s,a)^{\top} \blambda] \stackrel{(b)}{=}\underbrace{(\mathbb{E}[\mathbf{B}_{\omega}(s,a) \mathbf{R}(s,a)^\top])}_{=:\mathbf{H}}\blambda,
\end{equation}
\vspace{-1em}

where (a) holds \iclrwh{due to the fact that} $\blambda^{\top} \mathbf{R}(s,a)$ is a scalar and can be swapped in the matrix multiplication with $\mathbf{B}_{\omega}(s,a)$ and (b) follows from that $\blambda$ can be moved out of the expectation. 

Equation~(\ref{eq:opt_z_pref}) suggests that \textit{$\boldsymbol{z}_{\blambda}$ is in the span of $d$ preference-agnostic column vectors of the $d_z\times d$ matrix $\mathbf{H}$}, for any preference $\blambda$. In practice, since the number of objectives $d$ is usually much smaller than $d_z$, the coverage of $\{\boldsymbol{z}_{\blambda}\}_{\blambda\in\Lambda}$ in $\mathbb{R}^{d_z}$ can be extremely small. This leads to limited exploration of $\boldsymbol{z}$ during training such that the agent can commit to a set of improper $\boldsymbol{z}$, especially in the early training stage when \( \bbF_{\theta} \) and \( \bbB_{\omega} \) are not well trained.
\end{itemize}

\begin{itemize}[leftmargin=*]
\item \textbf{Constructing $\hat{\bm{z}}_{\bm{\lambda}}$ by mini-batch sampling for exploration}: To encourage exploration of $\boldsymbol{z}$ relevant to MORL, we propose a conceptually simple and yet effective technique that leverages mini-batch sampling to construct $\hat{\bm{z}}_{\bm{\lambda}}$. Specifically, we sample a batch of $n_s$ data samples (denoted by $\mathcal{D}$) from the replay buffer and compute $\hat{\bm{z}}_{\blambda}=\sum_{(s,a,\mathbf{r},s’)\in \mathcal{D}} \mathbf{B}_{\omega}(s,a)\mathbf{r}^\top \blambda/{n_s}$.
%sample preferences \( \boldsymbol{\lambda} \) uniformly from a unit simplex and then 
Figure~\ref{fig:dst_ablation} shows a comparison of training with ${\bm{z}}_{\bm{\lambda}}$, $\hat{\bm{z}}_{\bm{\lambda}}$ under various batch sizes, and $\bm{z}$ drawn from $\mathcal{N}(0, \mathbb{I}^{d_z})$ as in the original FB method, in Deep Sea Treasure (DST), which is a goal-oriented navigation task with two-dimensional rewards as (treasure value, step cost). As the true ${\bm{z}}_{\bm{\lambda}}$ is not available, we use $\hat{\bm{z}}_{\bm{\lambda}}$ with a large $n_s$ as a surrogate for ${\bm{z}}_{\bm{\lambda}}$.
We can see that: (i) $\hat{\bm{z}}_{\bm{\lambda}}$ indeed corresponds to learning more diverse behavior than just learning for $\boldsymbol{z}_{\blambda}$. (ii) $\hat{\bm{z}}_{\bm{\lambda}}$'s are more relevant to the reward functions in MORL encountered at test time than the $\bm{z}$ sampled from $\mathcal{N}(0, \mathbb{I}^{d_z})$, improving sample efficiency.
%We use~\Cref{eq:opt_z_pref} for both exploration and network updates.
\end{itemize}

%{\textbf{Sampling \( \bz \) as Auxiliary Tasks:}

\begin{itemize}[leftmargin=*]
\item \textbf{Learning induced by $\hat{\bm{z}}_{\bm{\lambda}}$ serves as auxiliary tasks:}
%For a given preference \( \blambda \), we will calculate different \( \bz_{\lambda} \) values (calculated based on the samples drawn from the replay buffer) during training, allowing the agent to train on more than one \( \bz \) for the same preference. 
Recall that in PG-Explore, we construct $\hat{\bm{z}}_{\blambda}=\sum_{(s,a,\mathbf{r},s’)\in \mathcal{D}} \mathbf{B}_{\omega}(s,a)\mathbf{r}^\top \blambda/{n_s}$ by mini-batch sampling.
%As defined in~\Cref{eq:opt_z_pref}, \( \bz \) is computed as the expected backward representation \( \bbB_{\omega}(s') \) sampled from the replay buffer, scaled by the preference-weighted reward. 
This means that for any given preference \( \blambda \), the agent can learn beyond $\bm{z}_{\blambda}$ and, moreover, from multiple values \( \bz \) from different batches of transitions, providing richer learning signals.
This approach is closely related to the auxiliary tasks in deep RL, where training objectives that are not directly or totally aligned with the target objective have been shown to accelerate learning \citep{jaderberg2016reinforcement,veeriah2019discovery,rafiee2022makes}. 
%A visualization shown in~\Cref{fig:walker_73} in the sequel shows the empirical return vectors that correspond to the sampled $\bm{z}$, manifesting how RFRL can be seen as an auxiliary task. Note that MORL-FB does not require additional samples for the inference of \( \bz \) at test time. In the testing phase, we directly leverage the samples from the replay buffer maintained during training.
\end{itemize}

{\textbf{Training Objective Functions:}
(i) \textit{Measure loss}: To train the \( \bbF_{\theta} \) and \( \bbB_{\omega} \) networks in MORL-FB, we use the standard measure loss $\mathcal{L}_{\text{M}}(\bbF_{\theta}, \bbB_{\omega}; \boldsymbol{z_{\blambda}})$, which minimizes the Bellman residual on the successor measure \citep{ahmed2023zero}. $\bbF_{\bar{\theta}}$ and $\bbB_{\bar{\omega}}$ are target networks. This loss is defined as: }
\begin{align}
\mathcal{L}_{\mathrm{M}}(\mathbf{F}_{\theta}, \mathbf{B}_{\omega}; \boldsymbol{z}_{\lambda}) &= \mathbb{E}_{\substack{(s_{t}, a_{t}, s_{t+1}) \sim \mathcal{D}\\ (s',a') \sim \mathcal{D}}} \left[ \left( \mathbf{F}_{\theta}(s_{t}, a_{t}, \boldsymbol{z}_{\lambda})^{\top} \mathbf{B}_{\omega}(s',a') \right.\right. \nonumber\\
&\quad \left.\left. - \gamma \mathbf{F}_{\bar{\theta}}(s_{t+1}, \pi(s_{t+1}, \boldsymbol{z}_{\lambda}), \boldsymbol{z}_{\lambda})^{\top} \mathbf{B}_{\bar{\omega}}(s',a') \right)^{2} \right] \nonumber\\
&\quad - 2 \mathbb{E}_{(s_{t}, a_{t}, s_{t+1}) \sim \mathcal{D}} [ \mathbf{F}_{\theta}(s_{t}, a_{t}, \boldsymbol{z}_{\lambda})^{\top} \mathbf{B}_{\omega}(s_{t+1},a_{t+1})].
\end{align}

\normalsize

(ii) \textit{Auxiliary Q loss}: In the context of MORL, we propose to employ an auxiliary Q loss to facilitate the learning of FB representations from the \textit{observed reward vectors}, instead of the \textit{pseudo rewards} in the original FB (also see the ablation study in Section~\ref{sec:exp:ablation}). Specifically, the Q-loss is constructed as the squared temporal difference error represented in $\mathbf{F_{\theta}}$ and $\mathbf{B_{\omega}}$, and the {transitions} are sampled from the replay buffer to compute $\bm{z}_{\bm{\lambda}}$ via our preference-guided function:
%(Reviewer3 Q5: Eq. (9) should be consistent with Eq. (12) in Appendix)

% \begin{footnotesize}

\begin{align}
\label{eq:q_loss_main}
\mathcal{L}_{Q}(\mathbf{F_{\theta}}; \boldsymbol{z_{\blambda}}) &= \mathbb{E}_{(s, a, \boldsymbol{r}, s') \sim \mathcal{D}} \big[ ( \mathbf{F_{\theta}}(s, a, \boldsymbol{z_{\blambda}})^{\top} \boldsymbol{z_{\blambda}} - ( \boldsymbol{\lambda}^{\top} \boldsymbol{r} + \gamma \mathbf{F_{\bar{\theta}}}(s', \pi(s', \boldsymbol{z_{\blambda}}), \boldsymbol{z_{\blambda}})^{\top} \boldsymbol{z_{\blambda}}) )^{2} \big].
\end{align}
% \end{footnotesize}
\vspace{-1em}

We summarize the implementation in \Cref{alg:Pseudo_MORL-FB}. The detailed pseudo code (\Cref{alg:MORL-FB}) and further details about loss functions and implementation are provided in \Cref{app:algorithm}. Note that the FB framework can use either state-dependent or state-action-dependent backward representation, and both perform well in practice (see Appendix~\ref{ap:sar}). As the original FB~\citep{ahmed2023zero} presumes a state-dependent design, we focus mainly on state-dependent ones in the subsequent experiments. 

\vspace{-0.6em}
\section{Experiments}
\label{sec:exp}
\vspace{-0.3em}
%In this section, we showcase the effectiveness of MORL-FB via extensive experiments on benchmark tasks and delineate how it achieves favorable sample efficiency and generalization across preferences. 

%\subsection{Setup}
{\textbf{Evaluation Domains.} We leverage the MO-Gymnasium benchmark suite~\citep{felten2023toolkit} and
consider various discrete and continuous control tasks. For the continuous control tasks, we consider robot locomotion tasks in Multi-objective MuJoCo with up to 5 objectives, including Walker2d, Halfcheetah2d, Ant3d, Hopper3d, Humanoid2d, and Humanoid5d. Each environment presents a unique set of objectives, \eg the goal of Ant3d is to optimize both x-axis and y-axis speeds while minimizing energy consumption. The experiments of the discrete control tasks are provided in Appendix~\ref{ap:discrete_env}. The detailed configurations for all tasks are provided in Appendix~\ref{app:experiment_config}.}

{\textbf{Benchmark Methods.} To evaluate the effectiveness of our proposed approach, we compare MORL-FB against various benchmark methods, including: (i) \textit{Single preference-conditioned policy methods}: PD-MORL \citep{basaklar2023pd}, Q-Pensieve \citep{weihung2023qp}, CAPQL \citep{lu2023multi}, Envelope Q-Learning (EQL) \citep{yang2019general}, and PCN \citep{reymond2022pareto}; (ii) \textit{Multi-policy MORL}: PG-MORL \citep{xu2020prediction}, SFOLS \citep{alegre2022optimistic}, MORL/D \citep{felten2024multi}, GPI-LS, and GPI-PD \citep{alegre2023sample} ; (iii) \textit{Reward-free RL}: We take the original FB approach \citep{ahmed2023zero} as a baseline.} 

Regarding PD-MORL, Q-Pensieve, and FB, we leverage their official implementations from~\citep{basaklar2023pd,weihung2023qp,ahmed2023zero}.
To ensure a fair comparison among all the benchmark methods, we adopt the standard PD-MORL without the auxiliary pre-trained preference interpolator, which essentially requires a substantial amount of additional data for pre-training and could bias the comparison. Regarding CAPQL, EQL, PCN, PG-MORL, MORL/D, GPI-LS, and GPI-PD, we leverage the implementation of MORL-Baselines \citep{felten2023toolkit} for better reproducibility. As the PG-MORL in MORL-Baselines can only support two-objective tasks, we extend this code base to accommodate those tasks beyond two objectives. On the other hand, the original SFOLS only focuses on discrete control tasks by default. For a more thorough comparison, we utilize its official implementation for discrete problems and extend SFOLS with a TD3 backbone for evaluation on continuous control. 
{Moreover, we apply hyperparameter optimization to the benchmark algorithms and MORL-FB. Appendix~\ref{sub:exp_setup} details the selection range of hyperparameters and the final selected values.}
For all the tasks, we run each algorithm for 3 million environment steps, which is comparable to most of the existing MORL studies. Below we report the average performance and the empirical standard deviation over 5 random seeds for each task.
More detailed configurations of the experiments and benchmark methods are provided in Appendix~\ref{app:experiment_config}.

\begin{figure*}[!h]
    \centering
    \includegraphics[width=0.85\textwidth]{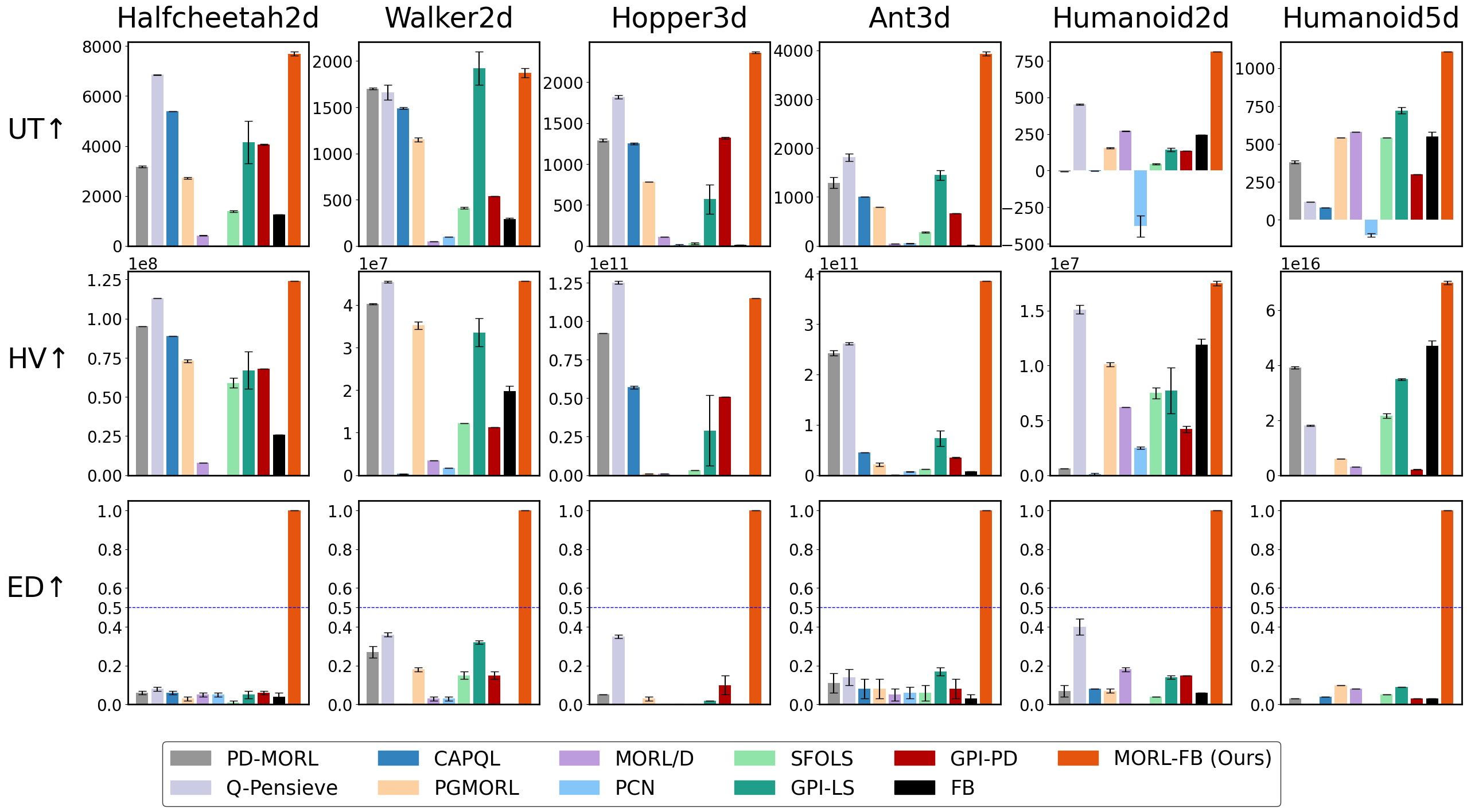}
    \caption{
    Evaluation of MORL-FB and several MORL benchmark algorithms on diverse continuous control tasks within the MO-Gymnasium suite, assessing performance using key metrics. These results demonstrate the clear advantage of MORL-FB across all tested benchmarks.}
    \label{fig:continuous_bar}
    \vspace{-1em}

\end{figure*}

%The metrics evaluated are HV, UT, and ED. In this figure, the ED metric is computed using MORL-FB as the baseline for comparison with other algorithms.
%\subsubsection{Evaluation metrics}
\textbf{Evaluation Metrics.}
We evaluate the performance of each algorithm using three metrics that are widely used in the MORL literature \citep{van2014multi,yang2019general,kyriakis2022pareto,basaklar2023pd,weihung2023qp,lu2023multi}: 
%\zwh{Cite more papers or more classic papers}
\begin{itemize}[leftmargin=*]
    \item \textbf{Utility (UT)}: To evaluate the scalarized total reward across different preferences at an aggregate level, we employ the utility metric defined as $\mathbb{E}_{\bm{\lambda}}[\sum_{t}\bm{\lambda}^\top \bm{r}_t]$, where the expectation is taken with respect to the uniform distribution over the preference set $\Lambda$ (\ie $d$-dimensional unit simplex).
    \item \textbf{Hypervolume (HV)}: As a standard metric in the literature of general multi-objective optimization, hypervolume naturally captures the inherent trade-off among different objective functions using one aggregate scalar value \citep{zitzler1999multiobjective}. Specifically, given a reference point $\bm{u}_{\text{ref}}\in \mathbb{R}^d$ and any collection for return vectors $\mathcal{U} \subseteq \mathbb{R}^d$, the hypervolume of $\mathcal{U} $ can be formally defined as $\HV(\mathcal{U}; \bm{u}_{\text{ref}}) := \mu\Big(\bigcup_{\bm{u}\in \mathcal{U}}\limits \big\{\bm{y}\big\rvert \bm{u}\succeq \bm{y} \succeq \bm{u}_{\text{ref}}\big\} \Big),$
    where $\mu(\cdot)$ denotes the $d$-dimensional Lebesgue measure. In practice, $\bm{u}_{\text{ref}}$ is selected based on the range of possible total return and is task-dependent. The configuration of $\bm{u}_{\text{ref}}$ for each MORL task is provided in Appendix~\ref{app:experiment_config}.  

    \item \textbf{Episodic Dominance (ED)}: As a metric complementary to HV and UT, ED is meant to capture the relative strength of a pair of algorithms under different preferences. Specifically, given any two algorithms $\texttt{ALG}_1, \texttt{ALG}_2$, we define $\text{ED}(\texttt{ALG}_1, \texttt{ALG}_2):=\mathbb{E}_{\bm{\lambda}}[\mathbb{I}\{\bm{\lambda}^\top \bm{g}(\tau_{\texttt{ALG}_{1}})\geq\bm{\lambda}^\top \bm{g}(\tau_{\texttt{ALG}_{2}})\}]$, where $\bm{g}(\cdot)$ denotes the trajectory-wise cumulative return vector, $\tau_{\texttt{ALG}_{1}}$ and $\tau_{\texttt{ALG}_{2}}$ are the trajectories generated under the policies of ${\texttt{ALG}_{1}}$ and ${\texttt{ALG}_{2}}$, and $\bm{\lambda}$ is drawn uniformly from $\Lambda$. {Note that we use 500 uniformly sampled preference vectors and evaluate across 5 distinct random seeds for each preference vector for statistical robustness.}
\end{itemize}

To ensure rigorous evaluation, we further follow the guidelines of \citep{agarwal2021deep} by taking the normalized UT scores and reporting the aggregated performance across tasks in median, mean, and interquartile mean (IQM).
Regarding the normalized scores, we follow the procedure in~\citep{fu2020d4rl}, which (i) employs a \textit{random policy}—where actions are selected uniformly at random— as the baseline with normalized score of 0 and (ii) an \textit{expert policy} trained by single-objective SAC—as the topline with normalized score of 100. The above normalization is done on a per-preference basis.

\label{sec:exp:results}
\textbf{Does the reward-free viewpoint of MORL-FB improve sample efficiency over the benchmark methods?}
\Cref{fig:continuous_bar} shows the performance of all the methods in UT, HV, and ED for continuous control tasks. Regarding ED, for each baseline algorithm $\texttt{ALG}$, we report $\text{ED}(\texttt{ALG},\texttt{MORL-FB})$ to show the pairwise comparison. We can make the following observations: (i) MORL-FB achieves either the best or close to the best UT and HV among all methods on all the tasks, regardless of the number of objectives. This showcases that MORL-FB is indeed sample-efficient in the sense that it can discover a diverse collection of high-performing policies across various preferences using only as few as 3 million samples as used by {the \textit{expert policy}}. (ii) Given that $\text{ED}(\texttt{ALG},\texttt{MORL-FB})$ are consistently smaller than 0.5 for all baselines, we see that MORL-FB outperforms all benchmark methods (including the state-of-the-art methods like PD-MORL and Q-Pensieve), under most preferences. (iii) PD-MORL and Q-Pensieve perform well on two-objective tasks (\eg Halfcheetah2d and Walker2d) but underperform when the number of objectives is larger (\eg Ant3d and Humanoid5d). {Additional results on discrete control environments can be found in \Cref{ap:discrete_env}.}

\begin{wrapfigure}{r}{0.5\textwidth} 
    \centering
    \vspace{-1em}
    \includegraphics[width=\linewidth]{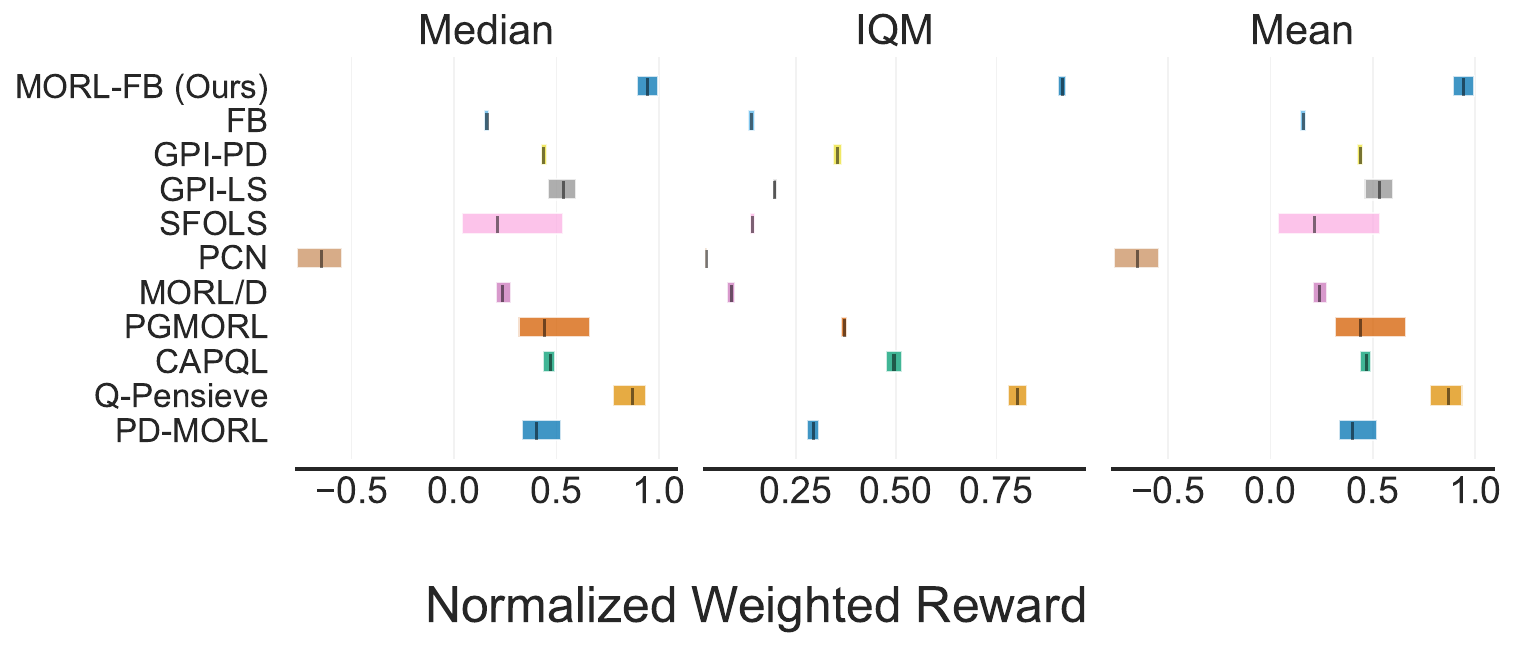}
    \caption{
    Evaluation of MORL-FB and several MORL benchmark algorithms using aggregate metrics, including median, mean, and interquartile mean (IQM). These results show the superior performance of MORL-FB across all metrics.}
    \label{fig:continuous-IQM}
    \vspace{-2em} 
\end{wrapfigure}

Moreover, regarding the aggregated results,~\Cref{fig:continuous-IQM} shows that MORL-FB reliably outperforms the benchmark methods both in conventional statistics (\eg mean and median) and robust metrics like IQM. MORL-FB achieves the best IQM scores by a large margin \textit{vis-à-vis} other methods, confirming the significant improvements over the state-of-the-art MORL.

%\textbf{MORL-FB as an Auxiliary Task} 
Recall that MORL-FB leverages PG-Explore to address the fundamental exploration issue of vanilla FB, which suffers from sample inefficiency in MORL. Remarkably, the per-task results in~\Cref{fig:continuous_bar} and aggregated results in~\Cref{fig:continuous-IQM} show that MORL-FB enjoys significantly better UT and HV across tasks. Accordingly, the ED scores $\text{ED}(\texttt{FB},\texttt{MORL-FB})$ remain nearly zero in all tasks.

\textbf{Does MORL-FB achieve effective generalization across preferences?}
To better assess the generalization capabilities of MORL-FB across preferences, we further evaluate the algorithms in a stylized setting where they are trained only on a small set of preference vectors $\Lambda_{\text{train}}$ (rather than the whole $\Lambda$) and aim for generalization over $\Lambda$ at test time. Specifically, for a $d$-objective task, we let $\Lambda_{\text{train}}$ include only the standard basis preferences, \ie $d$-dimensional one-hot vectors, and the uniform preference vector $[1/d,\cdots,1/d]$. 
The testing setup is exactly the same as that for \Cref{fig:continuous_bar}.
Here we focus on comparing MORL-FB to PD-MORL and Q-Pensieve, which are the top two benchmark methods in~\Cref{fig:continuous_bar} and utilize conditioned networks of structures different from MORL-FB.

\begin{figure}[!t]
    \centering
    \begin{minipage}[b]{0.48\textwidth}
        \centering
        \includegraphics[width=0.9\linewidth]{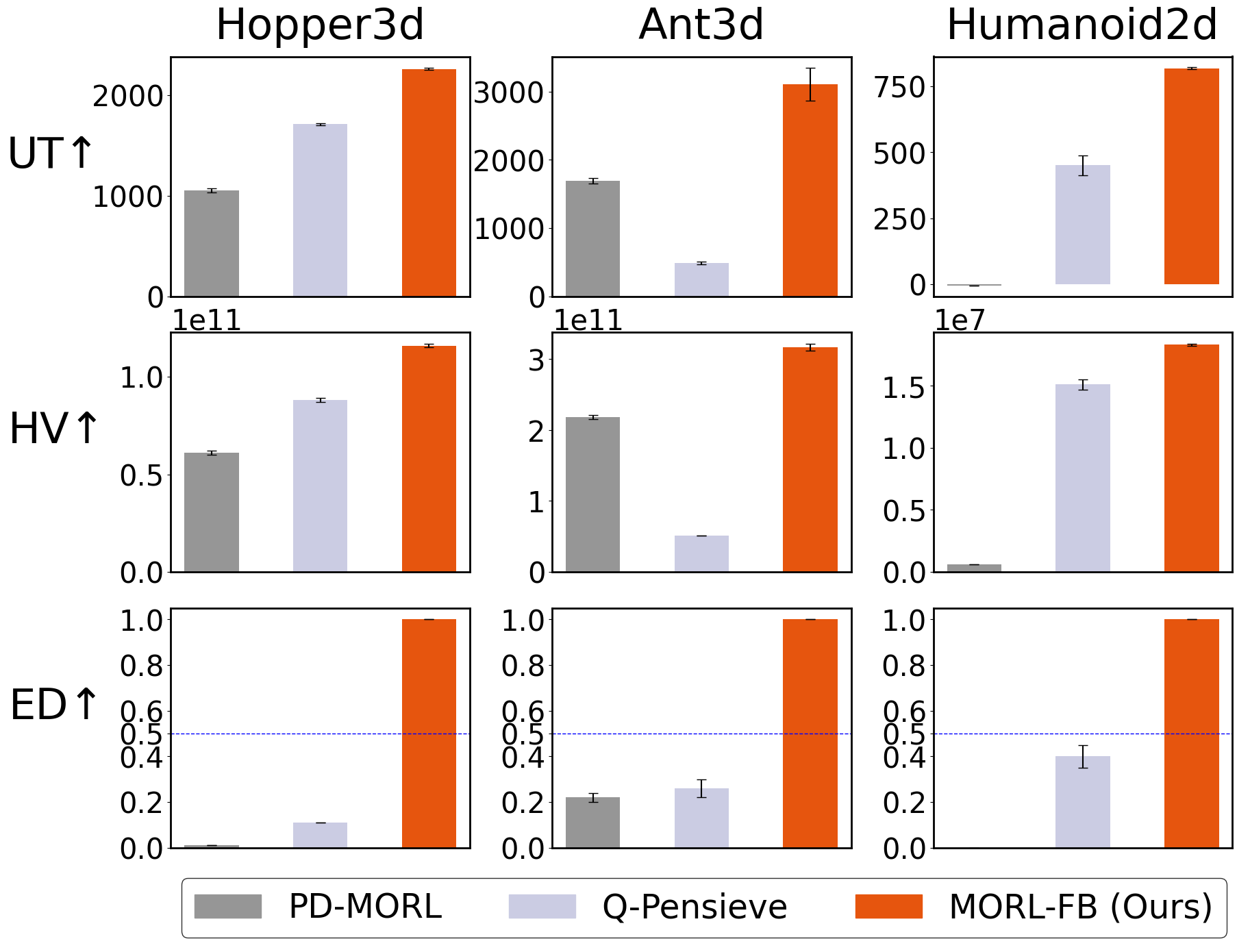}
        \caption{Evaluation of MORL-FB and benchmark methods (PD-MORL and Q-Pensieve) under a reduced preference set $\Lambda_{\text{train}}$ during training: These demonstrate the generalization capability of MORL-FB across preferences.}
        \label{fig:continuous_constraint_bar}
    \end{minipage}
    \hfill
    \begin{minipage}[b]{0.48\textwidth}
        \centering
          \includegraphics[width=0.7\linewidth]{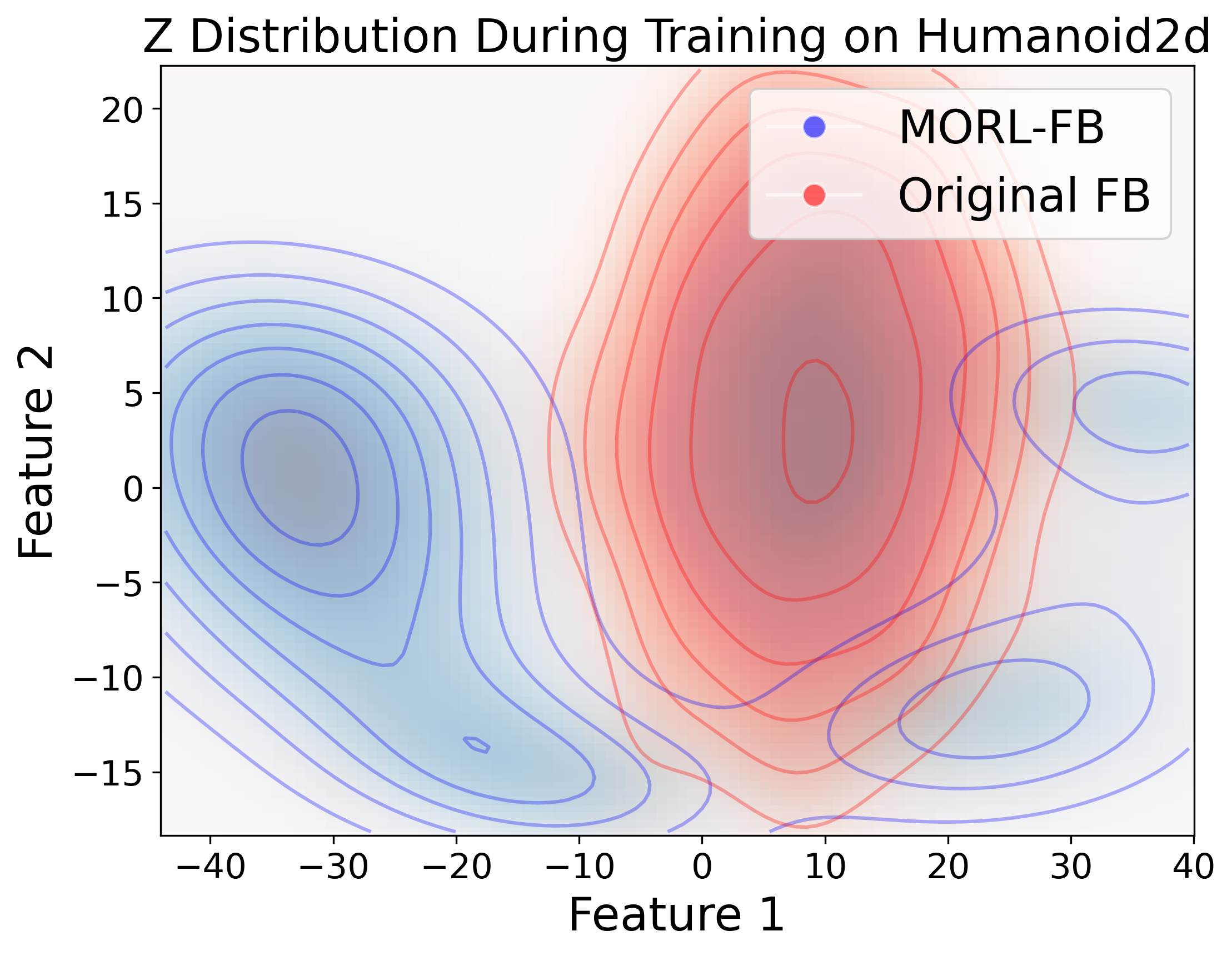}
        \caption{Empirical $\boldsymbol{z}$ distribution under t-SNE for Humanoid2d with MORL-FB (preference-guided sampling, blue) and original FB (standard normal, red): The multi-modal distribution observed with MORL-FB suggests a more diverse set of latent representations compared to unimodal nature of original FB.}
        %The visualization is performed using t-SNE to project the latent representations onto a two-dimensional space. 
        \label{fig:z_distribution}
    \end{minipage}
    \vspace{-0.6em}

\end{figure}

As shown in \Cref{fig:continuous_constraint_bar}, both PD-MORL and Q-Pensieve exhibit a notable decline in performance across all three metrics compared to those in~\Cref{fig:continuous_bar}. %This decline indicates that these benchmark methods struggle to generalize effectively when trained with a restricted set of preferences. 
In contrast, MORL-FB maintains consistent performance across the evaluated tasks, with only minimal degradation in UT and HV values compared to~\Cref{fig:continuous_bar}. These findings showcase that MORL-FB can generalize more effectively over the entire preference set, even when trained on a limited set of preference vectors.
%, thereby enhancing the generalization properties of the FB representation while excelling in multi-objective reinforcement learning tasks.
More detailed results, such as the numerical values and the aggregated performance (\eg IQM) are in Appendix~\ref{app:exp}.

\begin{figure}[!h]
    \centering
    \begin{minipage}[t]{0.48\textwidth}
    \centering
    \includegraphics[width=1\textwidth]{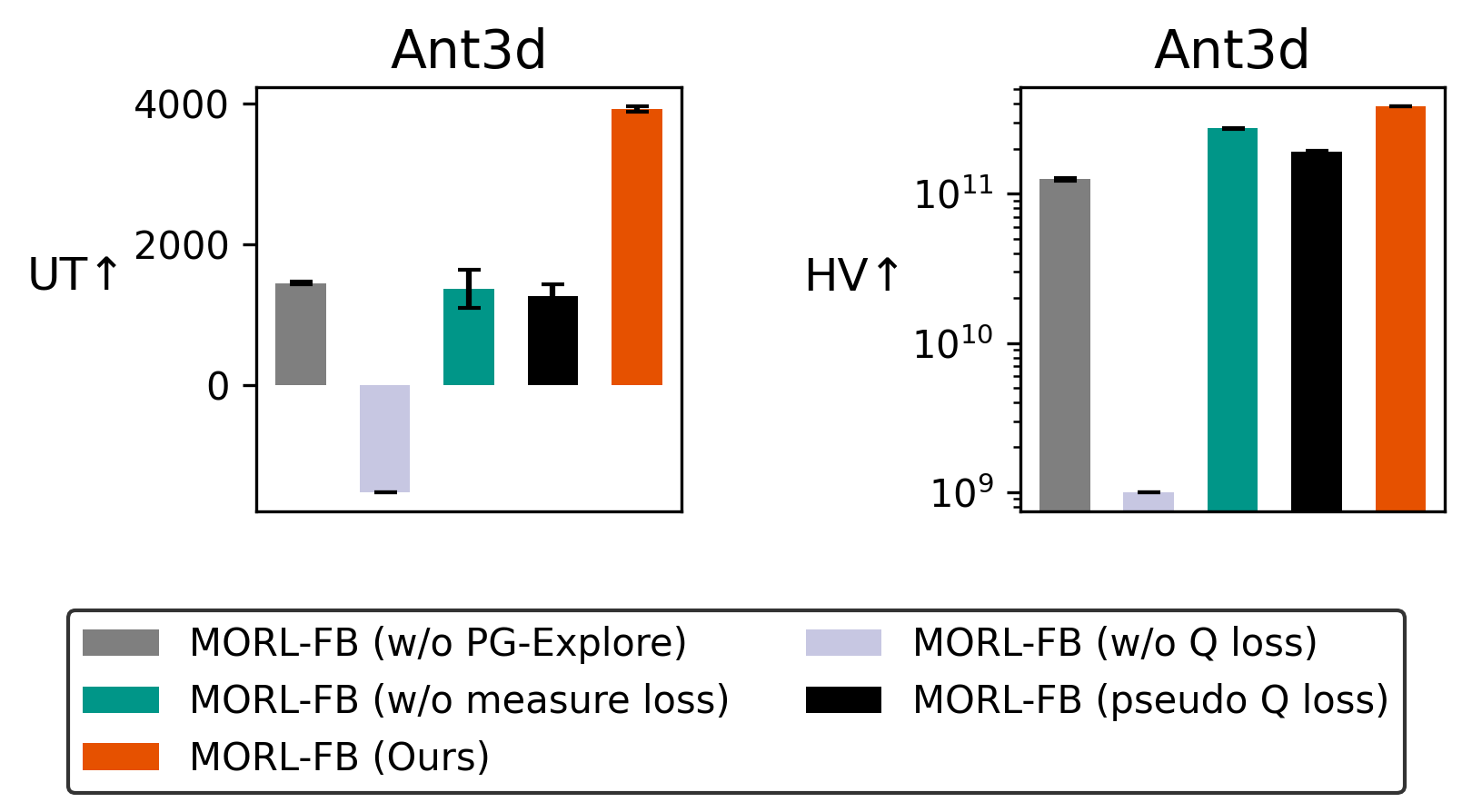}
    \caption{Evaluation of MORL-FB and its ablated versions on the Ant3d task. The results highlight the importance of PG-Explore and auxiliary losses, as removing these components leads to performance degradation.}
        \label{fig:ablation_ant3d}
    \end{minipage}
    \hfill
    \begin{minipage}[t]{0.48\textwidth}
        \centering
        \includegraphics[width=0.63\linewidth]{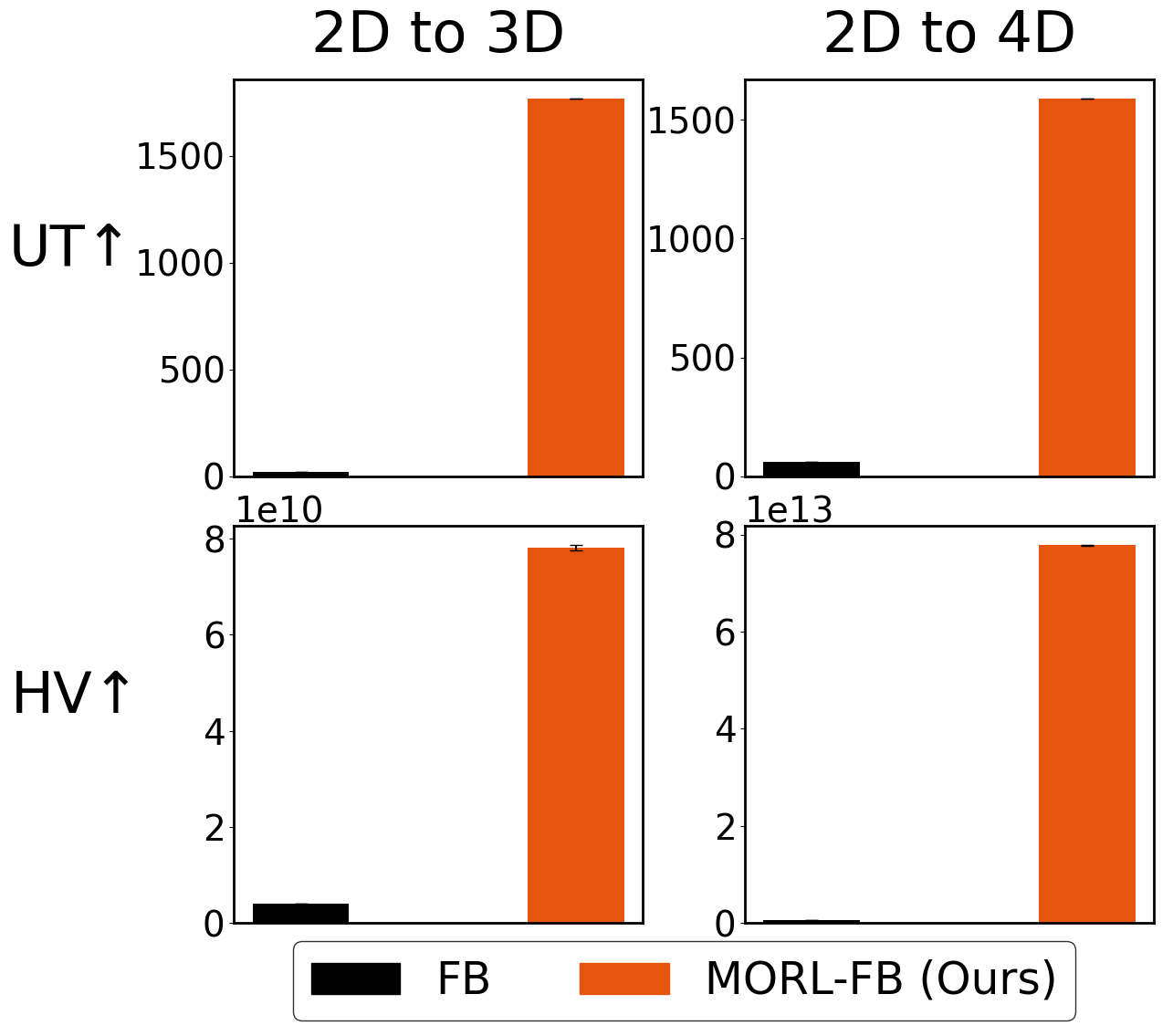}
        \caption{Zero-shot cross-objective transfer from Hopper2d to Hopper3d and Hopper4d using vanilla FB and MORL-FB:  Results demonstrate effective transfer by MORL-FB, supporting the efficacy of its proposed enhancements.}
        \label{fig:objective_transform_bar}
    \end{minipage}
    \vspace{-1.5em}
\end{figure}

\subsection{Ablation Study}
\label{sec:exp:ablation}

{\textbf{Preference-Guided Exploration (PG-Explore).}
To investigate the benefits of sampling $\bz$ from a preference-guided distribution, we perform an ablation study on comparing the proposed MORL-FB and a variant of MORL-FB that samples $\bz$ from $\mathcal{N}(0,\mathbb{I}^{d_z} )$, \ie the distribution adopted by the vanilla FB.
\Cref{fig:ablation_ant3d} (specifically the bars in gray and orange) shows that MORL-FB indeed benefits significantly from a preference-guided distribution across the tested task, highlighting its importance in enabling directed exploration and sample-efficient policy learning.} 

Moreover, we visualize the empirical distributions of the sampled $\bz$ of MORL-FB and the original FB. Specifically, we record the $\bz$ vectors used throughout training and apply t-SNE \citep{van2008visualizing} for visualization in a two-dimensional space. The results on Humanoid2d in \Cref{fig:z_distribution} show that sampling $\bz$ from a normal distribution results in a unimodal empirical distribution (contours in red). By contrast, MORL-FB with the preference-guided sampling exhibits a multi-modal distribution (contours in blue), indicating a richer and more diverse set of $\bz$ distributions. This multi-modality allows MORL-FB to better capture the underlying reward structure, achieving improved generalization and adaptation to various objectives. 
More visualization of $\bz$ distributions for other tasks can be found in Appendix~\ref{app:exp}.

\textbf{Auxiliary Q loss.} To corroborate the efficacy of the auxiliary Q loss, we further conduct an ablation study on this term. From~\Cref{fig:ablation_ant3d} (specifically the bars in black and orange), the Q loss can facilitate the learning of forward and backward representations in MORL-FB and thereby boost the performance in both UT and HV. More ablation results across environments are in Appendix~\ref{section:ablation}.
\vspace{-0.5em}
% {\color{blue}
% \textbf{FB decomposition.} (Reviewer4 2nd reply: Explain MO-TD3’s poor performance on MO-Humanoid?) We include an ablation comparing MORL-FB with MO-TD3 [Basaklar et al., 2023] with same hyperparameters, which uses vectorized Q-functions without FB. Results show that FB decomposition indeed improves both UT and HV. (https://imgur.com/a/Z7sAIMz)
% }

 \subsection{Zero-Shot Cross-Objective Transfer}
\label{sec:exp:cross}
{Recall that one salient feature of MORL-FB is to use $\bz$ to encode the $\bm{\lambda}$-dependent scalarized reward function. Accordingly, MORL-FB is endowed with the ability to achieve zero-shot transfer even across tasks of different number of objectives. 
Such \textit{zero-shot cross-objective transferability} allows us to add new factors to the reward function without the need for retraining and hence is a very useful feature in practice. 
To validate this, we use MORL-FB to learn the $\bbF$ and $\bbB$ networks on Hopper2d and directly evaluate them on Hopper3d and Hopper4d, which involve additional reward terms like ``jump height" and ``z-axis speed", at test time. Details of the environment configurations are in Appendox~\ref{app:experiment_config}.
We conducted the same evaluation for vanilla FB~\citep{ahmed2023zero} as a baseline. From~\Cref{fig:objective_transform_bar}, vanilla FB cannot achieve effective cross-objective transfer given that vanilla FB already suffers in the standard MORL setting (cf.~\Cref{fig:continuous_bar}). By contrast, MORL-FB achieves effective transfer across objectives in a zero-shot manner, corroborating the proposed enhancements.}

\vspace{-1em}

\section{Related Work}
\label{sec:related}
\vspace{-0.4em}

\subsection{Multi-Objective RL}
\label{sec:related:MORL}
\vspace{-2mm}

\textbf{Single preference-conditioned policy methods.} Single preference-conditioned policy methods learn one policy network that adapts to different objective trade-offs by conditioning on preference. Many of these methods employ scalarization techniques~\citep{van2013scalarized,yang2019general}, transforming multi-objective problems into weighted single-objective problems. They allow policies to dynamically adjust at inference time. However, relying solely on linear reward aggregation without proper representation learning can lead to suboptimal solutions. To address this, CAPQL~\citep{lu2023multi} introduced the concave reward terms for better optimization landscapes, while CN-DER~\citep{abels2019dynamic} proposed a preference-conditioned Q-network with an experience replay mechanism to handle dynamic weights and mitigate non-stationarity. Q-Pensieve~\citep{weihung2023qp} improved sample efficiency by reusing past policy snapshots. Without using scalarization,~\citet{abdolmaleki2020distributional} proposed to learn action distributions per objective and fitted a parametric policy via supervised learning. To improve adaptability to diverse and unseen preference vectors, methods like PCN~\citep{reymond2022pareto} formulated MORL as a classification problem, and MOAC~\citep{zhoufinite} finds Pareto-stationary points by adapting multi-gradient descent to MORL without scalarization. PD-MORL~\citep{basaklar2023pd} also trains a single preference-conditioned network but directly incorporates preference vectors, for example, through cosine similarity measures within its value-function update rule to efficiently learn a comprehensive set of policies across the continuous preference space.

\textbf{Multi-policy methods.} Multi-policy methods explicitly learn multiple policies to cover the Pareto front, capturing diverse trade-offs in the training process. A key challenge in multi-policy MORL is efficiently constructing a coverage set that represents the full Pareto front while maintaining scalability. To refine policy selection and handle dominated actions, \citet{lizotte2012linear} introduced a structured approach using linear value function approximation. Subsequent methods~\citep{kyriakis2022pareto,van2014multi,xu2020prediction} focused on improving exploration efficiency across the preference space but lacked structured learning mechanisms to generalize across diverse preferences. Building on the idea of incorporating structure into policy learning, {\citet{felten2024multi,mossalam2016multi}} extended structured learning for multi-objective RL, employing decomposition and sequential single-objective optimization to enhance efficiency. However, scalability and adaptability remained challenges. DG-MORL~\citep{lu2024demonstration} leveraged demonstrations and a self-evolving mechanism to improve scalability. As for improving adaptability, \citet{mossalam2016multi} extended Optimistic Linear Support (OLS) to deep RL, constructing a convex coverage set through a sequence of single-objective, providing a structured way to represent diverse trade-offs, but lacked effective transferability. 
%\pchreb{Successor features (SFs)~\citep{alegre2022optimistic} addressed this by enabling adaptation to new tasks without additional environment interactions, improving generalization across objectives. However, SF-based methods require handcrafted reward features $\phi(s,a,s')$, which could miss important aspects of the environment and limit adaptability. In contrast, FB learns the $F$ and $B$ representations without relying on handcrafted features.} 

%{Due to the page limit, we defer the related work on RFRL to Appendix~\ref{app:related_RFRL}.}

\vspace{-6mm}
\pchreb{\subsection{Successor Features for Transfer Across Reward Functions}}

\pchreb{One related transfer setting in RL is to learn policies for all reward functions that are linear combinations of a finite set of known features. \citet{barreto2017successor} proposed the \textit{successor feature} (SF), which reflects the state-action occupancy of a policy and can be viewed as an extension of the classic successor representation~\citep{dayan1993improving}. By design, SFs achieves transfer across reward functions in two ways: (i) For a fixed policy $\pi$, SFs can enable fast policy evaluation of $\pi$ across different reward functions; (ii) Given a set of policies, SFs can be combined with the generalized policy improvement (GPI) update to generate a new policy that is no worse than the given set of policies and has a sub-optimality gap characterized by the task differences, for any reward function~\citep{barreto2017successor,barreto2018transfer}. 
Accordingly, SFs has been combined with deep neural networks and applied to various tasks like subgoal extraction~\citep{kulkarni2016deep} and robot navigation~\citep{zhang2017deep}. 
Subsequently, \citet{borsa2019universal} extended the SFs by decoupling the policy and task description for better representational flexibility. Moreover, \citet{alegre2022optimistic} applied SFs and GPI to solve MORL by enabling transfer across different scalarized reward functions. \citet{chua2024learningsuccessorfeaturessimple} proposed a practical method for learning the SFs directly from pixels for image-based control. More recently, \citet{zhang2024sf} established the convergence analysis with generalization guarantees for SFs under neural function approximation.}

%We also refer the readers to~\citep{zhu2023transfer} for a detailed comparison of  SF-based methods.

\pchreb{Despite the transfer capability, the limitations of SFs and its variants are mainly two-fold: (i) SFs require a set of pre-defined reward features, which could miss important aspects of the environment and limit adaptability. (ii) While SFs can achieve fast policy evaluation across reward functions, SFs cannot directly induce an optimal policy when given an arbitrary new reward function at test time.}

%However, SF-based methods require handcrafted reward features $\phi(s,a,s')$, which could miss important aspects of the environment and limit adaptability. In contrast, FB learns the $F$ and $B$ representations without relying on handcrafted features. 

\vspace{-3mm}
\pchreb{\subsection{Reward-Free RL}}
\label{sec:related:RFRL}
\vspace{-2mm}

%\pchreb{\textbf{Successor features and successor measures.} 
\pchreb{Among the studies in the RFRL literature, \citep{touati2021learning,ahmed2023zero} are the most relevant to our work.} 
\pchreb{
To address the aforementioned issues of SFs, \citet{touati2021learning} adopted the reward-free MDP formulation and proposed a low-rank model termed the forward-backward (FB) representations, which capture the state-action occupancy of the optimal policies for all the reward functions by learning the required features directly from data. The FB framework has been implemented and validated for both reward-free discrete control~\citep{touati2021learning} and continuous control~\citep{ahmed2023zero}. Subsequently, the framework of FB representations has been extended to various settings, such as offline RL with low-quality data~\citep{jeen2024zero}, online unsupervised RL~\citep{sun2025unsupervised}, imitation learning~\citep{pirotta2024fast,tirinzoni2025zero}, and partially-observable MDPs~\citep{jeen2025zero}.} 

%That said, one known limitation of SFs and its variants is the need for a set of pre-defined features, which can be difficult to construct in practice. To address the above issue, \citep{touati2021learning} proposed a low-rank model termed the forward-backward (FB) representations, which capture the state-action occupancy of the optimal policies by learning the required features directly from data. The FB framework has been implemented and validated for both reward-free discrete control~\citep{touati2021learning} and continuous control~\citep{ahmed2023zero}.
%\textbf{Reward-free exploration.} 

\pchreb{Another line of RFRL research focuses on achieving provably efficient exploration without using any reward information, in both tabular settings \citep{jin2020reward,kaufmann2021adaptive,menard2021fast,wu2022gap} and under function approximation \citep{qiu2021reward,wagenmaker2022reward,wang2020reward,zanette2020provably,zhang2021reward}.}

\pchreb{Inspired by the RFRL literature, we propose to rethink MORL via RFRL and adapt the FB method to boost the sample efficiency and generalization in MORL.}

\section{Conclusion, Limitations, and Future Work}
\label{sec:conclusion}
\vspace{-0.6em}

We propose MORL-FB, \pchreb{which offers a new perspective by rethinking MORL through the lens of RFRL.
%and provides the first systematic adaptation of RFRL algorithms to solving MORL. 
By using RFRL as auxiliary tasks, MORL-FB provides the first systematic adaptation of RFRL to MORL and enhances its sample efficiency with critical algorithmic enhancements, including preference-guided exploration with mini-batch sampling and an auxiliary Q loss based on the observed reward vectors.} 
MORL-FB achieves strong performance in a variety of discrete and continuous control tasks, offering superior efficiency, better generalization, and zero-shot cross-objective transfer.
A key limitation, inherited from FB-based RFRL, is the need for more advanced exploration, especially in complex or sparse-reward environments, where dedicated strategies are crucial.
\pchreb{Future work includes exploring RFRL methods beyond FB, such as representation learning of successor measures~\citep{agarwal2025proto,farebrother} and learning distance-preserving state representations~\citep{park2024foundation}, to further reveal RFRL’s advantages in MORL.}

% successor features~\citep{chua2024learningsuccessorfeaturessimple}
%In this paper, we propose MORL-FB, which rethinks MORL from the perspective of RFRL. We leverage RFRL as a source of auxiliary tasks and boost the sample efficiency of RFRL by preference-guided exploration in the context of MORL. Through extensive experiments on MORL benchmark tasks, we show that MORL-FB indeed serves as a promising MORL method given its superior sample efficiency, improved generalization, and zero-shot cross-objective transfer capability.
%A key limitation of MORL-FB, inherited from FB representations of RFRL, is the necessity for more sophisticated exploration strategies on exploring environment. This is particularly pronounced in complex environments, especially those characterized by sparse rewards, where dedicated exploration techniques would be required to achieve optimal performance.
%One future work is to explore more RFRL methods beyond FB, such as other variants of learning successor measures~\citep{farebrother} or successor features~\citep{chua2024learningsuccessorfeaturessimple}, which could provide further insights into RFRL's advantages in MORL.

\section*{Acknowledgment}
This research is partially supported by the National Science and Technology Council (NSTC) of Taiwan under Grant Numbers 114-2628-E-A49-002 and 114-2634-F-A49-002-MBK. 
We also thank the National Center for High-performance Computing (NCHC) for providing computational and storage resources.

\section*{Ethics statement}
\vspace{-2mm}

Our work develops and evaluates reinforcement learning methods purely in simulated environments, without involving human subjects or sensitive data. This submission follows the code of ethics.
\section*{Reproducibility statement}
We release our code in the supplementary material and describe the commands needed to execute the code in a Readme file attached in the supplementary material. Additionally, we attach the list of package dependencies which can be used to build the environment.

\section*{The Use of Large Language Models (LLMs)}
Large language models (LLMs) were employed exclusively for language editing and polishing of the manuscript. They were not used for designing methods, conducting experiments, or analyzing results.
%\subsubsection*{Author Contributions}
%If you'd like to, you may include  a section for author contributions as is done in many journals. This is optional and at the discretion of the authors.

%\subsubsection*{Acknowledgments}
%Use unnumbered third level headings for the acknowledgments. All acknowledgments, including those to funding agencies, go at the end of the paper.

\bibliography{iclr2026_conference}

@inproceedings{pirotta2024fast,
  title={Fast Imitation via Behavior Foundation Models},
  author={Pirotta, Matteo and Tirinzoni, Andrea and Touati, Ahmed and Lazaric, Alessandro and Ollivier, Yann},
  booktitle={International Conference on Learning Representations},
  year={2024}
}

@inproceedings{sun2025unsupervised,
  title={Unsupervised Zero-Shot Reinforcement Learning via Dual-Value Forward-Backward Representation},
  author={Sun, Jingbo and Tu, Songjun and Li, Haoran and Liu, Xin and Chen, Yaran and Chen, Ke and Zhao, Dongbin and others},
  booktitle={International Conference on Learning Representations},
  year={2025}
}

@article{jeen2024zero,
  title={Zero-shot reinforcement learning from low quality data},
  author={Jeen, Scott and Bewley, Tom and Cullen, Jonathan},
  journal={Advances in Neural Information Processing Systems},
  volume={37},
  pages={16894--16942},
  year={2024}
}

@inproceedings{park2024foundation,
  title={Foundation Policies with {Hilbert} Representations},
  author={Park, Seohong and Kreiman, Tobias and Levine, Sergey},
  booktitle={International Conference on Machine Learning},
  pages={39737--39761},
  year={2024}
}

@inproceedings{jeen2025zero,
  title={Zero-Shot Reinforcement Learning Under Partial Observability},
  author={Jeen, Scott and Bewley, Tom and Cullen, Jonathan},
  booktitle={Reinforcement Learning Conference},
  year={2025}
}

@InProceedings{agarwal2025proto,
  title = {Proto Successor Measure: Representing the Behavior Space of an {RL} Agent},
  author = {Agarwal, Siddhant and Sikchi, Harshit and Stone, Peter and Zhang, Amy},
  booktitle = {International Conference on Machine Learning},
  pages = {566--586},
  year = 	{2025}
}

@inproceedings{tirinzoni2025zero,
  title={Zero-Shot Whole-Body Humanoid Control via Behavioral Foundation Models},
  author={Tirinzoni, Andrea and Touati, Ahmed and Farebrother, Jesse and Guzek, Mateusz and Kanervisto, Anssi and Xu, Yingchen and Lazaric, Alessandro and Pirotta, Matteo},
  booktitle={International Conference on Learning Representations},
  year={2025}
}

@inproceedings{barreto2018transfer,
  title={Transfer in deep reinforcement learning using successor features and generalised policy improvement},
  author={Barreto, Andre and Borsa, Diana and Quan, John and Schaul, Tom and Silver, David and Hessel, Matteo and Mankowitz, Daniel and Zidek, Augustin and Munos, Remi},
  booktitle={International Conference on Machine Learning},
  pages={501--510},
  year={2018}
}

@inproceedings{zhang2024sf,
  title={{SF-DQN}: Provable knowledge transfer using successor feature for deep reinforcement learning},
  author={Zhang, Shuai and Fernando, Heshan Devaka and Liu, Miao and Murugesan, Keerthiram and Lu, Songtao and Chen, Pin-Yu and Chen, Tianyi and Wang, Meng},
  booktitle={International Conference on Machine Learning},
  pages={58897--58934},
  year={2024}
}

@inproceedings{zhang2017deep,
  title={Deep reinforcement learning with successor features for navigation across similar environments},
  author={Zhang, Jingwei and Springenberg, Jost Tobias and Boedecker, Joschka and Burgard, Wolfram},
  booktitle={IEEE/RSJ International Conference on Intelligent Robots and Systems (IROS)},
  pages={2371--2378},
  year={2017}
}

@article{kulkarni2016deep,
  title={Deep successor reinforcement learning},
  author={Kulkarni, Tejas D and Saeedi, Ardavan and Gautam, Simanta and Gershman, Samuel J},
  journal={arXiv preprint arXiv:1606.02396},
  year={2016}
}

@inproceedings{jin2020reward,
  title={Reward-free exploration for reinforcement learning},
  author={Jin, Chi and Krishnamurthy, Akshay and Simchowitz, Max and Yu, Tiancheng},
  booktitle={International Conference on Machine Learning},
  year={2020}
}

@inproceedings{alegre2022optimistic,
  title={Optimistic linear support and successor features as a basis for optimal policy transfer},
  author={Alegre, Lucas Nunes and Bazzan, Ana and Da Silva, Bruno C},
  booktitle={International Conference on Machine Learning},
  year={2022}
}

@inproceedings{reymond2022pareto,
  title={Pareto conditioned networks},
  author={Reymond, Mathieu and Bargiacchi, Eugenio and Now{\'e}, Ann},
  booktitle={International Conference on Autonomous Agents and Multiagent Systems},
  year={2022}
}

@inproceedings{basaklar2023pd,
  title={{PD-MORL: Preference-driven multi-Objective reinforcement learning algorithm}},
  author={Basaklar, Toygun and Gumussoy, Suat and Ogras, Umit},
  booktitle={International Conference on Learning Representations},
  year={2023}
}

@article{dayan1993improving,
  title={Improving generalization for temporal difference learning: The successor representation},
  author={Dayan, Peter},
  journal={Neural computation},
  year={1993},
  publisher={MIT Press}
}

@inproceedings{borsa2019universal,
  title={Universal Successor Features Approximators},
  author={Borsa, Diana and Barreto, Andre and Quan, John and Mankowitz, Daniel J and van Hasselt, Hado and Munos, Remi and Silver, David and Schaul, Tom},
  booktitle={International Conference on Learning Representations},
  year={2019}
}

@inproceedings{kaufmann2021adaptive,
  title={Adaptive reward-free exploration},
  author={Kaufmann, Emilie and M{\'e}nard, Pierre and Domingues, Omar Darwiche and Jonsson, Anders and Leurent, Edouard and Valko, Michal},
  booktitle={Algorithmic Learning Theory},
  year={2021}
}

@inproceedings{zhang2021reward,
  title={Reward-free model-based reinforcement learning with linear function approximation},
  author={Zhang, Weitong and Zhou, Dongruo and Gu, Quanquan},
  booktitle={Advances in Neural Information Processing Systems},
  year={2021}
}

@inproceedings{zanette2020provably,
  title={Provably efficient reward-agnostic navigation with linear value iteration},
  author={Zanette, Andrea and Lazaric, Alessandro and Kochenderfer, Mykel J and Brunskill, Emma},
  booktitle={Advances in Neural Information Processing Systems},
  year={2020}
}

@inproceedings{wu2022gap,
  title={Gap-dependent unsupervised exploration for reinforcement learning},
  author={Wu, Jingfeng and Braverman, Vladimir and Yang, Lin},
  booktitle={International Conference on Artificial Intelligence and Statistics},
  year={2022}
}

@inproceedings{menard2021fast,
  title={Fast active learning for pure exploration in reinforcement learning},
  author={M{\'e}nard, Pierre and Domingues, Omar Darwiche and Jonsson, Anders and Kaufmann, Emilie and Leurent, Edouard and Valko, Michal},
  booktitle={International Conference on Machine Learning},
  year={2021}
}

@inproceedings{barreto2017successor,
  title={Successor features for transfer in reinforcement learning},
  author={Barreto, Andr{\'e} and Dabney, Will and Munos, R{\'e}mi and Hunt, Jonathan J and Schaul, Tom and van Hasselt, Hado P and Silver, David},
  booktitle={Advances in Neural Information Processing Systems},
  year={2017}
}

@inproceedings{wang2020reward,
  title={On reward-free reinforcement learning with linear function approximation},
  author={Wang, Ruosong and Du, Simon S and Yang, Lin and Salakhutdinov, Russ R},
  booktitle={Advances in Neural Information Processing Systems},
  year={2020}
}

@inproceedings{wagenmaker2022reward,
  title={{Reward-free RL is no harder than reward-aware RL in linear markov decision processes}},
  author={Wagenmaker, Andrew J and Chen, Yifang and Simchowitz, Max and Du, Simon and Jamieson, Kevin},
  booktitle={International Conference on Machine Learning},
  year={2022}
}

@inproceedings{weihung2023qp,
    author = {Wei Hung and Bo-Kai Huang and Ping-Chun Hsieh and Xi Liu},
    title = {{Q-Pensieve: Boosting sample efficiency of multi-objective RL through memory sharing of Q-snapshots}},
    booktitle = {International Conference on Learning Representations},
    year = {2023}
}

@inproceedings{ahmed2023zero,
    author = {Touati, Ahmed and Rapin, Jérémy and Ollivier, Yann},
    title = {Does Zero-Shot Reinforcement Learning Exist?},
    booktitle = {International Conference on Learning Representations},
    year = {2023}
}

@inproceedings{touati2021learning,
  title={Learning one representation to optimize all rewards},
  author={Touati, Ahmed and Ollivier, Yann},
  booktitle={Advances in Neural Information Processing Systems},
  year={2021}
}

@inproceedings{agarwal2021deep,
  title={Deep Reinforcement Learning at the Edge of the Statistical Precipice},
  author={Agarwal, Rishabh and Schwarzer, Max and Castro, Pablo Samuel
          and Courville, Aaron and Bellemare, Marc G},
  booktitle={Advances in Neural Information Processing Systems},
  year= {2021}
}

@inproceedings{yang2019general,
    author = {Runzhe Yang and Xingyuan Sun and Karthik Narasimhan},
    title = {A Generalized Algorithm for Multi-Objective Reinforcement Learning and Policy Adaptation},
    booktitle = {Advances in Neural Information Processing Systems},
    year = {2019}
}

@inproceedings{xu2020prediction,
  author={Xu, Jie and Tian, Yunsheng and Ma, Pingchuan and Rus, Daniela and Sueda, Shinjiro and Matusik, Wojciech},
  title={Prediction-Guided Multi-Objective Reinforcement Learning for Continuous Robot Control},
  booktitle={International Conference on Machine Learning},
  year={2020}
}

@inproceedings{abdolmaleki2020distributional,
    author = {Abbas Abdolmaleki and Sandy H. Huang and Leonard Hasenclever and Michael Neunert and H. Francis Song and Martina Zambelli and Murilo F. Martins and Nicolas Heess and Raia Hadsell and Martin Riedmiller},
    title = {A Distributional View on Multi-Objective Policy Optimization},
    booktitle = {International Conference on Machine Learning},
    year = {2020}
}

@article{vamplew2011empirical,
  title={Empirical evaluation methods for multiobjective reinforcement learning algorithms},
  author={Vamplew, Peter and Dazeley, Richard and Berry, Adam and Issabekov, Rustam and Dekker, Evan},
  journal={Machine learning},
  volume={84},
  number={1},
  pages={51--80},
  year={2011},
  publisher={Springer}
}

@article{fu2020d4rl,
  title={{D4RL: Datasets for deep data-driven reinforcement learning}},
  author={Fu, Justin and Kumar, Aviral and Nachum, Ofir and Tucker, George and Levine, Sergey},
  journal={arXiv preprint arXiv:2004.07219},
  year={2020}
}

@article{felten2024multi,
  title={{Multi-objective reinforcement learning based on decomposition: A taxonomy and framework}},
  author={Felten, Florian and Talbi, El-Ghazali and Danoy, Gr{\'e}goire},
  journal={Journal of Artificial Intelligence Research},
  year={2024}
}

@inproceedings{
    Hayes_2022,
    author = {Hayes, Conor F. and Rădulescu, Roxana and Bargiacchi, Eugenio and Källström, Johan and Macfarlane, Matthew and Reymond, Mathieu and Verstraeten, Timothy and Zintgraf, Luisa M. and Dazeley, Richard and Heintz, Fredrik and Howley, Enda and Irissappane, Athirai A. and Mannion, Patrick and Nowé, Ann and Ramos, Gabriel and Restelli, Marcello and Vamplew, Peter and Roijers, Diederik M.},
    title = {A practical guide to multi-objective reinforcement learning and planning},
    booktitle = {International Conference on Autonomous Agents and Multiagent Systems},
    year ={2022} 
}

@inproceedings{alegre2023sample,
  title={Sample-Efficient Multi-Objective Learning via Generalized Policy Improvement Prioritization},
  author={Alegre, Lucas Nunes and Bazzan, Ana LC and Roijers, Diederik M and Now{\'e}, Ann and da Silva, Bruno C},
  booktitle={International Conference on Autonomous Agents and Multiagent Systems},
  year={2023}
}

@article{sutton2018reinforcement,
  title={Reinforcement learning: An introduction},
  author={Sutton, Richard S. and Barto, Andrew G.},
  journal={A Bradford Book},
  year={2018}
}

@article{lu2024demonstration,
  title={Demonstration Guided Multi-Objective Reinforcement Learning},
  author={Lu, Junlin and Mannion, Patrick and Mason, Karl},
  journal={arXiv preprint arXiv:2404.03997},
  year={2024}
}

@inproceedings{lu2023multi,
  title={{Multi-objective reinforcement learning: Convexity, stationarity and pareto optimality}},
  author={Lu, Haoye and Herman, Daniel and Yu, Yaoliang},
  booktitle={International Conference on Learning Representations},
  year={2023}
}

@inproceedings{qiu2021reward,
  title={{On reward-free RL with kernel and neural function approximations: Single-agent MDP and Markov game}},
  author={Qiu, Shuang and Ye, Jieping and Wang, Zhaoran and Yang, Zhuoran},
  booktitle={International Conference on Machine Learning},
  year={2021}
}

@inproceedings{felten2023toolkit,
  title={A toolkit for reliable benchmarking and research in multi-objective reinforcement learning},
  author={Felten, Florian and Alegre, Lucas Nunes and Nowe, Ann and Bazzan, Ana and Talbi, El Ghazali and Danoy, Gr{\'e}goire and C da Silva, Bruno},
  booktitle={Advances in Neural Information Processing Systems},
  year={2023}
}

@article{lizotte2012linear,
  title={{Linear fitted-Q iteration with multiple reward functions}},
  author={Lizotte, Daniel J and Bowling, Michael and Murphy, Susan A},
  journal={Journal of Machine Learning Research},
  year={2012}
}

@article{van2014multi,
  title={Multi-objective reinforcement learning using sets of {Pareto} dominating policies},
  author={Van Moffaert, Kristof and Now{\'e}, Ann},
  journal={Journal of Machine Learning Research},
  year={2014}
}

@article{mossalam2016multi,
  title={Multi-objective deep reinforcement learning},
  author={Mossalam, Hossam and Assael, Yannis M and Roijers, Diederik M and Whiteson, Shimon},
  journal={arXiv preprint arXiv:1610.02707},
  year={2016}
}

@inproceedings{kyriakis2022pareto,
  title={Pareto policy adaptation},
  author={Kyriakis, Panagiotis and Deshmukh, Jyotirmoy},
  booktitle={International Conference on Learning Representations},
  year={2022}
}

@inproceedings{abels2019dynamic,
  title={Dynamic weights in multi-objective deep reinforcement learning},
  author={Abels, Axel and Roijers, Diederik and Lenaerts, Tom and Now{\'e}, Ann and Steckelmacher, Denis},
  booktitle={International Conference on Machine Learning},
  year={2019}
}

@article{zitzler1999multiobjective,
  title={{Multiobjective evolutionary algorithms: A comparative case study and the strength Pareto approach}},
  author={Zitzler, Eckart and Thiele, Lothar},
  journal={IEEE Transactions on Evolutionary Computation},
  year={1999}
}

@article{van2008visualizing,
  title={{Visualizing data using t-SNE}},
  author={Van der Maaten, Laurens and Hinton, Geoffrey},
  journal={Journal of machine learning research},
  year={2008}
}

@article{jaderberg2016reinforcement,
  title={Reinforcement learning with unsupervised auxiliary tasks},
  author={Jaderberg, Max and Mnih, Volodymyr and Czarnecki, Wojciech Marian and Schaul, Tom and Leibo, Joel Z and Silver, David and Kavukcuoglu, Koray},
  journal={arXiv preprint arXiv:1611.05397},
  year={2016}
}

@article{rafiee2022makes,
  title={What makes useful auxiliary tasks in reinforcement learning: investigating the effect of the target policy},
  author={Rafiee, Banafsheh and Jin, Jun and Luo, Jun and White, Adam},
  journal={arXiv preprint arXiv:2204.00565},
  year={2022}
}

@article{veeriah2019discovery,
  title={Discovery of useful questions as auxiliary tasks},
  author={Veeriah, Vivek and Hessel, Matteo and Xu, Zhongwen and Rajendran, Janarthanan and Lewis, Richard L and Oh, Junhyuk and van Hasselt, Hado P and Silver, David and Singh, Satinder},
  journal={Advances in Neural Information Processing Systems},
  volume={32},
  year={2019}
}

@inproceedings{eimer2023hyperparametersreinforcementlearningtune,
    author = {Theresa Eimer and Marius Lindauer and Roberta Raileanu},
    title = {Hyperparameters in Reinforcement Learning and How To Tune Them},
    booktitle = {International Conference on Machine Learning},
    year = {2023}
}

@inproceedings{fujimoto2018addressingfunctionapproximationerror,
    author = {Scott Fujimoto and Herke van Hoof and David Meger},
    title = {Addressing Function Approximation Error in Actor-Critic Methods},
    booktitle = {International Conference on Machine Learning},
    year = {2018}
}

@inproceedings{zhoufinite,
  title={Finite-Time Convergence and Sample Complexity of Actor-Critic Multi-Objective Reinforcement Learning},
  author={Zhou, Tianchen and Hairi, FNU and Yang, Haibo and Liu, Jia and Tong, Tian and Yang, Fan and Momma, Michinari and Gao, Yan},
  booktitle={International Conference on Machine Learning},
  year = {2024}
}

@inproceedings{van2013scalarized,
  title={Scalarized multi-objective reinforcement learning: Novel design techniques},
  author={Van Moffaert, Kristof and Drugan, Madalina M and Now{\'e}, Ann},
  booktitle={Adaptive Dynamic Programming and Reinforcement Learning},
  year={2013}
}

@inproceedings{farebrother,
    author = {Jesse Farebrother and Joshua Greaves and Rishabh Agarwal and Charline Le Lan and Ross Goroshin and Pablo Samuel Castro and Marc G. Bellemare},
    title = {Proto-Value Networks: Scaling Representation Learning with Auxiliary Tasks},
    booktitle = {International Conference on Learning Representations},
    year = {2023} 
}

@inproceedings{chua2024learningsuccessorfeaturessimple,
    author = {Raymond Chua and Arna Ghosh and Christos Kaplanis and Blake A. Richards and Doina Precup},
    title = {Learning Successor Features the Simple Way},
    booktitle = {Advances in Neural Information Processing Systems},
    year = {2024}
}

@inproceedings{romoff2018rewardestimationvariancereduction,
    author = {Joshua Romoff and Peter Henderson and Alexandre Piché and Vincent Francois-Lavet and Joelle Pineau},
    title = {Reward Estimation for Variance Reduction in Deep Reinforcement Learning},
    booktitle = {Conference on Robot Learning},
    year = {2018}
}

@inproceedings{
hu2022distributional,
title={Distributional Reward Estimation for Effective Multi-agent Deep Reinforcement Learning},
author={Jifeng Hu and Yanchao Sun and Hechang Chen and Sili Huang and haiyin piao and Yi Chang and Lichao Sun},
booktitle={Advances in Neural Information Processing Systems},
year={2022}
}

@inproceedings{tamar2015optimizing,
  title={Optimizing the CVaR via sampling},
  author={Tamar, Aviv and Glassner, Yonatan and Mannor, Shie},
  booktitle={Proceedings of the AAAI Conference on Artificial Intelligence},
  volume={29},
  number={1},
  year={2015}
}

@inproceedings{lin2024tchebycheffscalarization,
    author = {Xi Lin and Xiaoyuan Zhang and Zhiyuan Yang and Fei Liu and Zhenkun Wang and Qingfu Zhang},
    title = {Smooth Tchebycheff Scalarization for Multi-Objective Optimization},
    booktitle = {International Conference on Machine Learning},
    year = {2024}
}

@article{qiu2024traversingparetooptimalpolicies,
  title={Traversing {Pareto} Optimal Policies: Provably Efficient Multi-Objective Reinforcement Learning},
  author={Shuang Qiu and Dake Zhang and Rui Yang and Boxiang Lyu and Tong Zhang},
  journal={arXiv preprint arXiv:2407.17466},
  year={2024}
}
\bibliographystyle{iclr2026_conference}

\appendix
\newpage
\appendix
\onecolumn

\section*{Appendices}
\addcontentsline{toc}{section}{Appendices}

\startcontents[appendix]
\printcontents[appendix]{l}{1}{\setcounter{tocdepth}{2}}

\pchreb{\section{Additional Background: Successor Measure and Forward-Backward Representations}
\label{app:FB}
\subsection{Successor Measure}
Recall that in standard single-objective RL, the Q-function under a policy $\pi$ with respect to a reward function ${R}:\mathcal{S}\times \mathcal{A}\rightarrow {\mathbb{R}}$ is defined as 
\begin{equation}
    Q^\pi(s, a):= \mathbb{E}_{\pi} \left[ \sum_{t=0}^{\infty} \gamma^t {R}(s_t, a_t) \mid s_0=s, a_0=a \right],\label{eq:Q function}
\end{equation}
which captures the long-term expected discounted reward under the policy $\pi$. 
One way to interpret the Q function is through the lens of \textit{successor measure} $\mathcal{M}^{\pi}: \mathcal{S}\times \mathcal{A}\rightarrow \Delta(\mathcal{S}\times \mathcal{A})$, which reflects the discounted, expected future occupancy of the state-action pairs in $X$ when starting from $(s, a)$ and following policy $\pi$. Formally, the successor measure $\mathcal{M}^{\pi}$ is defined as follows: For any subset $X\subset \mathcal{S}\times\mathcal{A}$, define
\begin{equation}
    \mathcal{M}^{\pi}(s, a, X):=\mathbb{E}_{\pi} \left[ \sum_{t=0}^{\infty} \gamma^t \cdot \mathbb{I}\{(s_t,a_t)\in X\} \Big\rvert s_0=s, a_0=a \right],\label{eq:successor measure}
\end{equation}
where $\mathbb{I}\{\cdot\}$ is the indicator function. 
The key property is that this measure $\mathcal{M}^{\pi}$ is agnostic to the reward function ${R}$ as it only depends on the environment dynamics and the policy $\pi$. 
Note that viewing $\mathcal{M}^{\pi}$ as a measure can deal with both the discrete and continuous cases.
For ease of exposition, in the sequel, we focus on the case of discrete state and action spaces.
With that said, we also let $\mathcal{M}^{\pi}(s, a, s',a')$ denote the successor measure of the state-action pair $(s',a')$ when starting from $(s, a)$ and following policy $\pi$.
We can connect the Q function in~\Cref{eq:Q function} with the successor measure in~\Cref{eq:successor measure} as follows:
\begin{equation}
    Q^{\pi}(s, a) = \sum_{(s', a')\in \mathcal{S}\times\mathcal{A}} \mathcal{M}^{\pi}(s, a, s', a') \cdot {R}(s', a').
\end{equation}
Crucially, we can make two observations: (i) If we can directly learn the successor measure of a policy $\pi$, then we can derive the corresponding Q function in a zero-shot manner given any reward function. (ii) For any reward function $R$, if we can learn the successor measure of an optimal policy for $R$ (denoted by $\pi_{R}^*$), then we can learn the optimal Q function $Q^{\pi_{R}^*}$ and thereafter derive an optimal policy for $R$. Therefore, if we can directly encode the information about $\pi_R^{*}$ in the successor measure, then we can learn the optimal Q function for any possible reward function.
}
%where $\mathcal{M}^{\pi}(s, a)$ is the Successor Feature vector corresponding to state-action pair $(s, a)$. 
%If the reward function ${R}$ can be expressed as a linear combination of features, the Q-function can be re-expressed as a linear dot product:
%\begin{equation}
%    Q^{\pi}(s, a) = \sum_{s', a'} \mathcal{M}^{\pi}(s, a, s', a') \cdot \mathbf{R}(s', a') = \mathcal{M}^{\pi}(s, a)^\top \boldsymbol{\lambda},
%\end{equation}
%This enables zero-shot adaptation to any new preference vector $\boldsymbol{\lambda}$ by simply taking the dot product, without further environment interaction.

\pchreb{\subsection{Forward-Backward (FB) Representations}
Our MORL-FB method is built on the Forward-Backward (FB) representation approach~\citep{touati2021learning,ahmed2023zero}, which provides a compact, low-rank factorization of the successor measure. The FB framework decomposes the successor measure into the product of two vector-valued functions, namely a Forward function $\mathbf{F}:\mathcal{S}\times\mathcal{A}\times\mathbb{R}^{d_z}\rightarrow\mathbb{R}^{d_z} $ and a Backward function $\mathbf{B}:\mathcal{S}\times\mathcal{A}\rightarrow \mathbb{R}^{d_z}$.
Let $\pi_z$ be a policy that depends on some vector $z\in\mathbb{R}^{d_z}$.
Specifically, under the FB framework, the successor measure $\mathcal{M}^{\pi}$ is represented by:
\begin{equation}
    \mathcal{M}^{\pi_z}(s, a, s', a') =\mathbf{F}(s, a, z)^\top \mathbf{B}(s', a'),
\end{equation}
where ${z}$ acts as some latent vector that captures the required information about the policy $\pi_z$. Using this factorization, the Q-function for any policy $\pi_z$ can be expressed as
\begin{align}
    Q^{\pi_z}(s, a) &=\sum_{(s', a')\in \mathcal{S}\times\mathcal{A}}  \mathbf{F}(s, a, z)^\top \mathbf{B}(s', a') {R}(s', a'),\label{eq:Q FB 1}\\
    &=  \mathbf{F}(s, a, z)^\top \underbrace{\left(\sum_{(s', a')\in \mathcal{S}\times\mathcal{A}}\mathbf{B}(s', a') {R}(s', a') \right)}_{=:z_R}.\label{eq:Q FB 2}
\end{align}
%Here, the preference embedding $z_R$ is calculated as:
%\begin{equation}
%    z_R = \sum_{s',a'} \mathbf{B}_{\omega}(s',a')\, R(s',a').
%\end{equation}
Note that~\Cref{eq:Q FB 2} holds for \textit{any} $z$ (and hence any policy $\pi_z$). Therefore, we have the freedom to set $z_R$ to be the latent vector that results in an optimal policy for the reward function $R$, \ie $\pi_{z_R}\equiv \pi_R^{*}$. 
To achieve this, we shall simply enforce the following Bellman optimality equation~\citep{touati2021learning}, \ie
\begin{align}
\pi^{*}_R(s)=\arg\max_{a\in\mathcal{A}} \; \mathbf{F}(s, a, z_R)^{\top} z_R.\label{eq:FB optimal}
\end{align}
The final optimal policy for the given preference is then directly constructed by maximizing the approximated Q-function:
%\begin{align}
%\pi^{*}(s, \boldsymbol{\lambda}) &\approx \arg\max_{a} Q^{*}(s, a, \boldsymbol{\lambda}) \\
 %                               &=\arg\max_{a} \; F(s, a, z_R)^{\top} z_R .
%\end{align}
Therefore, if $\mathbf{F}$ and $\mathbf{B}$ are well learned, then one can directly retrieve an optimal policy for any reward function $R$ given at test time by computing $z_R$ and apply~\Cref{eq:FB optimal}. This factorization provides the foundation of our MORL-FB algorithm to efficiently explore and generalize across preferences during training.
}

\section{Detailed Pseudo Code of MORL-FB}
\label{app:algorithm}
\begin{algorithm}[h!]
    \caption{MORL-FB} 
	\begin{algorithmic}[1]
 
        \State \textbf{Input:} Network parameters $\theta, \bar{\theta}, \omega, \bar{\omega}, \eta, \bar{\eta}$, preference sampling distribution $\mathcal{P}_{\blambda}$, preference set $\Lambda$, actor learning rates $\mu_\pi$, FB presentation learning rate $\mu_{\text{FB}}$, $\bz$ dimension $d_z$, sample number $n_s$, update frequency $n_u$, warm up steps $n_w$, and target smoothing coefficient $\tau$
        \State Initialize networks $\mathbf{F}_{\theta}, \mathbf{B}_{\omega}, \pi_{\eta}$ and target networks $\mathbf{F}_{\bar{\theta}}, \mathbf{B}_{\bar{\omega}}, \pi_{\bar{\eta}}$
        \State Initialize replay buffer $\mathcal{M} \leftarrow \varnothing$
		\For {$\text{each} \ \text{iteration} \ i$}
            \State Sample a preference vector $\boldsymbol{\blambda}\sim \mathcal{P}_{\boldsymbol{\blambda}}$
            \If{$i \leq n_{w}$}
                \Comment Warm-up stage
                \State Sample $\bz$ from a multivariate normal distribution $\mathcal{N}(\boldsymbol{0},\mathbb{I}^{d_z})$
                \State Normalize $\bz$ such that $\boldsymbol{z} \leftarrow \sqrt{d_z}\frac{\boldsymbol{z}}{\Vert \boldsymbol{z} \Vert_{2}}$
                % \State $ \bz \leftarrow\sqrt{z_d}\bz$
            \Else
                \State $\bz \leftarrow \text{PG-Explore}(\boldsymbol{\blambda})$
            \EndIf
            \For {$\text{each} \ \text{environment} \ \text{step} \ t$}
                \State $a_{t} \sim \pi_{\eta}(\cdot \vert s_{t}; \boldsymbol{\bm{\lambda}})$
                \State $s_{t+1} \sim \mathcal{P}(\cdot \vert s_{t}, a_{t})$
                \State $\mathcal{M} \leftarrow \mathcal{M} \bigcup \ \{(s_{t}, a_{t}, \br_{t}, \boldsymbol{\blambda}_{t}, s_{t+1})\}$
                %\State Sample $N_{w}$ preference $\blambda'$
                % \For {$each \ additional \ numbers \ of \ preference \ N_{w}$}
                %    \State $\mathcal{M} \leftarrow \mathcal{M} \bigcup \ \{(s_{t}, a_{t}, r_{t}, \blambda', s_{t+1})\}$
                % \EndFor
            \EndFor
            \For {$\text{each} \ \text{gradient} \ \text{step} \ j$}
                \State Sample a batch of transitions $\{(s, a, \boldsymbol{r}, \boldsymbol{\blambda}, s')\}$ from the replay buffer $\mathcal{M}$
                % \State Sample preference $\blambda$ from $\mathcal{P}_{\blambda}$
                \State $\bz_{j} \leftarrow \text{PG-Explore}(\boldsymbol{\blambda})$
                \State $\theta \leftarrow \theta - \mu_{\text{FB}} \nabla_{\theta} (\mathcal{L}_{Q}(\theta;\boldsymbol{\blambda}) +  \mathcal{L}_{\text{M}}(\theta,\omega;\boldsymbol{\blambda}))$\;
                \State $\omega \leftarrow \omega - \mu_{\text{FB}} \nabla_{\omega} (\mathcal{L}_{n}(\omega;\boldsymbol{\blambda}) + \mathcal{L}_{\text{M}}(\theta,\omega;\boldsymbol{\blambda}))$\;
            \EndFor
            \If{$i \ \% \ n_u == 0$}
                     \State $\eta \leftarrow \eta - \mu_{\pi} \nabla_{\eta} \mathcal{L}_{\pi}(\eta;\boldsymbol{\blambda})$\;
                     \State $\bar{\theta} \leftarrow \tau \theta + (1 - \tau) \bar{\theta}$
                     \State $\bar{\omega} \leftarrow \tau \omega + (1 - \tau) \bar{\omega}$
                     \State $\bar{\eta} \leftarrow \tau \eta + (1 - \tau) \bar{\eta}$
                \EndIf
		\EndFor
    
        \Function{$\text{PG-Explore}$}{$\boldsymbol{\blambda}$}
            \State Sample a batch $\mathcal{D}$ of $n_s$ non-terminal transitions $\{(s, a, \boldsymbol{r}, s')\}$ from $\mathcal{M}$
            \State $\bz \leftarrow \sum_{(s,a,\boldsymbol{r},s')\in \mathcal{D}}\frac{\mathbf{B}_{\omega}(s,a)\boldsymbol{r}^{\top}\boldsymbol{\blambda}}{n_s}$
             % \State Normalize $\bz$ such that $\lVert\bz\rVert_2=\bz_d$ 
            \State Normalize $\bz$ such that $\boldsymbol{z} \leftarrow \sqrt{d_z}\frac{\boldsymbol{z}}{\Vert \boldsymbol{z} \Vert_{2}}$
            \State \Return $\bz$
        \EndFunction
  
	\end{algorithmic}
    \label{alg:MORL-FB}
\end{algorithm}
%\bsw{(from reviewer3 Q1: move the descriptions of FB in Appendix A to the main text) (from reviewer4 Q9: In Algorithm 2 in Appendix A, z computed in line 10 is not used. Reply: In Line 14, $\pi$ takes the latent vector $z$ (derived from the sampled preference 
%) as input.)}
\Cref{alg:MORL-FB} details the proposed MORL-FB method. Initially, during the warm-up phase (lines 6-8), the latent vector $\boldsymbol{z}$ is sampled from a standard multivariate normal distribution.  After the warm-up, $\boldsymbol{z}$ is determined using the preference-guided sampling scheme (line 10). This $\boldsymbol{z}$ is then used to generate trajectories within the environment, which are stored in the replay buffer $\mathcal{M}$ (lines 12-16). Model updates are performed by sampling transitions from $\mathcal{M}$ (lines 17-22). A delayed actor update mechanism is employed for the actor model (lines 23-24), and target networks are updated via a soft update scheme (lines 25-27). The Preference-Guided Exploration function (lines 30-35) normalizes the sampled latent vector $\boldsymbol{z}$ (line 33) as $\boldsymbol{z} \leftarrow \sqrt{d_z}\frac{\boldsymbol{z}}{\Vert \boldsymbol{z} \Vert_{2}}$. This normalization step, motivated by the prior work~\citep{ahmed2023zero}, has been observed to improve performance.

As shown in~\Cref{alg:MORL-FB}, the training of MORL-FB involves the following loss functions:

\textbf{Measure Loss.} The Measure loss, $\mathcal{L}_{\text{M}}(\theta, \omega; \boldsymbol{z_{\lambda}})$, is central to learning a task-agnostic representation of environment dynamics, $\mathbf{F}_{\theta}(s_{t}, a_{t}, \boldsymbol{z}_{\boldsymbol{\lambda}})$, encoding command-conditioned successor measures. It enforces Bellman consistency for these measures when projected onto a learned basis $\mathbf{B}_{\omega}(s',a')$, as shown in~\Cref{eq:measure_loss}. This mechanism, drawn from~\citep{ahmed2023zero}, aims to separate the environment structure from specific rewards. This disentanglement is crucial for enabling zero-shot generalization, allowing the agent to understand ``what happens next" irrespective of the immediate goal, forming a reusable foundation for various tasks.
\begin{comment}
\begin{align}
\label{eq:measure_loss}
\mathcal{L}_{\text{M}}(\theta, \omega; \boldsymbol{z_{\lambda}}) &= 
\mathbb{E}_{\substack{(s_{t}, a_{t}, s_{t+1}) \sim \mathcal{D} \\ s' \sim \mathcal{D}}} \left[ 
\left( 
\mathbf{F}_{\theta}(s_{t}, a_{t}, \boldsymbol{z}_{\boldsymbol{\lambda}})^{\top} 
\mathbf{B}_{\omega}(s') 
- \gamma 
\mathbf{F}_{\bar{\theta}}(s_{t+1}, \pi_{\bar{\eta}}(s_{t+1}), \boldsymbol{z}_{\boldsymbol{\lambda}})^{\top} 
\mathbf{B}_{\bar{\omega}}(s') 
\right)^{2} 
\right] \notag \\
&\quad 
- 2 \, \mathbb{E}_{(s_{t}, a_{t}, s_{t+1}) \sim \mathcal{D}} \left[ 
\mathbf{F}_{\theta}(s_{t}, a_{t}, \boldsymbol{z}_{\boldsymbol{\lambda}})^{\top} 
\mathbf{B}_{\omega}(s_{t+1}) 
\right],
\end{align}  
\end{comment}

\begin{align}
\mathcal{L}_{\text{M}}(\mathbf{F_{\theta}},\mathbf{B_{\omega}}; \boldsymbol{z_{\blambda}}) = &\mathbb{E}_{\substack{(s_{t}, a_{t}, s_{t+1}) \sim \mathcal{D}\\ (s',a') \sim \mathcal{D}}} \big[ ( \mathbf{F_{\theta}}(s_{t}, a_{t}, \boldsymbol{z_{\blambda}})^{\top} \mathbf{B_{\omega}}(s',a') \nonumber\\
& - \gamma \mathbf{F_{\bar{\theta}}}(s_{t+1}, \pi(s_{t+1}, \boldsymbol{z_{\blambda}}), \boldsymbol{z_{\blambda}})^{\top} \mathbf{B_{\bar{\omega}}}(s',a') )^{2} \big] \nonumber\\
& - 2 \ \mathbb{E}_{(s_{t}, a_{t}, s_{t+1}) \sim \mathcal{D}} [ \mathbf{F_{\theta}}(s_{t}, a_{t}, \boldsymbol{z_{\blambda}})^{\top} \mathbf{B_{\omega}}(s_{t+1},a_{t+1})].
\label{eq:measure_loss}
\end{align}

where $\rho$ denotes the underlying distributions of the dataset.

\textbf{Auxiliary Q Loss.} To ensure the learned representation $\mathbf{F}_{\theta}$ is relevant for decision-making, the Auxiliary Q Loss, $\mathcal{L}_{Q}(\theta; \boldsymbol{z_{\blambda}})$, connects it to task-specific values. When explicit reward signals $\boldsymbol{r}_{t}$ and corresponding preferences $\blambda$ are available, \Cref{eq:q_loss} minimizes a standard temporal difference error. This is vital for MORL contexts, effectively teaching $\mathbf{F}_{\theta}$ to support optimizing diverse rewards.
\begin{align}
\label{eq:q_loss}
\mathcal{L}_{Q}(\theta; \boldsymbol{z_{\blambda}}) &= \mathbb{E}_{(s_{t}, a_{t}, \boldsymbol{r}_{t}, s_{t+1}) \sim \mathcal{D}} \left[ \left( \mathbf{F}_{\theta}(s_{t}, a_{t}, \boldsymbol{z_\blambda})^{\top} \boldsymbol{z}_{\boldsymbol{\blambda}} - \left( \boldsymbol{\blambda}^{\top} \boldsymbol{r}_{t} + \gamma \mathbf{F}_{\bar{\theta}}(s_{t+1}, \pi_{\bar{\eta}}(s_{t+1}),  \boldsymbol{z_\blambda})^{\top} \boldsymbol{z}_{\boldsymbol{\blambda}} \right) \right)^{2} \right].
\end{align}

\textbf{Orthonormality Loss.} The Orthonormality Regularization Loss, $\mathcal{L}_{\text{n}}(\omega)$, acts as a crucial regularizer for the learned basis functions $\mathbf{B}_{\omega}(s,a)$. Its purpose, as reflected in \Cref{eq:ortho_loss}, is to promote a well-conditioned and non-degenerate basis. By encouraging properties such as orthogonality between basis vectors and unit norm, this loss helps prevent representational collapse and redundancy within $\mathbf{B}_{\omega}$. This, in turn, ensures that the successor measures are projected onto a stable and diverse set of features, enhancing the robustness and quality of the learned representations $\mathbf{F}_{\theta}$.

\begin{align}
\label{eq:ortho_loss}
\mathcal{L}_{\text{n}}(\omega) &= \mathbb{E}_{(s,a) \sim \mathcal{D}, (s',a') \sim \mathcal{D}} \left[ \left( \mathbf{B}_{\omega}(s,a)^{\top} \mathbf{B}_{\omega}(s',a') \right)^{2} - \| \mathbf{B}_{\omega}(s,a) \|_{2}^{2} - \| \mathbf{B}_{\omega}(s',a') \|_{2}^{2} \right].
\end{align}

\textbf{Policy Loss.} The agent's behavior is refined through the Policy Optimization Loss, $\mathcal{L}_{\pi}(\eta; \boldsymbol{z_{\blambda}})$, which trains the policy $\pi_{\eta}$ within an actor-critic paradigm. The actor's objective is to maximize the Q-values estimated by the critic, where these Q-values are derived from the learned representation as $Q(s, a; \boldsymbol{z_{\blambda}}) = \mathbf{F}_{\theta}(s, a, \boldsymbol{z_{\blambda}})^{\top} \boldsymbol{z}_{\boldsymbol{\blambda}}$ (\Cref{eq:policy_loss}). This loss drives the policy to select actions that are optimal for the task specified by the current command $\boldsymbol{z_{\blambda}}$. It thus enables the agent to translate its universal understanding of the environment into effective, task-adaptive behavior.

\begin{align}
\label{eq:policy_loss}
\mathcal{L}_{\pi}(\eta; \boldsymbol{z_{\blambda}}) &= \mathbb{E}_{s \sim \mathcal{D}} \left[ -Q(s, \pi_{\eta}(s); \boldsymbol{z_{\blambda}}) \right], \quad \text{where } Q(s, a; \boldsymbol{z_{\blambda}}) = \mathbf{F}(s, a, \boldsymbol{z_{\blambda}})^{\top} \boldsymbol{z}_{\boldsymbol{\blambda}}.
\end{align}

\section{Detailed Configurations of Experiments}
\label{app:experiment_config}

In this section, we describe the experimental setup used to evaluate the performance of our approach. We detail the hyperparameters used in our experiments, as well as the reference points chosen for HV evaluation across different environments. 

\subsection{Evaluation Environments}

\iclrwh{We evaluate the performance of our proposed method, MORL-FB, across a diverse set of multi-objective reinforcement learning environments. These environments, detailed below, encompass both established MuJoCo-based locomotion tasks and discrete problems, allowing us to assess its adaptability across distinct settings with varying state spaces, action spaces, and objective numbers.}

\iclrwh{\textbf{MuJoCo-Based Continuous Control:}}
\begin{itemize}
    \item \textbf{Halfcheetah2d:} The state space and action space are defined as $\mathcal{S} \subseteq \mathbb{R}^{17}$ and $\mathcal{A} \subseteq \mathbb{R}^{6}$, respectively. The two objectives for this environment are maximizing moving speed along the x-axis and minimizing energy cost.
    \item \textbf{Walker2d:} The state space and action space are defined as $\mathcal{S} \subseteq \mathbb{R}^{17}$ and $\mathcal{A} \subseteq \mathbb{R}^{6}$, respectively. The two objectives for this environment are maximizing moving speed along the x-axis and minimizing energy cost.
    \item \textbf{Hopper3d:} The state space and action space are defined as $\mathcal{S} \subseteq \mathbb{R}^{11}$ and $\mathcal{A} \subseteq \mathbb{R}^{3}$. The three objectives include maximizing moving speed along the x-axis, maximizing jumping height along the z-axis, and minimizing energy cost.
    \item \textbf{Ant3d:} The state space and action space are defined as $\mathcal{S} \subseteq \mathbb{R}^{27}$ and $\mathcal{A} \subseteq \mathbb{R}^{8}$. The three objectives are maximizing moving speed along the x-axis, maximizing moving speed along the y-axis, and minimizing energy cost.

    \item \textbf{Humanoid2d:} The state space and action space are defined as $\mathcal{S} \subseteq \mathbb{R}^{376}$ and $\mathcal{A} \subseteq \mathbb{R}^{17}$. The two objectives are maximizing moving speed along the x-axis and minimizing energy cost. Additionally, we set the healthy reward parameter to 1.0 to encourage exploration and stability.

    \item \textbf{Humanoid5d:} The state space and action space are defined as $\mathcal{S} \subseteq \mathbb{R}^{376}$ and $\mathcal{A} \subseteq \mathbb{R}^{17}$. This environment has five objectives: maximizing moving speed along the x-axis, maximizing moving speed along the y-axis, maximizing angular velocity on the left elbow, maximizing angular velocity on the right elbow, and minimizing energy cost. Similar to Humanoid2d, the healthy reward parameter is set to 1.0 to ensure meaningful evaluation.
\end{itemize}
\iclrwh{\textbf{Classic Control:}
\begin{itemize}
    \item \textbf{Deep Sea Treasure (DST):} The Deep Sea Treasure environment is a classic MORL problem in which the agent controls a submarine in a 2D grid world~\citep{vamplew2011empirical}. The two conflicting objectives are typically maximizing collected treasure value and minimizing the time cost of collection.
    \item \textbf{Fruit Tree Navigation (FTN):} This environment is structured as a full binary tree of depth $d=5, 6,$ or $7$~\citep{yang2019general}. The agent navigates from the root to a leaf node. Every leaf contains a fruit with values for six objectives: Protein, Carbs, Fats, Vitamins, Minerals, and Water.
\end{itemize}}

\subsection{Experimental Setup}
\label{sub:exp_setup}
To begin with, we describe the hyperparameters of the benchmark MORL methods and the proposed MORL-FB for better reproducibility.

\textbf{Hyperparameters for Experiments.} 
To ensure a fair comparison, for those benchmark methods that already provide tuned task-specific hyperparameters on MuJoCo, we primarily refer to their original papers for the hyperparameter configurations, including PGMORL~\citep{xu2020prediction} and CAPQL~\citep{lu2023multi}. \Cref{tab:hyperparam_pgmorl} and \Cref{tab:hyperparam_capql} list the detailed hyperparameters used in our experiments. For PGMORL, the hyperparameters reflect its evolutionary population-based design. The parameter $n$ defines the number of parallel reinforcement learning tasks in each generation. Each task includes $m_w$ warm-up iterations and $m_t$ evolutionary iterations. $P_{\text{num}}$ and $P_{\text{size}}$ define the number and size of the performance buffers. The PPO parameters used across all environments are summarized in \Cref{tab:ppo_param}. For CAPQL, the hyperparameter $\alpha$ controls the strength of a concave regularization term added to the reward. The general hyperparameters shared by CAPQL are listed in \Cref{tab:capql_shared}.

\begin{table}[h!]
    \centering
    \caption{Hyperparameters of PGMORL.}
    \label{tab:hyperparam_pgmorl}
    \begin{tabular}{lccccccc}
        \toprule
        \textbf{Environments}  & $\boldsymbol{n}$ & $\boldsymbol{m_w}$ & $\boldsymbol{m_t}$ & $\boldsymbol{P_{\text{num}}}$ & K& $\boldsymbol{P_{\text{size}}}$ & $\boldsymbol{\alpha}$ \\ \midrule
        HalfCheetah2d & $6$ & $80$  & $20$  & $100$ & $2$ & $7$ & $-1$ \\
        Walker2d    & $6$ & $80$  & $20$  & $100$ & $2$ & $7$ & $-1$ \\
        Hopper3d      & $15$ & $200$ & $40$  & $210$ & $2$ & $7$ & $-10^6$ \\ 
        Ant3d         & $15$ & $200$ & $40$  & $210$ & $2$ & $7$ & $-10^6$ \\
        Humanoid2d    & $6$ & $200$ & $40$  & $100$ & $2$ & $7$ & $-1$ \\
        Humanoid5d    & $35$ & $200$ & $40$  & $550$ & $2$ & $7$ & $-10^6$ \\
        \bottomrule
    \end{tabular}
\end{table}

\begin{table}[h!]
    \centering
    \caption{PPO hyperparameters used in PGMORL.}
    \label{tab:ppo_param}
    \begin{tabular}{ll}
        \toprule
        \textbf{Parameter} & \textbf{Value} \\
        \midrule
        Timesteps per actor batch & 2,048 \\
        Processes number & 4 \\
        Learning rate & $3 \times 10^{-4}$ \\
        Discount factor ($\gamma$) & 0.995 \\
        GAE lambda & 0.95 \\
        Batch size & 32 \\
        PPO epochs & 10 \\
        Entropy coefficient & 0 \\
        Value loss coefficient & 0.5 \\
        \bottomrule
    \end{tabular}
\end{table}

\begin{table}[h!]
    \centering
    \caption{Augmentation strength of CAPQL.}
    \label{tab:hyperparam_capql}
    \begin{tabular}{ll}
        \toprule
        \textbf{Environments} & $\boldsymbol{\alpha}$ \\ \midrule
        HalfCheetah2d & 0.1 \\ 
        Walker2d & 0.05 \\
        Hopper3d & 0.2 \\ 
        Ant3d & 0.2 \\ 
        Humanoid2d & 0.005 \\ 
        Humanoid5d & 0.005 \\ 
        \bottomrule
    \end{tabular}
\end{table}

\begin{table}[h!]
    \centering
    \caption{Hyperparameters of CAPQL.}
    \label{tab:capql_shared}
    \begin{tabular}{ll}
        \toprule
        \textbf{Parameter} & \textbf{Value} \\
        \midrule
        Optimizer & Adam \\
        Learning rate & $3 \times 10^{-4}$ \\
        Discount factor ($\gamma$) & 0.99 \\
        Number of hidden units per layer & 256 \\
        Replay buffer size & $10^6$ \\
        Batch size & 256 \\
        Nonlinearity & ReLU \\
        Target smoothing coefficient ($\tau$) & 0.005 \\
        \bottomrule
    \end{tabular}
\end{table}

For algorithms without officially tuned or specified hyperparameters, we perform hyperparameter optimization (HPO) by following~\citep{eimer2023hyperparametersreinforcementlearningtune}. Specifically, we employ Bayesian optimization by using the Weights \& Biases Sweeps. Each episode refers to a simulation run for 1M environment steps, and the optimization is performed over 10 episodes. \Cref{tab:fb_tuning,tab:qp_tuning,tab:pdmorl_tuning,tab:pcn_tuning,tab:morld_tuning,tab:sfols_tuning,tab:gpils_tuning} summarize the search ranges and the final selected hyperparameters.

\begin{itemize}
    \item \textbf{MORL-FB:} For MORL-FB, which is built on the implementation of the Twin Delayed Deep Deterministic Policy Gradient (TD3) algorithm \citep{fujimoto2018addressingfunctionapproximationerror}, we prioritized the tuning of parameters inherent to TD3. Specifically, this included critical elements such as learning rates for actor and critic networks, the target network update rate, and policy delay, which are known to significantly influence the stability and performance of TD3-based agents.
    \item \textbf{Q-Pensieve:} Our tuning efforts centered on parameters associated with its core Q-Snapshot mechanism~\citep{weihung2023qp}. Given that Q-Pensieve's efficacy in improving sample efficiency stems from the storage and utilization of these Q-function snapshots, parameters governing the snapshot buffer, the frequency of snapshot capture, and their influence on the policy update were key areas of focus during hyperparameter optimization.
    \item \textbf{PD-MORL:} For PD-MORL, which can be viewed as the multi-objective extensdion of TD3~\citep{fujimoto2018addressingfunctionapproximationerror}, we prioritized the tuning of parameters inherent to TD3. This included critical elements such as learning rates for actor and critic networks, the target network update rate, and policy delay, which are known to significantly influence the stability and performance of TD3-based agents.
    \item \textbf{MORL/D:} For MORL/D, which is built upon SAC, we tuned the standard hyperparameters of SAC. In addition, we focused on four components that are central to the MORL/D framework: population size, neighborhood size, scalarization method, and weight adaptation method. These elements directly affect how the algorithm decomposes the multi-objective space and maintains policy diversity.
    \item \textbf{PCN:} For PCN, since the hyperparameter settings for continuous control tasks such as MuJoCo environments were not specified in~\citep{reymond2022pareto}, we tuned three critical parameters: learning rate, batch size, and hidden dimension, which affect training stability, sample efficiency, and the model’s ability to generalize across diverse preference vectors.
    \item \textbf{SFOLS:} For SFOLS, since we utilize its official implementation for discrete problems and extend it with a TD3 backbone for evaluation on continuous control tasks, we tuned the parameters inherent to TD3, including learning rates for the actor and critic networks, policy noise, and target policy smoothing noise. These components are known to significantly influence the stability and performance of TD3-based agents.
    {\item \textbf{GPI-PD:} For GPI-PD, since GPI-PD relies on a learned dynamics model for sample generation, we tuned four key parameters: dynamics rollout length, rollout frequency, model training frequency, and real data ratio. These factors critically affect model accuracy, stability of planning updates, and the overall effectiveness of Dyna-style training.}
\end{itemize}

\begin{table}[h!]
    \centering
    \caption{Hyperparameter tuning for MORL-FB.}
    \label{tab:fb_tuning}
    \begin{tabular}{lll}
        \toprule
        \textbf{Parameter} & \textbf{Value} &\textbf{Optimal Value} \\ \midrule
        Learning rate & $[0.0001, 0.01]$ & $0.0001$ \\ \midrule
        Policy update delay & $\{1, 2, 5, 10\}$ & $10$ \\ \midrule
        Steps per update & $\{1, 2, 5, 10\}$ & $1$ \\ \midrule
        Latent dimension ($\boldsymbol{z}$ dimension) & $\{50 ,150, 300\}$ & $300$ \\ \midrule
        Exploration noise std. & $[0.1, 1.0]$ & $0.1$ \\ \midrule
        Target policy smoothing noise std. & $[0.1, 1.0]$ & $0.2$ \\
        \bottomrule
    \end{tabular}
\end{table}

\begin{table}[h!]
    \centering
    \caption{Hyperparameter tuning for Q-Pensieve.}
    \label{tab:qp_tuning}
    \begin{tabular}{lll}
        \toprule
        \textbf{Parameter} & \textbf{Value} & \textbf{Optimal Value} \\ \midrule
        Learning rate & $[0.0001, 0.01]$ & $0.0001$ \\ \midrule
        Q replay buffer size & $\{1, 2, 4\}$ & $4$ \\ \midrule
        Preference set size & $\{1, 2, 4\}$ & $4$ \\
        \bottomrule
    \end{tabular}
\end{table}

\begin{table}[h!]
    \centering
    \caption{Hyperparameter tuning for PD-MORL.}
    \label{tab:pdmorl_tuning}
    \begin{tabular}{lll}
        \toprule
        \textbf{Parameter} & \textbf{Value} & \textbf{Optimal Value} \\ \midrule
        Learning rate & $[0.0001, 0.01]$ & $0.0003$ \\ \midrule
        Exploration noise std. & $[0.1, 1.0]$ & $0.15$ \\ \midrule 
        Target policy smoothing noise std. & $[0.1, 1.0]$ & $0.25$ \\ \midrule
        Policy update delay & $\{1, 2, 5, 10\}$ & $10$ \\
        \bottomrule
    \end{tabular}
\end{table}

\begin{table}[h!]
    \centering
    \caption{Hyperparameter tuning for PCN.}
    \label{tab:pcn_tuning}
    \begin{tabular}{lll}
        \toprule
        \textbf{Parameter} & \textbf{Value} & \textbf{Optimal Value} \\ \midrule
        Learning rate & $[0.0001, 0.01]$ & $0.0023$ \\ \midrule
        Batch size & $\{64, 128, 256\}$ & $64$ \\ \midrule
        Number of hidden units per layer & $\{256, 512, 1024\}$ & $1,024$ \\ \bottomrule
    \end{tabular}
\end{table}

\begin{table}[h!]
    \centering
    \caption{Hyperparameter tuning for MORL/D.}
    \label{tab:morld_tuning}
    \begin{tabular}{lll}
        \toprule
        \textbf{Parameter} & \textbf{Value} & \textbf{Optimal Value} \\ \midrule
        Learning rate & $[0.0001, 0.01]$ & $0.0013$ \\ \midrule
        Batch size & $\{64, 128, 256\}$ & $256$ \\ \midrule
        Number of hidden units per layer & $\{256, 512, 1024\}$ & $1,024$ \\ \midrule
        Pop size & $\{4, 6, 8\}$ & $6$ \\ \midrule
        Neighborhood size & $\{0, 1, 2\}$ & $1$ \\ \midrule
        Scalarization method & $\{none, ws\}$ & $ws$ \\ \midrule
        Weight adaptation method & $\{none, PSA\}$ & $PSA$ \\\bottomrule
    \end{tabular}
\end{table}

\begin{table}[h!]
    \centering
    \caption{Hyperparameter tuning for SFOLS.}
    \label{tab:sfols_tuning}
    \begin{tabular}{lll}
        \toprule
        \textbf{Parameter} & \textbf{Value} & \textbf{Optimal Value} \\ \midrule
        Learning rate & $[0.0001, 0.01]$ & $0.0006$ \\ \midrule
        Batch size & $\{64, 128, 256\}$ & $256$ \\ \midrule
        Number of hidden units per layer & $\{256, 512, 1024\}$ & $1,024$ \\ \midrule
        Target smoothing coefficient ($\tau$) & $[0.001, 0.02]$ & $0.0061$ \\ \midrule
        Exploration noise std. & $[0.1, 0.3]$ & $0.1736$ \\ \midrule
        Target policy smoothing noise std. & $[0.1, 0.5]$ & $0.4232$ \\ \bottomrule
    \end{tabular}
\end{table}

\begin{table}[h!]
    \centering
    \caption{Hyperparameter tuning for GPI-PD.}
    \label{tab:gpils_tuning}
    \begin{tabular}{lll}
        \toprule
        \textbf{Parameter} & \textbf{Value} & \textbf{Optimal Value} \\ \midrule
        Learning rate & $[0.0001, 0.01]$ & $0.0003$ \\ \midrule
        Batch size & $\{64, 128, 256\}$ & $256$ \\ \midrule
        Number of hidden units per layer & $\{256, 512, 1024\}$ & $1,024$ \\ \midrule
        Dynamics rollout length & $[5, 20]$ & $8$ \\ \midrule
        Dynamics rollout freqency & $[50, 500]$ & $312$ \\ \midrule
        Dynamics train freqency & $[100, 500]$ & $139$ \\ \midrule
        Dynamics real ratio & $[0.05, 0.2]$ & $0.0845$ \\ \bottomrule
    \end{tabular}
\end{table}

\textbf{Hyperparameters of MORL-FB.}
The hyperparameters for the MORL-FB experiments were chosen to ensure fair evaluation and stable learning, guided by prior research and the HPO results. \bswreb{The hyperparameters tuned via HPO are presented in Table \ref{tab:fb_tuning}, while the remaining values were primarily drawing from the default settings of original FB and TD3 algorithms.} The complete configuration is detailed in \Cref{tab:hyperparameters}.

\begin{table}[h!]
    \centering
    \caption{Hyperparameter configuration for MORL-FB experiments.}
    \label{tab:hyperparameters}
    \begin{tabular}{ll}
        \toprule
        \textbf{Parameter} & \textbf{Value} \\ \midrule
        Total number of environment steps & $3 \times 10^{6}$ \\
        Replay buffer size & $1 \times 10^{6}$ \\
        Latent dimension ($\boldsymbol{z}$ dimension) & 300 \\
        Interface batch size & 5,120 \\
        Number of hidden units per layer & 1,024 \\
        Preprocessing feature dimension & 512 \\
        Batch size & 1,024 \\
        Target smoothing coefficient ($\tau$) & 0.01 \\
        Discount factor ($\gamma$) & 0.99 \\
        Learning rate & $1 \times 10^{-4}$ \\
        Policy update delay & 10 \\
        Steps per update & 1 \\
        Exploration noise std. & 0.1 \\
        Target policy smoothing noise std. & 0.2 \\
        Clipping parameter & 0.5 \\
        Value loss coefficient & 1 \\ 
        \bottomrule
    \end{tabular}
\end{table}

\textbf{Reference Points for HV Evaluation.}
We compute the HV indicator using predefined reference points  (Ref. Point) for each environment. These reference points serve as baselines to measure the coverage of the Pareto front obtained during training. \Cref{tab:referencepoint} provides the specific reference points used in different environments.

\begin{table*}[h!]
    \centering
    \caption{Reference points used for HV calculation in different environments.}
    \label{tab:referencepoint}
    \begin{tabular}{@{}lc@{}}
        \toprule
        \textbf{Environment} & \textbf{Ref. Point} \\ \midrule
        Deep Sea Treasure & (0, -50) \\
        Fruit Tree Navigation & (0, 0, 0, 0, 0, 0) \\ 
        Halfcheetah2d & (0, -8000) \\
        Walker2d & (0, -8000) \\
        Hopper3d & (0, 0, -8000) \\
        Hopper4d & (0, 0, -8000, 0) \\
        Ant3d & (0, 0, -8000) \\
        Humanoid2d & (0, -8000) \\
        Humanoid5d & (0, 0, 0, 0, -8000)\\
        \bottomrule
    \end{tabular}
\end{table*}

\subsection{Compute Resources}
\label{app:sub:compute}

All models were trained on a workstation featuring a single NVIDIA RTX 4090 GPU, an Intel Core i7-13700K CPU, and 64 GB of system memory.

\newpage

\section{Additional Experimental Results}
\label{app:exp}
\subsection{Visualization of \texorpdfstring{$\boldsymbol{z}$}{z} Distribution in Different Environments}
\begin{figure}[h!]
    \centering
    \includegraphics[width=0.5\linewidth]{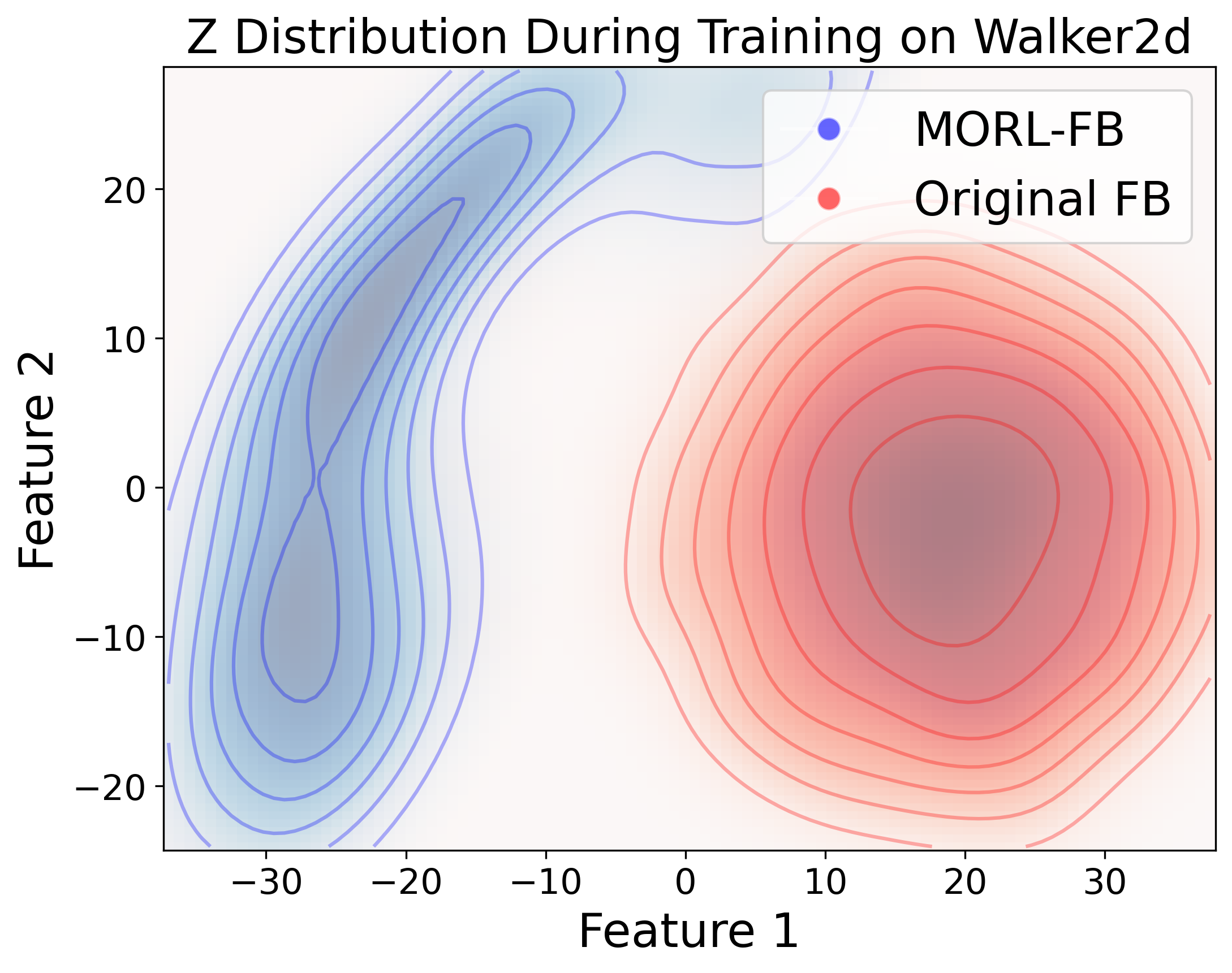}
    \caption{Empirical $\boldsymbol{z}$ distribution under MORL-FB with preference-guided sampling versus Original FB with simple normal distributions on Walker2d.}
    \label{fig:walker}
\end{figure}

\begin{figure}[h!]
    \centering
    \includegraphics[width=0.5\linewidth]{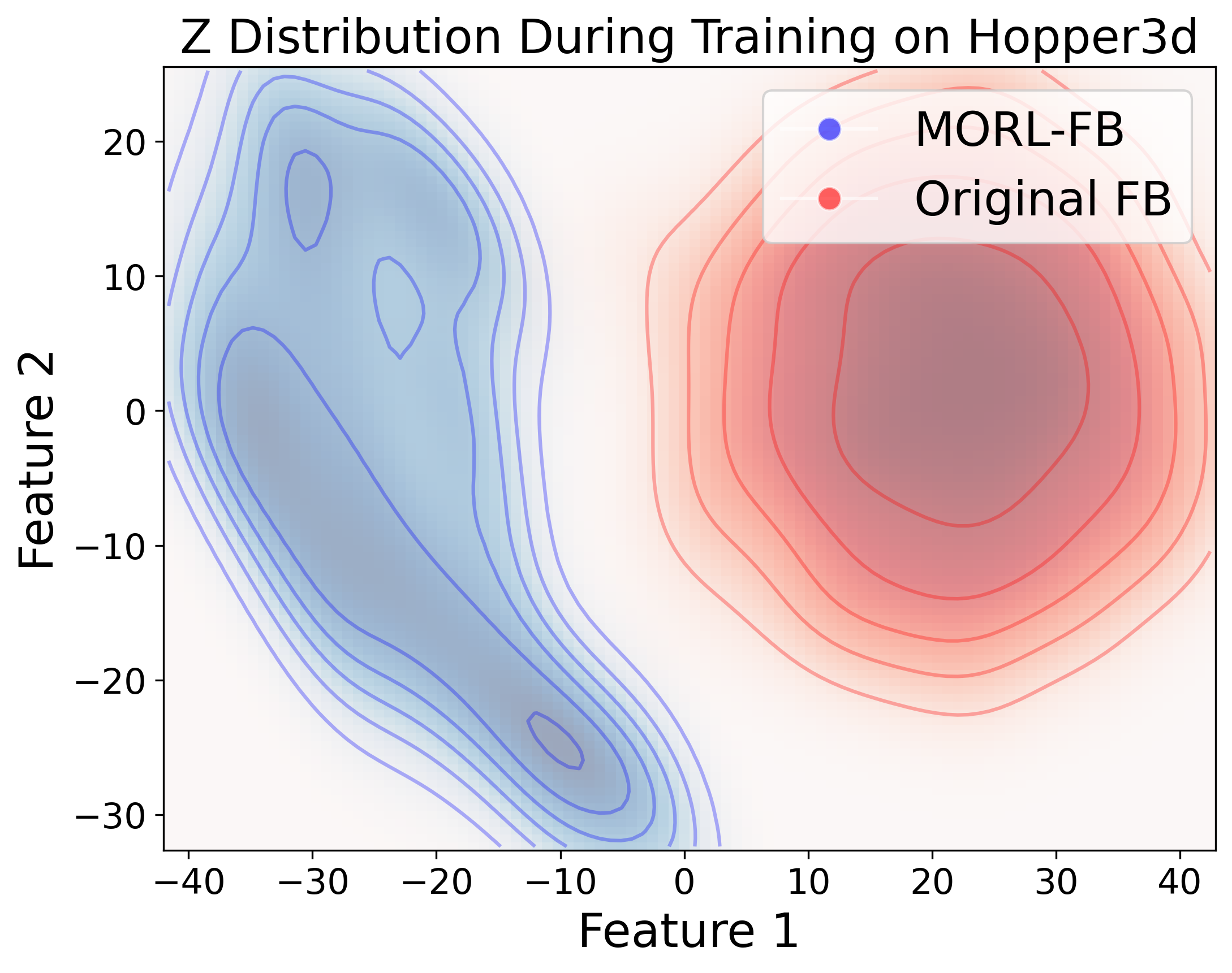}
    \caption{Empirical $\boldsymbol{z}$ distribution under MORL-FB with preference-guided sampling versus Original FB with simple normal distributions on Hopper3d.}
    \label{fig:hopper}
\end{figure}

\begin{figure}[h!]
    \centering
    \includegraphics[width=0.5\linewidth]{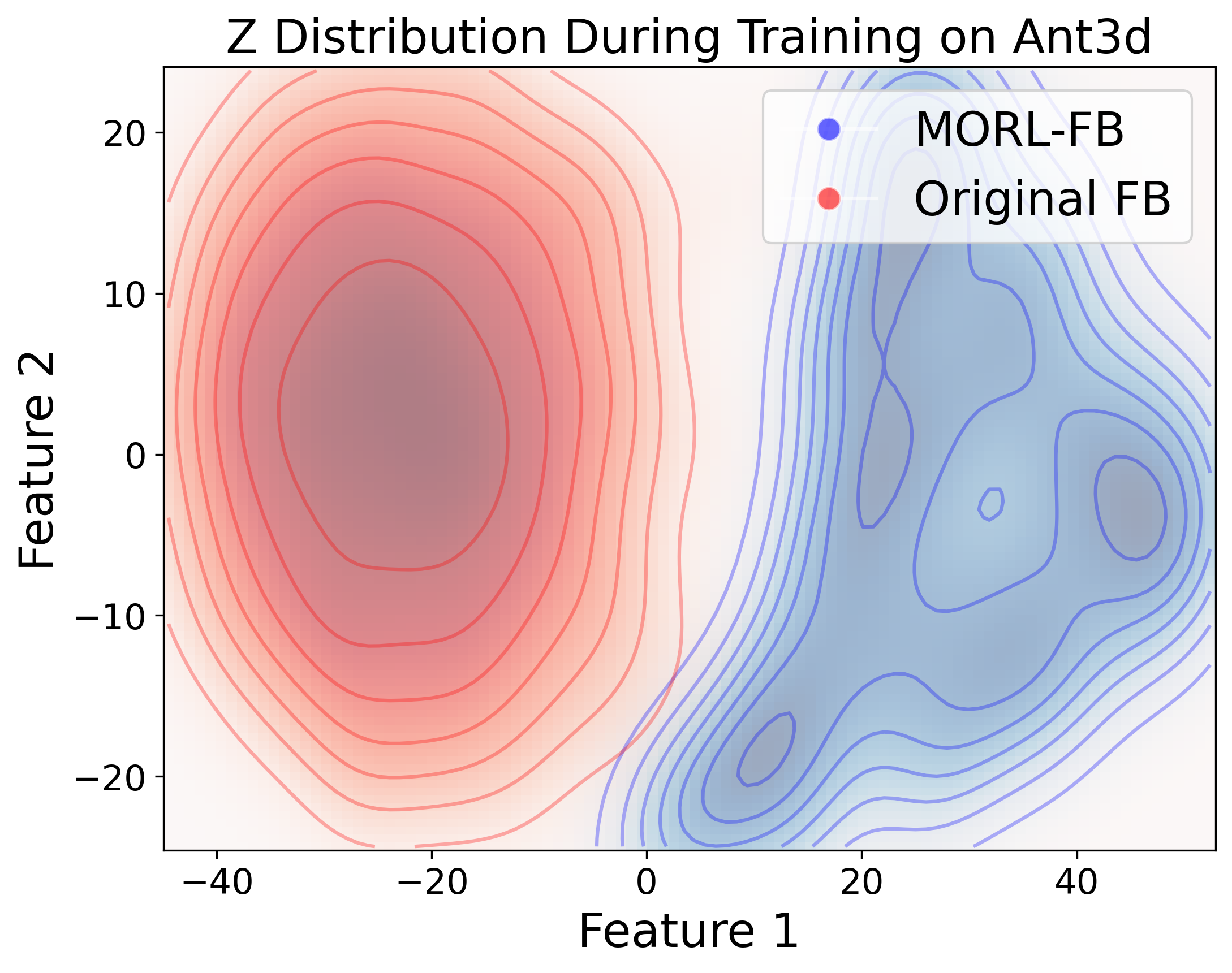}
    \caption{Empirical $\boldsymbol{z}$ distribution under MORL-FB with preference-guided sampling versus Original FB with simple normal distributions on Ant3d.}
    \label{fig:ant}
\end{figure}

We analyze the learned latent representations of MORL-FB by visualizing the distribution of sampled z across various environments using t-SNE \citep{van2008visualizing}.  These visualizations offer insight into the method's ability to capture the underlying structure of the multi-objective tasks.  Consistent with the observations on Humanoid2d (\Cref{fig:z_distribution}), the visualizations for Walker2d, Hopper3d, and Ant3d (\Cref{fig:walker}, \Cref{fig:hopper}, and \Cref{fig:ant}, respectively) demonstrate that preference-guided sampling yields distinct distributions compared to sampling from a standard normal distribution.  These results further support the hypothesis that preference-guided sampling promotes the exploration of a more diverse set of latent representations, which may contribute to improved generalization and adaptation on various objectives.

\begin{figure}[h!]
    \centering
    \includegraphics[width=0.7\linewidth]{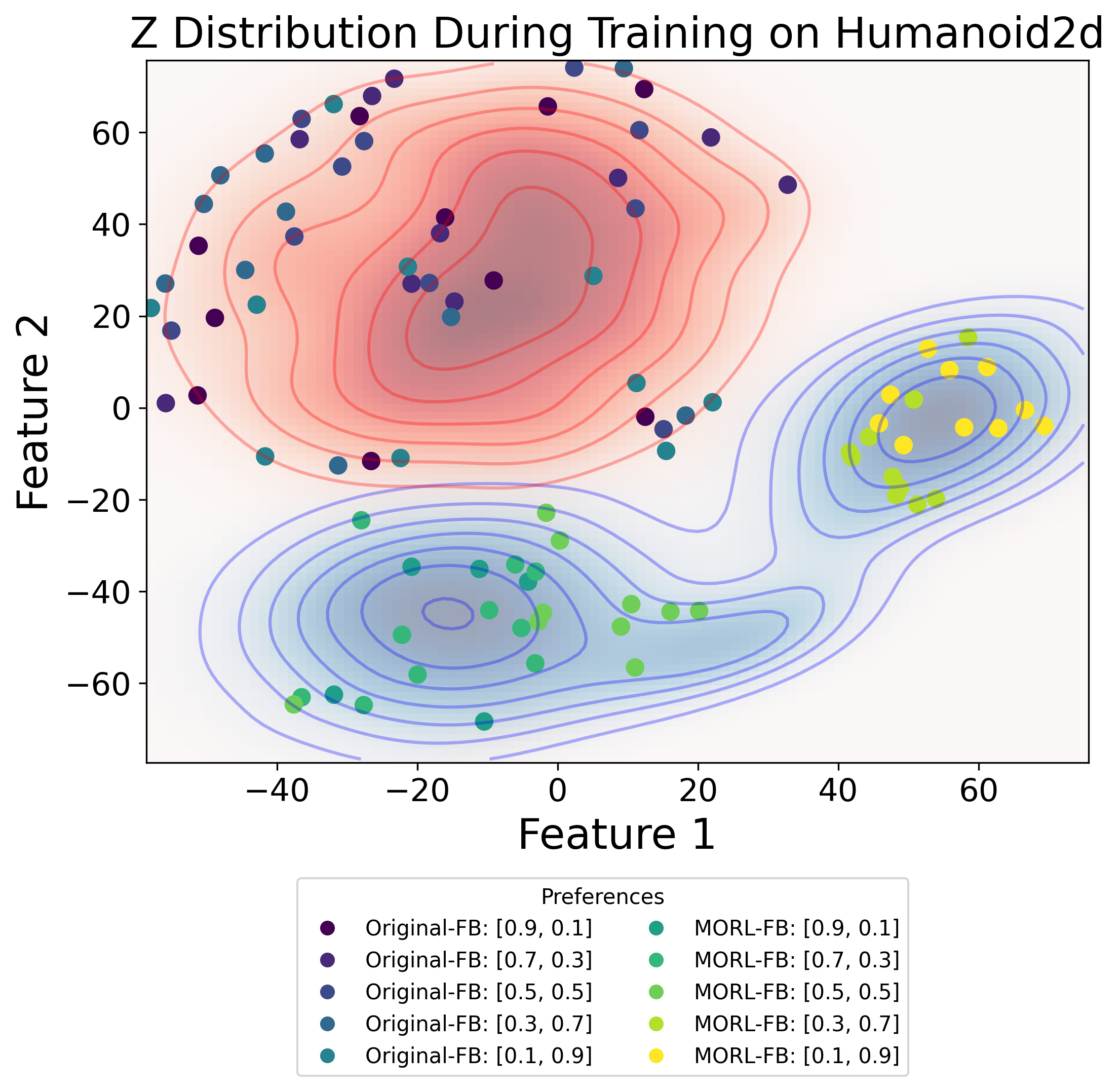}
    \caption{Empirical z distribution under MORL-FB with preference-guided sampling (blue) versus
Original FB with simple normal distributions (red) on Humanoid2d. This figure is slightly different from Figure \ref{fig:z_distribution} due to the additional preference points and the inherent randomness of t-SNE.}
    \label{fig:z_distribution_add}
\end{figure}

{
 To shows the multi-modality of MORL-FB, we visualized the positions of latent variables $z$ inferred from different preferences on the t-SNE plot in Figure \ref{fig:z_distribution_add}. This demonstrates that MORL-FB effectively encodes different preferences into separate regions of the latent space, leading to more diverse policies. To further illustrate this, we provide a demo of the policies learned by MORL-FB and vanilla FB in this link: \url{https://imgur.com/a/ehx1v7q}, where the z’s are selected from different positions on the t-SNE plot.
}

\subsection{Ablation Study}
\label{section:ablation}
\subsubsection{Experiment on Preference-Guided Exploration}
To assess the data efficiency of our proposed MORL-FB, we compared its performance against PD-MORL \citep{basaklar2023pd}. PD-MORL necessitates three to five times more training samples to train its preference interpolator. Can MORL-FB maintain superior performance while reducing data requirements?

\Cref{exp:with_interpolator} reveals that MORL-FB consistently outperforms PD-MORL across various environments. Specifically, MORL-FB achieves higher UT values in five out of six tasks, demonstrating its effectiveness in most scenarios. In more complex settings, such as MO-Humanoid, how does MORL-FB compare in HV results? The results indicate that MORL-FB remains competitive, underscoring its significant data efficiency gains. By eliminating the need for additional data to pretrain an interpolator, MORL-FB achieves competitive or superior performance while requiring significantly fewer training samples in multi-objective environments.

\begin{table*}[h!]
    \centering
    \caption{Performance comparison between MORL-FB and PD-MORL (with interpolator) across key metrics (UT, HV, and ED) on various multi-objective tasks.}
     \scalebox{0.85}{\begin{tabular}{@{}lccccccccc@{}}
        \\
        \toprule
        \multirow{2}{*}{\raisebox{0.3ex}}{Environments} & \multirow{2}{*}{\raisebox{0.3ex}}{Metrics} & \multirow{2}{*}{\raisebox{0.3ex}}{PD-MORL} & \multirow{2}{*}{\raisebox{0.3ex}}{MORL-FB} \\
        & & (w/i interpolator) & \\ \midrule

        \multirow{3}{*}{Halfcheetah2d} 
        & UT($\times$ \num{e3}) & 5.62 ± 0.05 & \textbf{7.69 ± 0.08}  \\
        & HV($\times$ \num{e8}) & 1.08 ± 0.00 & \textbf{1.24 ± 0.00} \\
        % & SP($\times$ \num{e5}) & 82.48 ± 11.29 & 33.6 ± 6.6 \\
        & ED & 0.07 $\pm$ 0.01 & - \\ \midrule
        
        \multirow{3}{*}{Walker2d} 
        & UT($\times$ \num{e3}) & 2.18 ± 0.02 & \textbf{2.23 ± 0.03} \\
        & HV($\times$ \num{e7}) & \textbf{5.45 ± 0.01} & 4.32 ± 0.02 \\
        % & SP($\times$ \num{e5}) & 0.06 ± 0.01 & 0.02 ± 0.01 \\
        & ED & 0.56 $\pm$ 0.00 & - \\ \midrule
        
        \multirow{3}{*}{Hopper3d} 
        & UT($\times$ \num{e3}) & 2.26 ± 0.01 & \textbf{2.36 ± 0.01} \\
        & HV($\times$ \num{e11}) & \textbf{1.08 ± 0.00} & 1.15 ± 0.00 \\
        % & SP($\times$ \num{e4}) & 0.12 ± 0.03 & 0.23 ± 0.05 \\
        & ED & 0.20 $\pm$ 0.01 & - \\ \midrule
        
        \multirow{3}{*}{Ant3d} 
        & UT($\times$ \num{e3}) & \textbf{3.59 ± 0.06} & 3.43 ± 0.22 \\
        & HV($\times$ \num{e11}) & \textbf{4.20 ± 0.04} & 4.18 ± 0.04 \\
        % & SP($\times$ \num{e5}) & 0.79 ± 0.008 & 0.532 ± 0.131 \\
        & ED & 0.60 $\pm$ 0.00 & -  \\ \midrule

        \multirow{3}{*}{Humanoid2d} 
        & UT($\times$ \num{e2}) & 2.93 ± 0.07 & \textbf{8.13 ± 0.01} \\
        & HV($\times$ \num{e7}) & 1.06 ± 0.01 & \textbf{1.75 ± 0.02} \\
        % & SP($\times$ \num{e5}) & 0.07 ± 0.02 & 22.92 ± 2.86 \\
        & ED & 0.33 $\pm$ 0.01 & - \\ \midrule

        \multirow{3}{*}{Humanoid5d}
        & UT($\times 10^{3}$) & 0.93 ± 0.04 & \textbf{1.11 ± 0.00} \\
        & HV($\times 10^{15}$) & 6.64 ± 0.09 & \textbf{6.99 ± 0.06} \\
        % & SP($\times 10^{3}$) & 0.018 ± 0.0006 & 1.03 ± 0.137 \\
        & ED & 0.43 $\pm$ 0.01 & - \\
        
        \bottomrule
    \end{tabular}}
    \label{exp:with_interpolator}
\end{table*}

\subsubsection{Experiments on \texorpdfstring{$\boldsymbol{z}$}{z} Dimension}

We investigate the impact of the $\boldsymbol{z}$ dimension in the Hopper3d environment. As shown in \Cref{tab:ablation_z_dim}, the performance metric increases with the $\boldsymbol{z}$ dimension. However, when the $\boldsymbol{z}$ dimension reaches 300, performance declines, likely due to insufficient training steps for the larger network.

\begin{table*}[!h]
    \centering
    \caption{Empirical study on $\boldsymbol{z}$ dimension on Hopper3d.}
     \scalebox{0.9}{\begin{tabular}{@{}lccccccccc@{}}
        \\
        \toprule
        \multirow{2}{*}{\raisebox{0.3ex}}{$\boldsymbol{z}$ dimension} & \multirow{2}{*}{\raisebox{0.3ex}}{Metrics}
        & \multirow{2}{*}{\raisebox{0.3ex}}{MORL-FB}  \\
        \midrule

        \multirow{2}{*}{50} 
        & UT($\times$ \num{e3}) & 2.21 $\pm$ 0.01 \\
        & HV($\times$ \num{e11}) & 1.03 $\pm$ 0.05 \\
        % & SP($\times$ \num{e5}) & 33.6 $\pm$ 6.6 & 2.03 $\pm$ 1.38  \\
        % & ED & 0.07 $\pm$ 0.03 & - \\ \midrule
        \midrule

        \multirow{2}{*}{100} 
        & UT($\times$ \num{e3}) & 2.25 $\pm$ 0.00 \\
        & HV($\times$ \num{e11}) & 1.13 $\pm$ 0.01 \\
        % & SP($\times$ \num{e5}) & 0.02 $\pm$ 0.01 & 1.24 $\pm$ 0.31 & 0.1 $\pm$ 0.05 \\
        % & ED & 0.05 $\pm$ 0.01 & -  \\ 
        % \midrule
        \midrule

        \multirow{2}{*}{150} 
        & UT($\times$ \num{e3}) & 2.36 $\pm$ 0.01 \\
        & HV($\times$ \num{e11}) & 1.15 $\pm$ 0.00 \\
        \midrule

        \multirow{2}{*}{300} 
        & UT($\times$ \num{e3}) & 2.01 $\pm$ 0.01 \\
        & HV($\times$ \num{e11}) & 0.97 $\pm$ 0.02 \\
        
        \bottomrule
    \end{tabular}}
    \label{tab:ablation_z_dim}
\end{table*}

\begin{figure}[!h]
        \centering
        \includegraphics[width=0.8\linewidth]{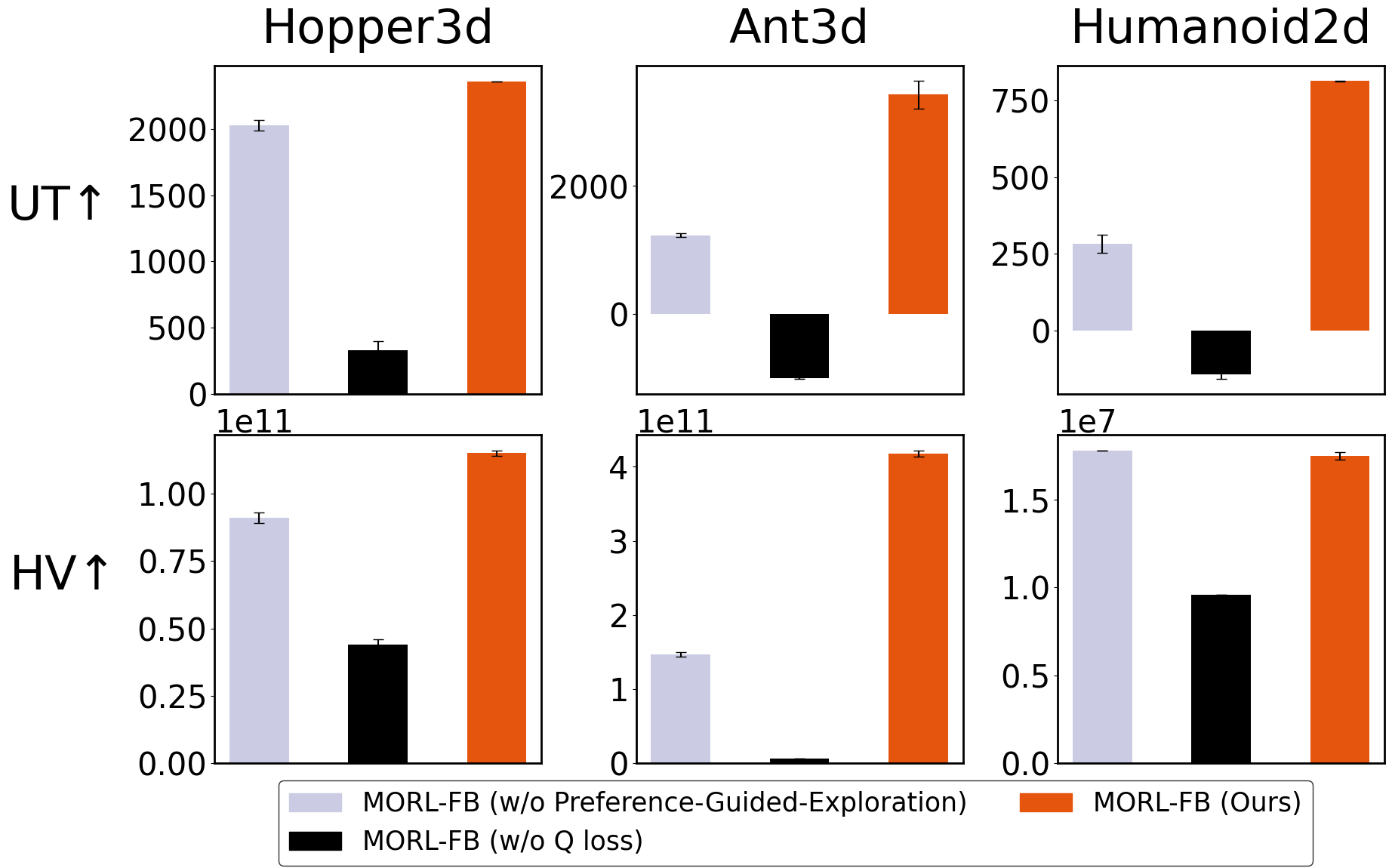}
        \caption{Evaluation of MORL-FB and its ablated versions across different environments. The results highlight the importance of preference-guided sampling and the auxiliary Q-loss for MORL performance.}
        \label{fig:MORL_ablation_bar}
		
		\end{figure}

\rev{\subsubsection{Experiments on Auxiliary Loss}}
\rev{While \citet{ahmed2023zero} also includes a similar auxiliary loss, there is one salient difference between theirs and our Q loss \Cref{eq:q_loss_main}: As the original FB is designed for RFRL, it does not have the reward signal available at training and hence needs to construct a pseudo reward as $B^\top \mathbb{E}[BB^\top]^{-1}z$ (cf. Equation (9) in~\citep{ahmed2023zero}). On the other hand, as MORL-FB addresses MORL and can observe vector reward signals at training, we propose to use the actual scalarized reward $\lambda^\top r$ in the Q loss. While this algorithmic difference appears seemingly subtle, this design makes a huge difference in the performance. Below we show an ablation study that compares MORL-FB with our Q loss (denoted as “Original”) and MORL-FB with the auxiliary loss using pseudo reward in FB (denoted as “Pseudo Q Loss”). The results are summarized in \Cref{tab:ablation_z_lambda_reward,fig:MORL_ablation_bar}. This shows that the auxiliary loss of FB cannot be directly applied and needs to be adapted properly in the context of MORL. Note that we use the term “auxiliary” since the original FB is directly built on the measure loss and hence the Q loss is auxiliary for learning FB representation, rather than being unimportant for MORL.}

\begin{table*}[!h]
    \centering
    \caption{Ablation study of MORL-FB on preference-guided exploration.}
    \scalebox{0.9}{
     \begin{tabular}{@{}lccccccccc@{}}
        \toprule
        \multirow{2}{*}{\raisebox{0.3ex}}{Environments} & \multirow{2}{*}{\raisebox{0.3ex}}{Metrics} & \multirow{2}{*}{\raisebox{0.3ex}}{MORL-FB} 
        & \multirow{2}{*}{\raisebox{0.3ex}}{MORL-FB} & \multirow{2}{*}{\raisebox{0.3ex}}{MORL-FB} \\ & &  (w/o PG-Explore) & (Pseudo Q Loss) & (Ours) \\
        \midrule

        \multirow{2}{*}{Ant3d} 
        & UT($\times$ \num{e3}) & 1.45 $\pm$ 0.02 & 1.27 $\pm$ 0.16 & \textbf{3.93} $\pm$ \textbf{0.04} \\
        & HV($\times$ \num{e11}) & 1.25 $\pm$ 0.02 & 1.91 $\pm$ 0.02 & \textbf{3.85} $\pm$ \textbf{0.01} \\
        %\midrule
        
        %\multirow{2}{*}{Hopper3d} 
        %& UT($\times$ \num{e3}) & 1.46 $\pm$ 0.30 & 0.58 $\pm$ 0.00 & 1.11 %$\pm$ 0.02 \\
        %& HV($\times$ \num{e11}) & 0.89 $\pm$ 0.01 & 0.06 $\pm$ 0.00 & 1.08 %$\pm$ 0.02 \\
        \bottomrule
    \end{tabular}}
    \label{tab:ablation_z_lambda_reward}
\end{table*}

\rev{\subsubsection{Experiments on Measure Loss}}
\rev{Q-loss is surely important in our method as it guides our representation to learn the reward function used in the environment. On the other hand, both the preference-guided exploration and the measure loss are also essential to the success of MORL-FB based on Figure \ref{fig:MORL_ablation_bar} and an additional ablation study on the measure loss shown below.}

\begin{table*}[!h]
    \centering
    \caption{Ablation study of MORL-FB on measure loss and Q loss.}
    \scalebox{0.9}{
     \begin{tabular}{@{}lccccccccc@{}}
        \toprule
        \multirow{2}{*}{\raisebox{0.3ex}}{Environments} & \multirow{2}{*}{\raisebox{0.3ex}}{Metrics} & \multirow{2}{*}{\raisebox{0.3ex}}{MORL-FB} 
        & \multirow{2}{*}{\raisebox{0.3ex}}{MORL-FB} & \multirow{2}{*}{\raisebox{0.3ex}}{MORL-FB} \\ & &  (w/o measure loss) & (w/o Q loss) & (Ours) \\
        \midrule

        \multirow{2}{*}{Ant3d} 
        & UT($\times$ \num{e3}) & 1.37 $\pm$ 0.28 & -1.53 $\pm$ 0.00 & \textbf{3.93} $\pm$ \textbf{0.04} \\
        & HV($\times$ \num{e11}) & 2.74 $\pm$ 0.02 & 0.00 $\pm$ 0.00 & \textbf{3.85} $\pm$ \textbf{0.01} \\
        % \midrule
        
        %\multirow{2}{*}{Hopper3d} 
        %& UT($\times$ \num{e3}) & 1.35 $\pm$ 0.01 & 0.37 $\pm$ 0.00 & 1.11 %$\pm$ 0.02 \\
        %& HV($\times$ \num{e11}) & 0.96 $\pm$ 0.01 & 0.07 $\pm$ 0.03 & 1.08 %$\pm$ 0.02 \\
        \bottomrule
    \end{tabular}}
    \label{tab:ablation_loss}
\end{table*}

\iclrwh{
\subsubsection{Experiments on Q-loss Coefficient}
In our implementation, the measure loss $\mathcal{L}_M$ and the Q loss $\mathcal{L}_Q$ are directly added together without an additional weighting term. We chose this default configuration because it already yields strong and stable performance across all evaluated tasks, and we did not observe a need for further hyperparameter tuning during development. Moreover, we have conducted an additional sensitivity analysis to better understand how the balance between these two losses affects performance. Specifically, we varied the Q-loss coefficient $\alpha_Q$ (with $\alpha_Q=1$ as the default setting) and reported the corresponding results in Ant3d. \Cref{tab:Q_coeff} shows that MORL-FB’s performance is largely insensitive to the choice of $\alpha_Q$, indicating that directly adding $\mathcal{L}_M$ and $\mathcal{L}_Q$ is a reasonable and robust design choice.
\begin{table*}[!h]
    \centering
    \caption{Performance of MORL-FB with different Q-loss coefficients.}
    \label{tab:Q_coeff}
    \scalebox{0.9}{
    \begin{tabular}{@{}lcccccc@{}}
        \toprule
        \multirow{2}{*}{\raisebox{0.3ex}{Environments}} & \multirow{2}{*}{\raisebox{0.3ex}{Metrics}} & \multicolumn{4}{c}{MORL-FB} \\
        \cmidrule(lr){3-6}
        & & $\alpha_Q=0.25$ & $\alpha_Q=0.5$  & $\alpha_Q=1$ (Default) & $\alpha_Q=2$\\
        \midrule
        \multirow{2}{*}{Ant3d} 
        & UT($\times \num{e3}$) & 3.76 $\pm$ 0.41 & 3.79 $\pm$ 0.32 & 3.77 $\pm$ 0.11 & \textbf{3.95 $\pm$ 0.13} \\
        & HV($\times \num{e11}$) & 4.03 $\pm$ 0.22 & 4.13 $\pm$ 0.19 & 3.95 $\pm$ 0.13 & \textbf{4.23 $\pm$ 0.04} \\
        \bottomrule
    \end{tabular}}
\end{table*}
}
\subsection{Performance Comparison of MORL-FB Under Different Q-loss Coefficients}
\label{ap:discrete_env}
%\iclrwh{
We evaluate MORL-FB and several baselines on two classic discrete control tasks: Deep Sea Treasure (DST) and Fruit Tree Navigation (FTN). DST features the fundamental trade-off between two conflicting objectives. FTN, on the other hand, involves a tree search process with 6-dimensional terminal rewards, as previously detailed in the evaluation environments section. The detailed configurations for these experiments are provided in \Cref{app:experiment_config}.
%}

\begin{figure}[H]
    \centering
    \includegraphics[width=0.7\linewidth]{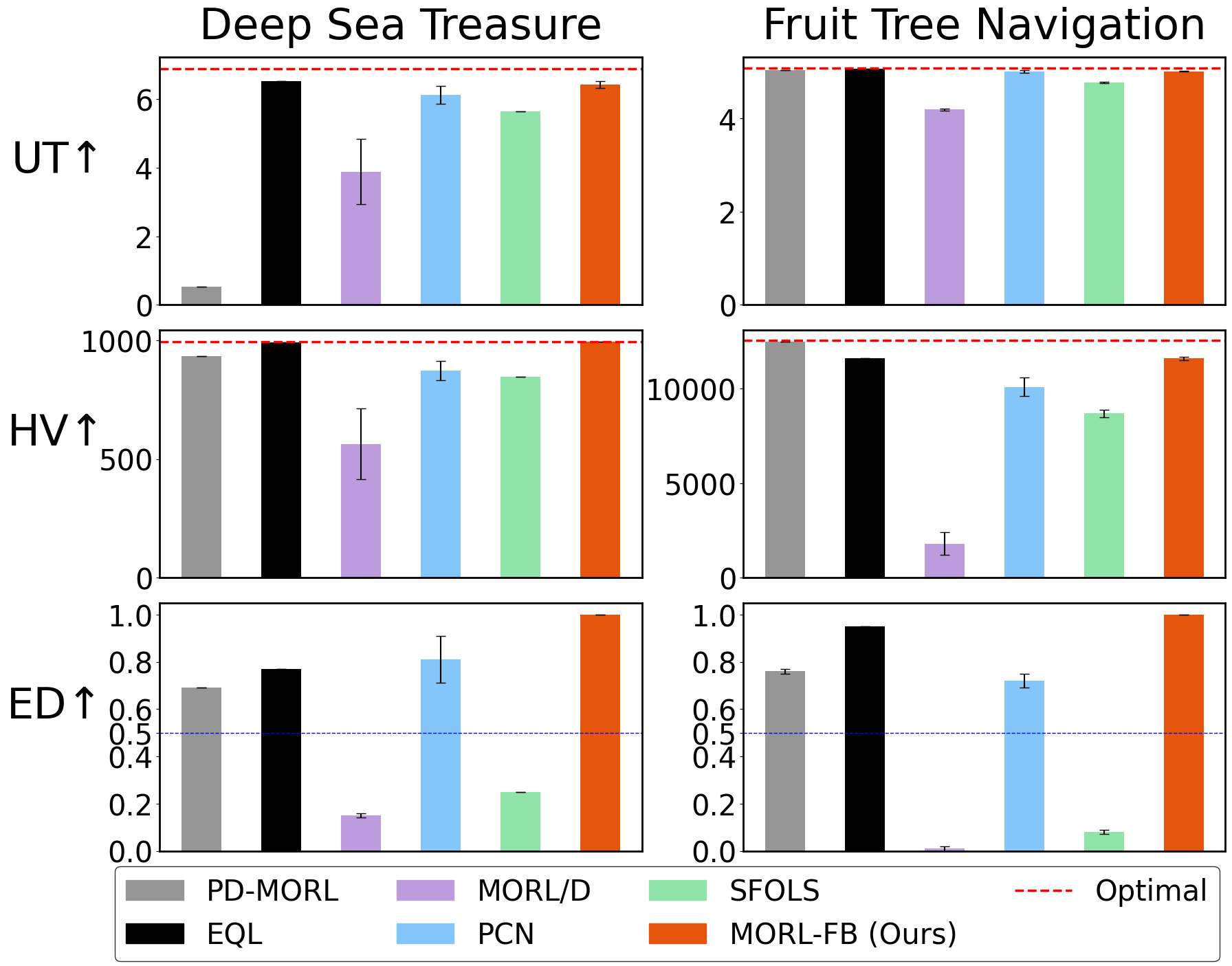}
    \caption{\textbf{Competitive Results of MORL-FB on Discrete Control Tasks.} We evaluate MORL-FB and several benchmark MORL algorithms on classic discrete control tasks in MO-Gymnasium. Performance is measured using UT, HV and ED. MORL-FB demonstrates competitive results against specialized discrete MORL algorithms.}
    \label{fig:discrete_bar}
\end{figure}

\Cref{fig:discrete_bar} shows the performance of all the methods in UT, HV, and ED for discrete control tasks. Regarding ED, for each baseline algorithm $\texttt{ALG}$, we report $\text{ED}(\texttt{ALG},\texttt{MORL-FB})$ to show the pairwise comparison. We can observate that MORL-FB consistently achieves competitive or superior performance across all three metrics on the discrete control tasks Deep Sea Treasure and Fruit Tree Navigation.

\subsection{Performance Comparison of MORL-FB Under State-Action-Based Rewards and State-Based Rewards}
\label{ap:sar}
\begin{figure*}[!h]
    \centering
    \includegraphics[width=0.7\textwidth]{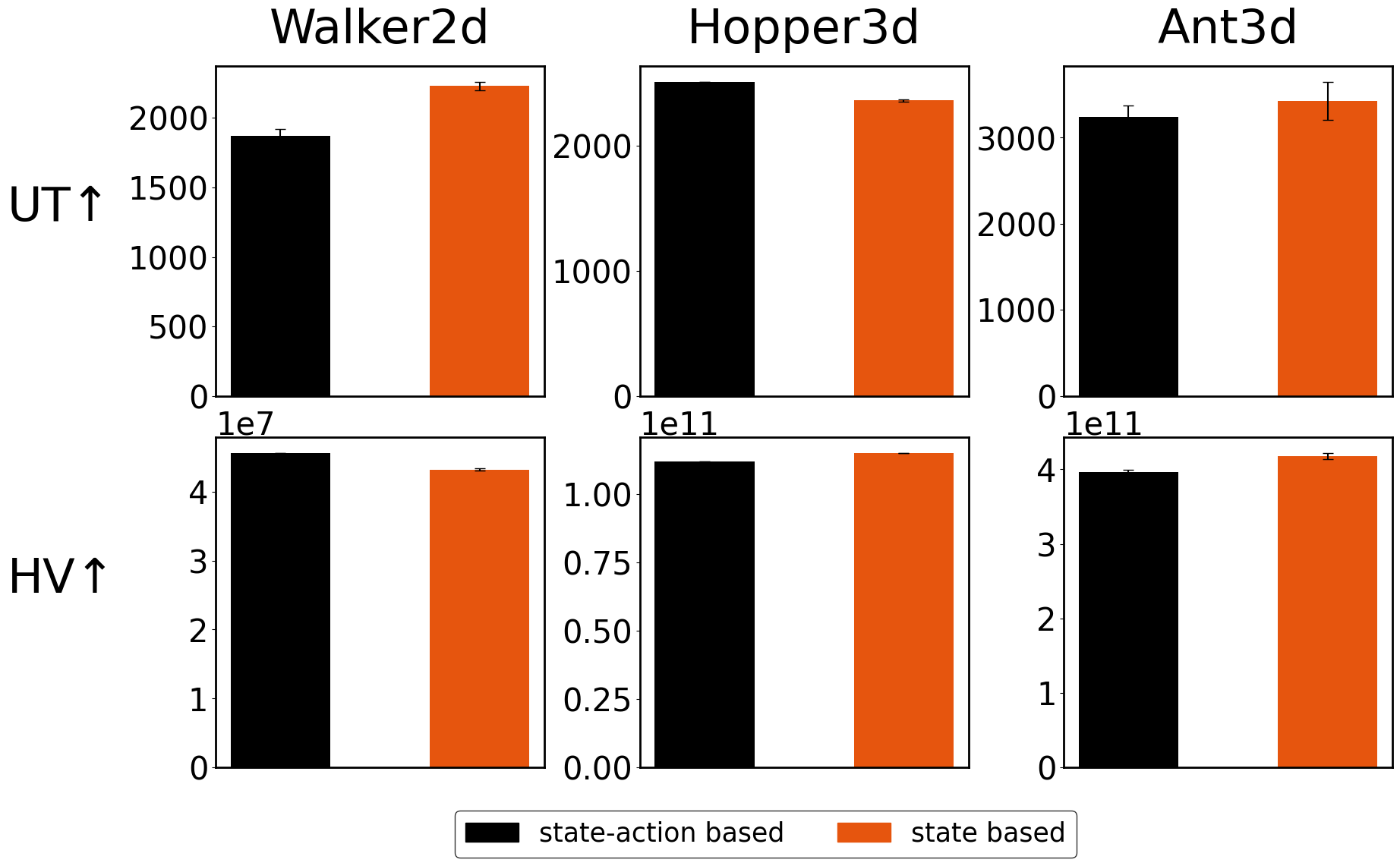}
    \caption{\textbf{Evaluation of MORL-FB with different reward function representations.} This figure presents the performance of MORL-FB when reward functions depend on states only (\ie $\bm{R}(s)$) versus state-action pairs (\ie $\bm{R}(s,a)$).}
    \label{fig:state_action}
\end{figure*}

 In this paper, we primarily focus on state-based rewards. However, as the original FB supports both state-based~\citep{ahmed2023zero} and state-action-based rewards~\citep{touati2021learning}, MORL-FB can also be extended to state-action-based rewards by replacing $B(s)$ with $B(s,a)$. Since some MO MuJoCo rewards depend on both states and actions, we compare MORL-FB and the extended one. As shown in the \Cref{fig:state_action}, the state-action-based variant yields slight performance improvements on several tasks.

\begin{figure*}[!h]
    \centering
    \includegraphics[width=0.35\textwidth]{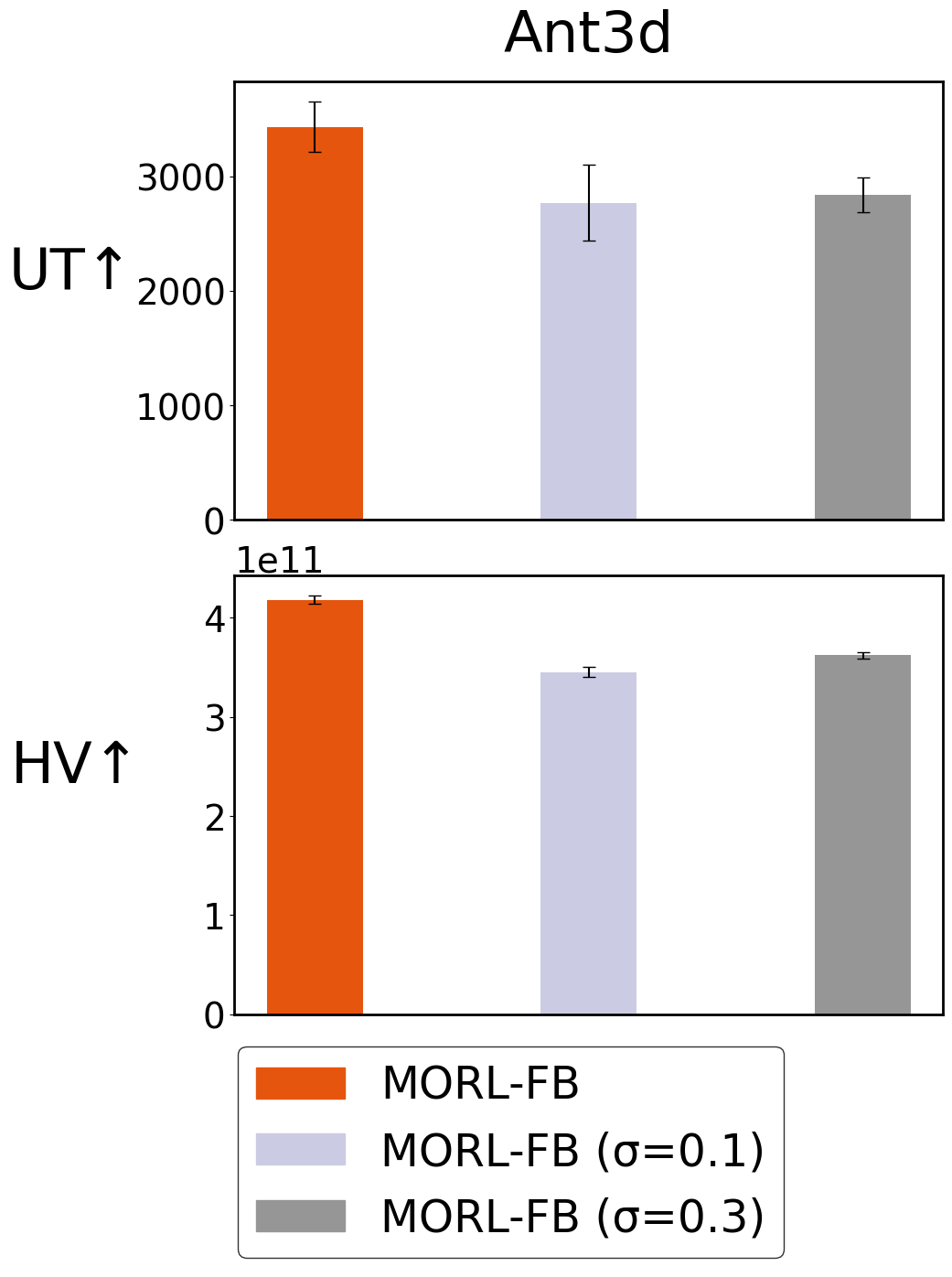}
    \caption{\textbf{Evaluation of MORL-FB under stochastic reward.} This figure assesses the performance of MORL-FB in environments featuring stochastic reward functions.}
    \label{fig:MORLFB_noise}
\end{figure*}

{
\subsection{Performance Comparison of MORL-FB Under Stochastic Rewards}
As vanilla FB naturally handles stochastic rewards, MORL-FB inherits this capability. To further demonstrate this, we evaluated MORL-FB under stochastic rewards by adding zero-mean Gaussian noise $\mathcal{N}(0,\sigma^2)$, similar to prior work~\citep{romoff2018rewardestimationvariancereduction,hu2022distributional}. The result is shown in \Cref{fig:MORLFB_noise}.
}

\rev{\subsection{Performance Comparison of MORL-FB With Nonlinear Scalarization}}
\rev{While we focus on linear scalarization in this paper, MORL-FB can be readily extended to nonlinear scalarization schemes by replacing $r^\top\lambda$ with $f_\lambda(r)$ when sampling $z$ for preference-guided exploration and when computing Q loss, where $f_\lambda(r)$ is the general scalarization function. This is feasible since the original FB is designed to handle any scalar reward function, and MORL-FB inherits this property from FB and can also handle nonlinear scalarization. To demonstrate this generalizability, we further evaluate MORL-FB on Halfcheetah2d by training under smooth Tchebycheff scalarization as $f_\lambda(r) = \mu \log \left(\sum_{i=1}^m \exp\left(\frac{\lambda_i(r-r_{ref})}{\mu}\right)\right)$, where $\lambda_i$ = is the $i$-th entry of preference vector, $\mu$ is the smoothing parameter and set to 0.1, and $r_{ref}$ is set to [2.0, 0.0] across training~\citep{lin2024tchebycheffscalarization, qiu2024traversingparetooptimalpolicies}. We see that MORL-FB still achieves comparably strong performance in HV and UT under nonlinear scalarization.}

\begin{table*}[!h]
    \centering
    \caption{Performance of MORL-FB with Smooth Tchebycheff scalarization.}
    \scalebox{0.9}{
     \begin{tabular}{@{}lccccccccc@{}}
        \toprule
        \multirow{2}{*}{\raisebox{0.3ex}}{Environments} & \multirow{2}{*}{\raisebox{0.3ex}}{Metrics} & \multirow{2}{*}{\raisebox{0.3ex}}{MORL-FB}
        & \multirow{2}{*}{\raisebox{0.3ex}}{MORL-FB} \\ & & {(linear scalarization)} & {(Smooth Tchebycheff scalarization)} \\
        \midrule

        \multirow{2}{*}{HalfCheetah2d} 
        & UT($\times$ \num{e3}) & 7.69 $\pm$ 0.08 & 6.33 $\pm$ 0.02  \\
        & HV($\times$ \num{e8}) & 1.24 $\pm$ 0.00 & 1.00 $\pm$ 0.01  \\
        
        \bottomrule
    \end{tabular}}
    \label{tab:stch}
\end{table*}

\subsection{Sample Efficiency of MORL-FB}
\begin{figure*}[!h]
    \centering
    \includegraphics[width=0.9\textwidth]{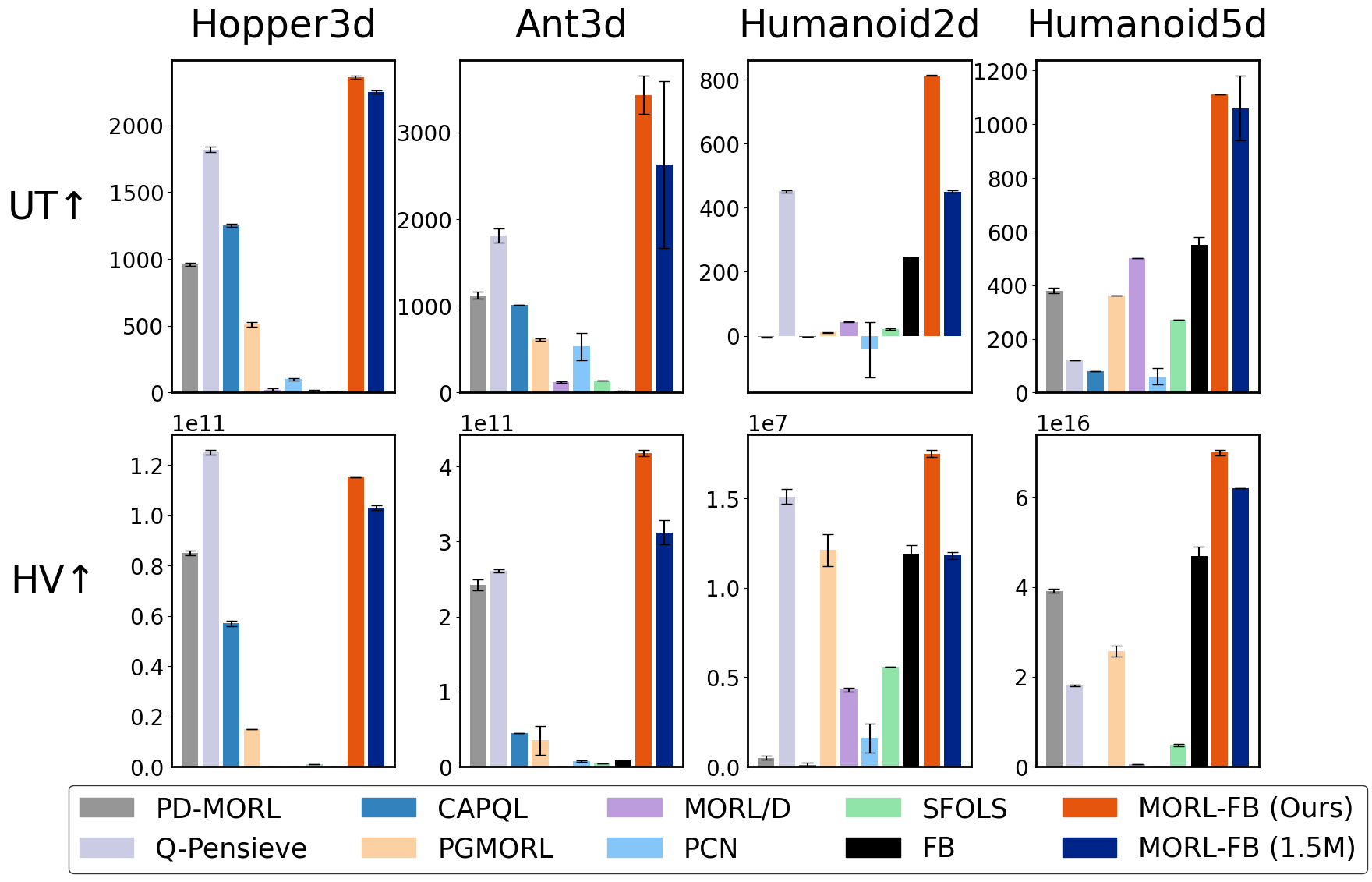}
    \caption{\textbf{Performance of MORL-FB on continuous control tasks.} We evaluate MORL-FB (1.5M training steps) against several benchmark MORL algorithms (3M training steps) on diverse continuous control tasks from MO-Gymnasium. Utilizing key metrics, these results demonstrate that MORL-FB outperforms baselines, achieving superior HV and UT across most tasks despite significantly fewer training steps.}
    \label{fig:continuous_half}
\end{figure*}

\begin{figure}[!h]
    \centering
    \begin{minipage}[t]{0.48\textwidth}
    \centering
    \includegraphics[width=1\textwidth]{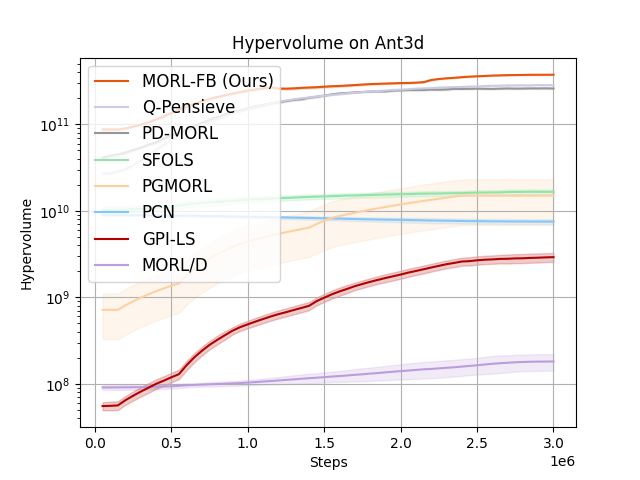}
    \caption{\textbf{Learning curves for MORL-FB and benchmark algorithms on Ant3d.} This figure presents the learning curves in terms of Hypervolume (HV) for MORL-FB and several benchmark MORL algorithms evaluated on Ant3d.}
    \label{fig:learn_hv}
    \end{minipage}
    \hfill
    \begin{minipage}[t]{0.48\textwidth}
    \centering
    \includegraphics[width=1\textwidth]{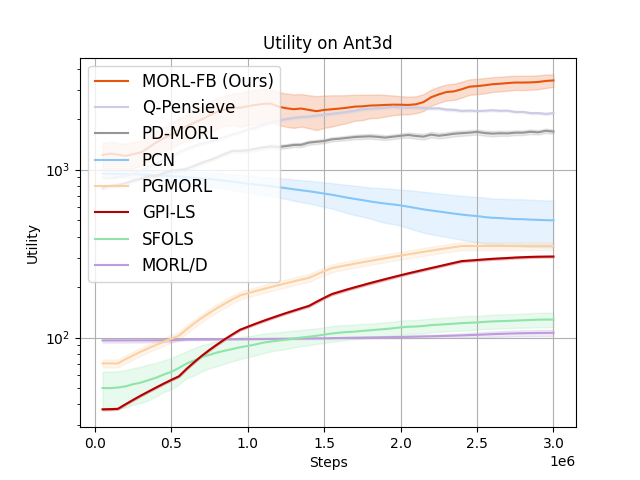}
    \caption{\textbf{Learning curves for MORL-FB and benchmark algorithms on Ant3d.} This figure presents the learning curves in terms of Utility (UT) for MORL-FB and several benchmark MORL algorithms evaluated on Ant3d.}
    \label{fig:learn_ut}
    \end{minipage}
\end{figure}

We demonstrate the sample efficiency of MORL-FB by evaluating its performance at an intermediate stage of 1.5M training steps. Notably, as shown in \Cref{fig:continuous_half}, MORL-FB achieves superior HV and UT scores across most tasks compared to baseline methods trained for a full 3M steps. This indicates that MORL-FB can attain high performance with significantly fewer environment interactions. Further evidence, presented in \Cref{fig:learn_hv} and \Cref{fig:learn_ut}, corroborates that MORL-FB reaches proficient performance levels with reduced training data.

\subsection{Cross-Objective Transfer Capability of MORL-FB}

To investigate how well MORL-FB handles transfer across different numbers of objectives, we conducted an empirical study on Hopper across different objective dimensions. We analyze the following cases:

\begin{itemize}
\item Hopper2d: Moving forward speed on the x-axis, control cost of the action
\item Hopper3d: Moving forward speed on the x-axis, jumping height on the z-axis, control cost of the action
\item Hopper4d: Moving forward speed on the x-axis, jumping height on the z-axis, jumping up speed on the z-axis, control cost of the action
\end{itemize}

\Cref{tab:objective_transfer} summarizes the quantitative results presented visually in \Cref{fig:objective_transform_bar}. MORL-FB consistently outperforms FB across all configurations in terms of utility and hypervolume. 

%% Cross-Objective Transfer
\begin{table*}[!h]
    \centering
    \caption{\textbf{Zero-shot cross-objective transfer from Hopper2d to Hopper3d and Hopper4d using vanilla FB and MORL-FB:} This figure presents the results demonstrating effective transfer by MORL-FB, supporting the efficacy of its proposed enhancements.}
     \scalebox{0.8}{\begin{tabular}{@{}lccccccccc@{}}
        \\
        \toprule
        \multirow{2}{*}{\raisebox{0.3ex}}{Environments} & \multirow{2}{*}{\raisebox{0.3ex}}{Metrics} & \multirow{2}{*}{\raisebox{0.3ex}}{FB} 
        & \multirow{2}{*}{\raisebox{0.3ex}}{MORL-FB}  \\
        \midrule

        \multirow{3}{*}{2D to 3D} 
        & UT($\times$ \num{e3}) & 0.02 $\pm$ 0.00 & \textbf{1.77 $\pm$ 0.00}  \\
        & HV($\times$ \num{e10}) & 0.40 $\pm$ 0.00 & \textbf{7.81 $\pm$ 0.06}  \\

        & ED & 0.07 $\pm$ 0.03 & - \\ \midrule
        
        \multirow{3}{*}{2D to 4D} 
        & UT($\times$ \num{e3}) & 0.06 $\pm$ 0.00 & \textbf{1.59 $\pm$ 0.00}  \\
        & HV($\times$ \num{e13}) & 0.05 $\pm$ 0.00 & \textbf{7.78 $\pm$ 0.10}  \\

        & ED & 0.05 $\pm$ 0.01 & -  \\ 
        % \midrule
        
        \bottomrule
    \end{tabular}}
    \label{tab:objective_transfer}
\end{table*}

\begin{figure}[t]
    \centering
    \includegraphics[width=0.5\linewidth]{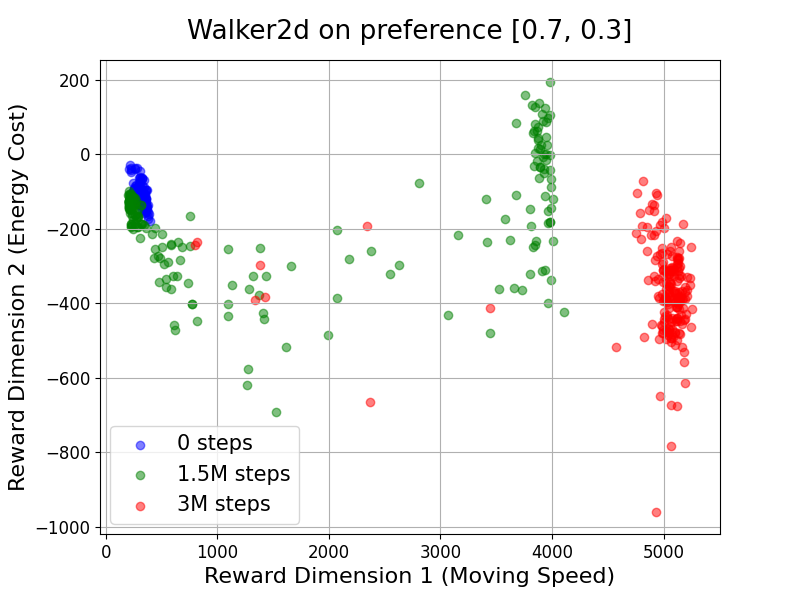}
    \caption{Return vectors (Moving Speed vs. Energy Cost) achieved at the initial, intermediate, and final training stages in Walker2d with preference [0.7, 0.3]: The scatter plot, particularly at 1.5M steps, highlights the diverse policies beyond the CCS policy, supporting that RFRL serves as auxiliary tasks beneficial for MORL.}
    \label{fig:walker_73}
\end{figure}

\rev{\subsection{RFRL as a source of auxiliary tasks}
During training, the $\bm{z}$ vectors computed for each preference $\lambda$ are diverse, covering both CCS and non-CCS policies. Learning from non-CCS policies serves as auxiliary tasks. From~\Cref{fig:walker_73}, we find that the return vectors achieved by those $\bm{z}$-induced polices at the 1.5 million training step span both non-CCS and CCS regions.}

\begin{figure}[H]
    \centering
    \begin{minipage}{0.48\linewidth}
        \centering
        \includegraphics[width=\linewidth]{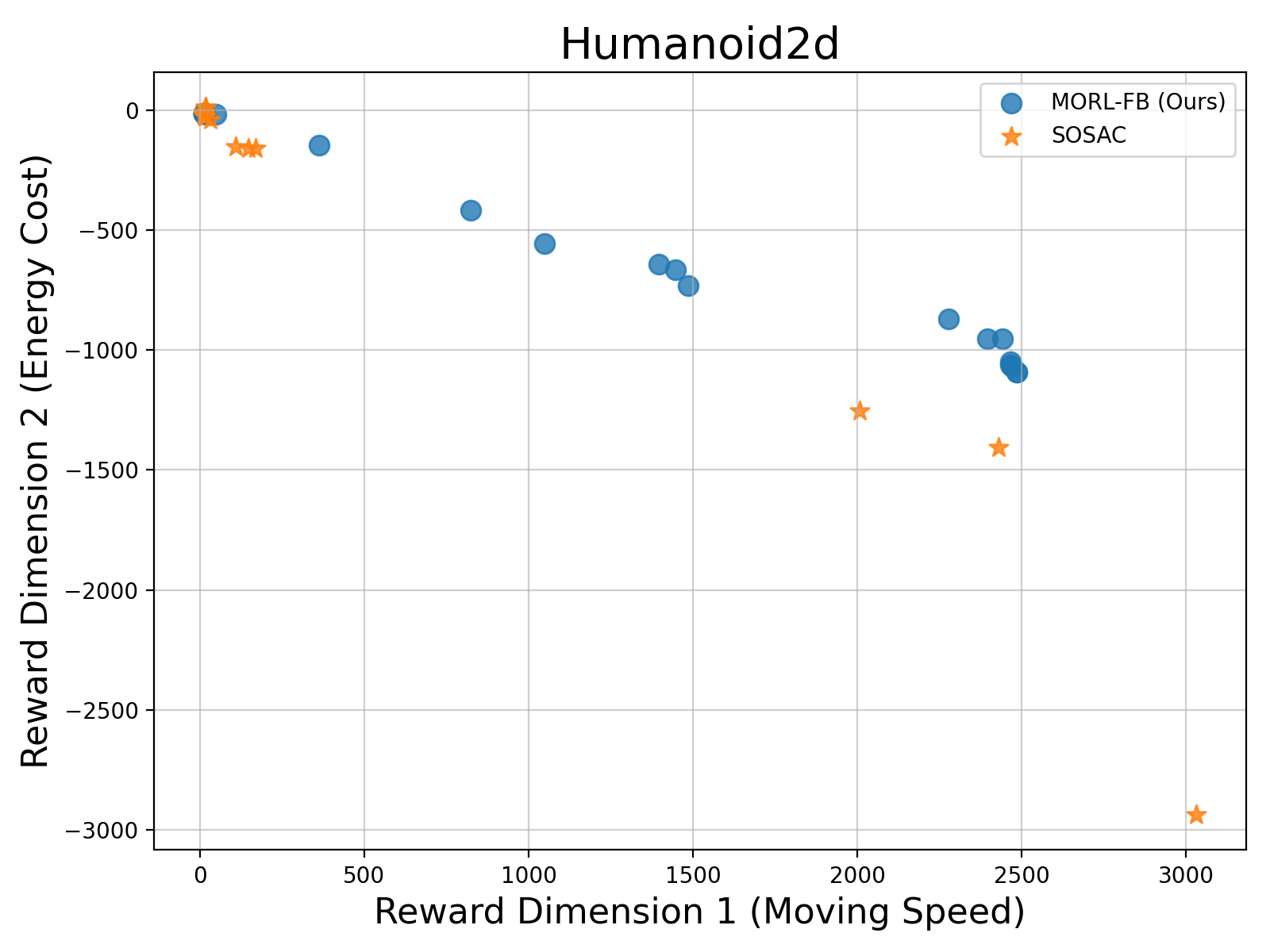}
        \subcaption{Humanoid2d}
    \end{minipage}
    \hfill
    \begin{minipage}{0.48\linewidth}
        \centering
        \includegraphics[width=\linewidth]{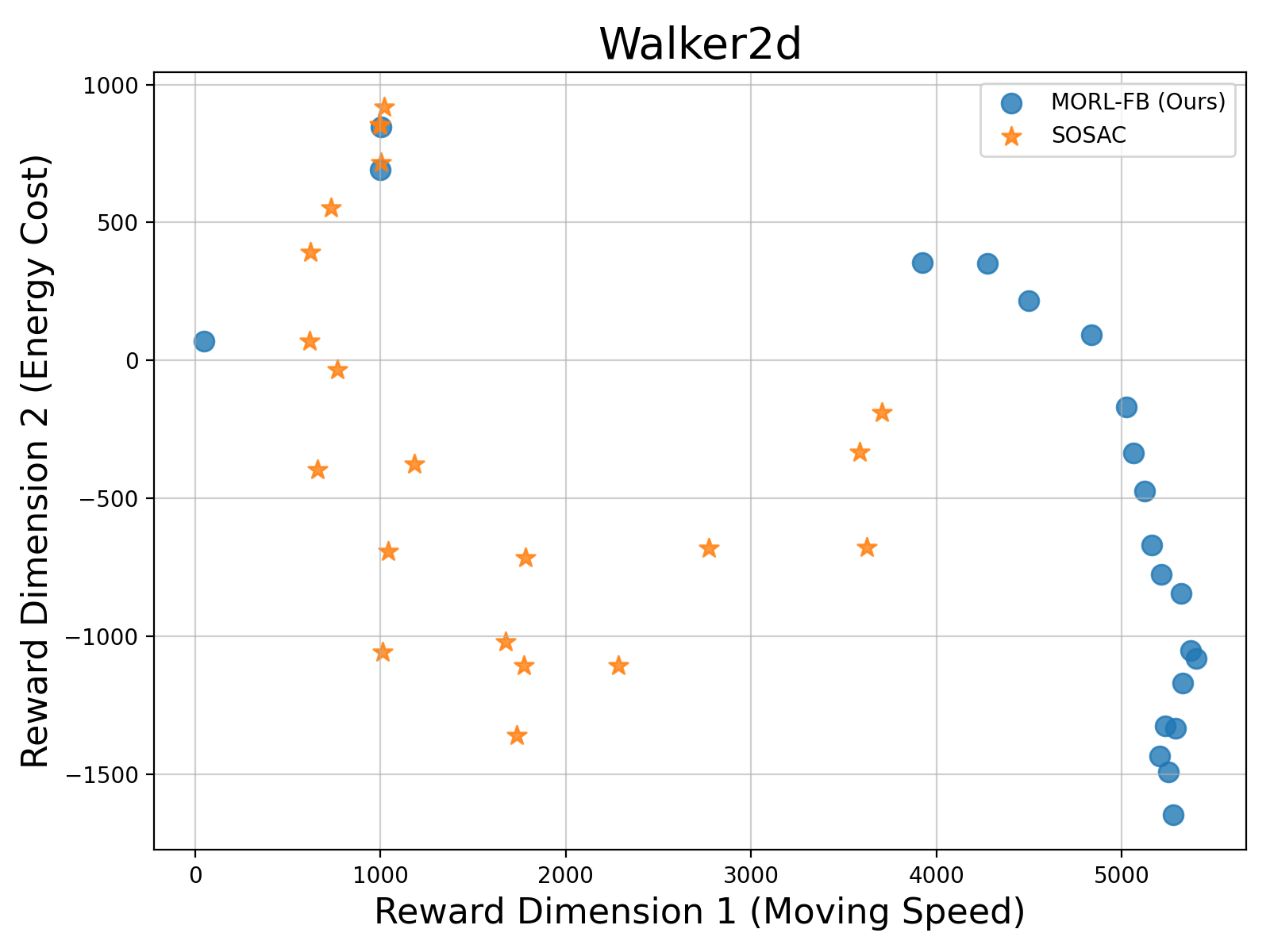}
        \subcaption{Walker2d}
    \end{minipage}
    \caption{Return vectors (Moving Speed vs.\ Energy Cost) achieved under 21 different preference vectors $[0,1], [0.05,0.95], \ldots, [0.95,0.05], [1,0]$ across different methods. Each scatter cloud corresponds to the learned policies under a specific preference, illustrating how MORL methods adapt to diverse trade-offs between objectives.}
    \label{fig:pareto_halfcheetah_walker}
\end{figure}

\rev{\subsection{Comparison of Pareto Fronts}
To evaluate the sample efficiency of MORL-FB, we conduct experiments on 2-objective MuJoCo tasks with a whole range of 21 preference vectors ([0.0, 1.0], [0.05, 0.95], [0.1, 0.9], ···, [1.0, 0.0]). As a comparison baseline, we consider SORL which trains a separate policy for each individual preference. Each single-object SAC (SOSAC) model requires 3M steps, resulting in a total training budget of 63M steps for all 21 preferences. By contrast, MORL-FB only uses 3M steps in total to learn policies that generalize across the entire preference set. \Cref{fig:pareto_halfcheetah_walker} shows the return vectors attained by MORL-FB and the collection of 21 SOSAC models on the Walker2d task. MORL-FB achieves comparable or even superior return vectors with only 1/21 of the samples, demonstrating its strong sample efficiency and generalization ability across diverse preference vectors.}

\subsection{Robustness under Worst-Case Preferences}
\label{sec:wcp}
\rev{
Beyond average performance metrics, it is crucial to assess the robustness of multi-objective reinforcement learning (MORL) algorithms under unfavorable preference settings. In practice, policies deployed in dynamic environments may encounter user preferences that significantly diverge from those seen during training. To capture this aspect, we evaluate each method under \textit{worst-case preferences} using the Conditional Value-at-Risk (CVaR). Following the risk-sensitive reinforcement learning formulation of \citet{tamar2015optimizing}, we compute CVaR@0.1 as the mean of the lowest $10\%$ scalarized returns. Specifically, for each algorithm we uniformly sample $500$ preference vectors, apply linear scalarization to obtain utility values, sort the results, and average the worst $10\%$. This quantifies the expected return conditioned on being in the lowest $\alpha$-quantile:
\begin{align}
\phi(\theta) = \mathbb{E}^\theta[R \mid R \leq \nu_\alpha(\theta)],
\end{align}
\Cref{tab:cvar} reports CVaR@0.1 across HalfCheetah2d and Walker2d. MORL-FB achieves the highest CVaR in both environments, significantly outperforming prior baselines. These results indicate that MORL-FB not only excels on mean metrics but also demonstrates superior robustness against adverse preferences, avoiding catastrophic failures more effectively than existing methods.}
\begin{table}[h]
\centering
\caption{CVaR@0.1 performance across HalfCheetah2d and Walker2d. Higher is better.}
\label{tab:cvar}
\begin{tabular}{lcc}
\toprule
\textbf{Algorithm} & \textbf{HalfCheetah2d} & \textbf{Walker2d} \\
\midrule
PD-MORL     & 2425.91 & 284.89 \\
Q-Pensieve  & 6275.89 & 1516.20 \\
CAPQL       & 4874.64 & 1297.40 \\
PGMORL      & 2630.02 & 1113.49 \\
MORL/D      & 407.66  & 40.47 \\
PCN         & -0.11   & 98.28 \\
SFOLS       & 1390.73 & 402.19 \\
GPI-LS      & 3884.67 & 1546.96 \\
GPI-PD      & 3999.36 & 276.74 \\
MORL-FB (Ours) & \textbf{7123.87} & \textbf{2304.02} \\
\bottomrule
\end{tabular}
\end{table}

\rev{\subsection{Computational Cost Analysis}}
\rev{To provide a fair and comprehensive comparison of computational efficiency, we evaluated all methods under the same hardware environment. Each algorithm was trained for 100K environment steps on a workstation equipped with a single NVIDIA RTX 4090 GPU, an Intel Core i7-13700K CPU, and 64 GB of system memory. \Cref{tab:time_comparison} reports the total wall-clock time required by each method, ensuring a fair comparison of time and resource usage across methods. Compared to strong baselines such as PD-MORL, Q-Pensieve, and GPI-LS, FB-MORL maintains a reasonable training time. While some methods require higher computational costs, our approach strikes a good balance between performance and efficiency.}

\begin{table}[ht]
    \centering
    \caption{Wall-clock time comparison (100K steps).}
    \label{tab:time_comparison}
    \begin{tabular}{l c}
        \hline
        \textbf{Algorithm} & \textbf{Wall-Clock Time (seconds)} \\
        \hline
        PD-MORL        & 1166  \\
        CAPQL          & 3369  \\
        MORL/D         & 793   \\
        SFOLS          & 550   \\
        Q-Pensieve     & 12960 \\
        PGMORL         & 4237  \\
        PCN            & 4445  \\
        GPI-LS         & 3611  \\
        GPI-PD         & 5237  \\
        \textbf{MORL-FB (Ours)} & \textbf{1874} \\
        \hline
    \end{tabular}
\end{table}

\subsection{Detailed Experimental Results of \Cref{sec:exp}}

In this part, we will provide the table we use on plotting the bar chart on \Cref{sec:exp}.

The results in \Cref{tab:discrete,tab:continuous_constraint,tab:MORL_ablation_bar,tab:continuous} correspond to the visualizations shown in \Cref{fig:continuous_bar}, \Cref{fig:continuous_constraint_bar}, \Cref{fig:MORL_ablation_bar} and \Cref{fig:discrete_bar}, respectively.

\begin{table*}[h!]
    \centering
    \caption{Performance of MORL-FB and benchmark algorithms in discrete MORL environments.}
    \label{tab:discrete}
    \scalebox{0.8}{\begin{tabular}{@{}lccccccccc@{}}
        \\
        \toprule
        \multirow{2}{*}{\raisebox{0.3ex}}{Environments} & \multirow{2}{*}{\raisebox{0.3ex}}{Metrics} & \multirow{2}{*}{\raisebox{0.3ex}}{PD-MORL} 
        & \multirow{2}{*}{\raisebox{0.3ex}}{Envelope} & \multirow{2}{*}{\raisebox{0.3ex}}{PCN} & \multirow{2}{*}{\raisebox{0.3ex}}{MORL/D}
        & \multirow{2}{*}{\raisebox{0.3ex}}{SFOLS} & \multirow{2}{*}{\raisebox{0.3ex}}{MORL-FB} & Optimal \\
        & & (w/i interpolator) & & & & & (Ours) & \\ \midrule

        \multirow{3}{*}{Deep Sea Treasure} 
        & UT & 0.52 $\pm$ 0.00 & 6.53 $\pm$ 0.00 & 6.12 $\pm$ 0.95 & 3.88 $\pm$ 0.26 & 5.65 $\pm$ 0.95 & 6.43 $\pm$ 0.10 & 6.89 \\
        & HV($\times 10^{2}$) & 9.33 $\pm$ 0.00 & 9.91 $\pm$ 0.00 & 8.71 $\pm$ 1.49 & 5.63 $\pm$ 0.41 & 8.46 $\pm$ 0.01 & 9.92 $\pm$ 0.00 & 9.92 \\
        & ED & 0.69 $\pm$ 0.00 & 0.77 $\pm$ 0.00 & 0.81 $\pm$ 0.01 & 0.15 $\pm$ 0.10 & 0.25 $\pm$ 0.00 & - & - \\ 
        \midrule

        \multirow{3}{*}{Fruit Tree Navigation} 
        & UT & 5.03 $\pm$ 0.00 &  5.07 $\pm$ 0.00 & 5.00 $\pm$ 0.02 & 4.19 $\pm$ 0.03 &  4.77 $\pm$  0.01 & 5.01 $\pm$ 0.01 & 5.08 \\
        & HV($\times 10^{4}$) & 1.25 $\pm$ 0.00 & 1.16 $\pm$ 0.00 & 1.01 $\pm$ 0.06 & 0.18 $\pm$ 0.05 & 0.87 $\pm$ 0.01 & 1.16 $\pm$ 0.01 & 1.25 \\
        & ED & 0.76 $\pm$ 0.01 & 0.95 $\pm$ 0.00 & 0.72 $\pm$ 0.03 & 0.01 $\pm$ 0.01 & 0.08 $\pm$ 0.01 & - & - \\

        \bottomrule
    \end{tabular}}
\end{table*}

\begin{table*}
    \centering
    \caption{Performance of MORL-FB, PD-MORL, and Q-Pensieve under constrained preference training.}
    \label{tab:continuous_constraint}
    \scalebox{0.85}{\begin{tabular}{@{}lccccccccc@{}}
        \\
        \toprule
        Environments & Metrics & PD-MORL & Q-Pensieve & MORL-FB \\
        \midrule

        \multirow{3}{*}{Hopper3d} 
        & UT($\times$ \num{e3}) & 1.05 $\pm$ 0.02 & 1.72 $\pm$ 0.01 & \textbf{2.26 $\pm$ 0.01} \\
        & HV($\times$ \num{e11}) & 0.61 $\pm$ 0.01 & 0.88 $\pm$ 0.01 & \textbf{1.16 $\pm$ 0.01} \\
        & ED & 0.01 $\pm$ 0.00 & 0.11 $\pm$ 0.00 & - \\ \midrule
        
        \multirow{3}{*}{Ant3d} 
        & UT($\times$ \num{e3}) & 1.69 $\pm$ 0.04 & 0.49 $\pm$ 0.02 & \textbf{3.11 $\pm$ 0.24} \\
        & HV($\times$ \num{e11}) & 2.18 $\pm$ 0.03 & 0.51 $\pm$ 0.00 & \textbf{3.17 $\pm$ 0.05} \\
        & ED & 0.22 $\pm$ 0.02 & 0.26 $\pm$ 0.04 & - \\ \midrule

        \multirow{3}{*}{Humanoid2d} 
        & UT($\times$ \num{e2}) & -0.04 $\pm$ 0.00\phantom{-} & 4.51 $\pm$ 0.38 & \textbf{8.19 $\pm$ 0.03} \\
        & HV($\times$ \num{e7}) & 0.06 $\pm$ 0.00 & 1.51 $\pm$ 0.04 & \textbf{1.83 $\pm$ 0.01} \\
        & ED & 0.00 $\pm$ 0.00 & 0.40 $\pm$ 0.05 & - \\
        
        \bottomrule
    \end{tabular}}
    \label{tab:cconstrained preference}
\end{table*}

\begin{table*}
    \centering
    \caption{\textbf{Performance comparison of MORL-FB and its ablated versions across environments.} This table evaluates MORL-FB against variants lacking preference-guided exploration or the Q-loss component, showing their performance across different environments.}
    \label{tab:MORL_ablation_bar}
    \begin{tabular}{@{}lccccccccc@{}}
        \\
        \toprule
        \multirow{2}{*}{\raisebox{0.3ex}}{Environments} & \multirow{2}{*}{\raisebox{0.3ex}}{Metrics} & \multirow{2}{*}{\raisebox{0.3ex}}{MORL-FB} 
        & \multirow{2}{*}{\raisebox{0.3ex}}{MORL-FB} & \multirow{2}{*}{\raisebox{0.3ex}}{MORL-FB} \\
        & & (w/o preference-guided exploration) & (w/o q loss) &   \\ \midrule

        \multirow{2}{*}{Hopper3d} 
        & UT($\times$ \num{e3})  & 2.03 $\pm$ 0.42 & 0.33 $\pm$ 0.07 & \textbf{2.36 $\pm$ 0.00}\\
        & HV($\times$ \num{e11}) & 0.91 $\pm$ 0.02 & 0.44 $\pm$ 0.02 & \textbf{1.15 $\pm$ 0.01}\\ 
        \midrule
        
        \multirow{2}{*}{Ant3d} 
        & UT($\times$ \num{e3}) & 1.23 $\pm$ 0.03 & -1.01 $\pm$ 0.01\phantom{-} & \textbf{3.43 $\pm$ 0.22}  \\
        & HV($\times$ \num{e11}) & 1.47 $\pm$ 0.03 & 0.06 $\pm$ 0.00 & \textbf{4.18 $\pm$ 0.04} \\ 
        \midrule

        \multirow{2}{*}{Humanoid2d} 
        & UT($\times$ \num{e2}) & 2.38 $\pm$ 0.30 & -1.43 $\pm$ 0.14\phantom{-} & \textbf{8.13 $\pm$ 0.01} \\
        & HV($\times$ \num{e7}) & \textbf{1.78 $\pm$ 0.00} & 0.96 $\pm$ 0.00 & 1.75 $\pm$ 0.02 \\
        
        \bottomrule
    \end{tabular}
    \label{tab:inference}
\end{table*}

\begin{landscape}
\begin{table}
    \centering
    \caption{Comparative performance of MORL-FB and various benchmark algorithms across continuous control tasks in MuJoCo.}
    \label{tab:continuous}
    % \begin{adjustbox}{angle=90}
    \scalebox{0.8}{\begin{tabular}{@{}lcccccccccccc@{}}
        \\
        \toprule
        \multirow{2}{*}{\raisebox{0.3ex}}{Environments} & \multirow{2}{*}{\raisebox{0.3ex}}{Metrics} & \multirow{2}{*}{\raisebox{0.3ex}}{PD-MORL} 
        & \multirow{2}{*}{\raisebox{0.3ex}}{Q-Pensieve} & \multirow{2}{*}{\raisebox{0.3ex}}{CAPQL} & \multirow{2}{*}{\raisebox{0.3ex}}{PGMORL}
        & \multirow{2}{*}{\raisebox{0.3ex}}{MORL/D} & \multirow{2}{*}{\raisebox{0.3ex}}{PCN} & \multirow{2}{*}{\raisebox{0.3ex}}{SFOLS} & \multirow{2}{*}{\raisebox{0.3ex}}{GPI-LS} & \multirow{2}{*}{\raisebox{0.3ex}}{GPI-PD} &  \multirow{2}{*}{\raisebox{0.3ex}}{FB} & \multirow{2}{*}{\raisebox{0.3ex}}{MORL-FB} \\
        & & (w/o interpolator) & & & & & & & & & & (Ours)\\ \midrule
        
        \multirow{3}{*}{Halfcheetah2d}
        & UT($\times 10^{3}$) & 3.17 $\pm$ 0.04 & 6.85 $\pm$ 0.01 & 5.39 $\pm$ 0.01 & 2.72 $\pm$ 0.03 & 0.41 $\pm$ 0.01 & 0.00 $\pm$ 0.00 & 1.39 $\pm$ 0.03 & 4.15 $\pm$ 0.84& 4.06 $\pm$ 0.03 & 1.26 $\pm$ 0.00 & \textbf{7.69 $\pm$ 0.08} \\
        & HV($\times 10^{8}$) & 0.95 $\pm$ 0.00 & 1.13 $\pm$ 0.00 & 0.89 $\pm$ 0.00 & 0.73 $\pm$ 0.01 & 0.08 $\pm$ 0.00 & 0.00 $\pm$ 0.00 & 0.59 $\pm$ 0.03 & 0.67 $\pm$ 0.12& 0.68 $\pm$ 0.00 & 0.26 $\pm$ 0.00 & \textbf{1.24 $\pm$ 0.00} \\
        & ED & 0.06 $\pm$ 0.01 & 0.08 $\pm$ 0.01 & 0.06 $\pm$ 0.01 & 0.03 $\pm$ 0.01 & 0.05 $\pm$ 0.01 & 0.05 $\pm$ 0.01 & 0.01 $\pm$ 0.01 & 0.05 $\pm$ 0.02& 0.06 $\pm$ 0.01 & 0.04 $\pm$ 0.02 & - \\ \midrule

        \multirow{3}{*}{Walker2d} 
        & UT($\times 10^{3}$) & 1.70 $\pm$ 0.01 & 1.66 $\pm$ 0.08 & 1.49 $\pm$ 0.01 & 1.15 $\pm$ 0.02 & 0.05 $\pm$ 0.00 & 0.10 $\pm$ 0.00 & 0.41 $\pm$ 0.01 & \textbf{1.92 $\pm$ 0.18}& 0.54 $\pm$ 0.00 & 0.29 $\pm$ 0.01 & 1.87 $\pm$ 0.05 \\
        & HV($\times 10^{7}$) & 4.02 $\pm$ 0.01 & 4.54 $\pm$ 0.02 & 0.03 $\pm$ 0.01 & 3.52 $\pm$ 0.09 & 0.35 $\pm$ 0.00 & 0.17 $\pm$ 0.00 & 1.22 $\pm$ 0.00 & 3.35 $\pm$ 0.33& 1.13 $\pm$ 0.00 & 1.97 $\pm$ 0.12 & \textbf{4.56 $\pm$ 0.00}  \\
        & ED & 0.27 $\pm$ 0.03 & 0.36 $\pm$ 0.01 & 0.00 $\pm$ 0.00 & 0.18 $\pm$ 0.01 & 0.03 $\pm$ 0.01 & 0.03 $\pm$ 0.01 & 0.15 $\pm$ 0.02 & 0.32 $\pm$ 0.01& 0.15 $\pm$ 0.02 & 0.00 $\pm$ 0.00 & - \\ \midrule

        \multirow{3}{*}{Hopper3d}
        & UT($\times 10^{3}$) & 1.29 $\pm$ 0.02 & 1.82 $\pm$ 0.02 & 1.25 $\pm$ 0.01 & 0.78 $\pm$ 0.00 & 0.11 $\pm$ 0.00 & 0.01 $\pm$ 0.01 & 0.03 $\pm$ 0.01 & 0.57 $\pm$ 0.23& 1.32 $\pm$ 0.01 & 0.01 $\pm$ 0.00 & \textbf{2.36 $\pm$ 0.01} \\
        & HV($\times 10^{11}$) & 0.92 $\pm$ 0.00 & \textbf{1.25 $\pm$ 0.01} & 0.57 $\pm$ 0.01 & 0.01 $\pm$ 0.00 & 0.01 $\pm$ 0.00 & 0.00 $\pm$ 0.00 & 0.03 $\pm$ 0.00 & 0.29 $\pm$ 0.00& 0.51 $\pm$ 0.00 & 0.00 $\pm$ 0.00 & 1.15 $\pm$ 0.00 \\
        & ED & 0.05 $\pm$ 0.00 & 0.35 $\pm$ 0.01 & 0.00 $\pm$ 0.00 & 0.03 $\pm$ 0.01 & 0.00 $\pm$ 0.00 & 0.00 $\pm$ 0.00 & 0.00 $\pm$ 0.00 & 0.02 $\pm$ 0.10 & 0.10 $\pm$ 0.05 & 0.00 $\pm$ 0.00 & - \\ \midrule

        \multirow{3}{*}{Ant3d} 
        & UT($\times 10^{3}$) & 1.29 $\pm$ 0.11 & 1.81 $\pm$ 0.08 & 1.01 $\pm$ 0.00 & 0.80 $\pm$ 0.00 & 0.04 $\pm$ 0.00 & 0.05 $\pm$ 0.01 & 0.28 $\pm$ 0.01 & 1.45 $\pm$ 0.10& 0.66 $\pm$ 0.00 & 0.01 $\pm$ 0.01 & \textbf{3.43 $\pm$ 0.22} \\
        & HV($\times 10^{11}$) & 2.42 $\pm$ 0.05 & 2.61 $\pm$ 0.02 & 0.45 $\pm$ 0.00 & 0.21 $\pm$ 0.03 & 0.01 $\pm$ 0.00 & 0.07 $\pm$ 0.01 & 0.12 $\pm$ 0.00 & 0.73 $\pm$ 0.15 & 0.35 $\pm$ 0.01 & 0.08 $\pm$ 0.00 & \textbf{4.18 $\pm$ 0.04} \\
        & ED & 0.11 $\pm$ 0.05 & 0.14 $\pm$ 0.04 & 0.08 $\pm$ 0.05 & 0.08 $\pm$ 0.05 & 0.05 $\pm$ 0.03 & 0.06 $\pm$ 0.03 & 0.06 $\pm$ 0.04 & 0.17 $\pm$ 0.02 & 0.08 $\pm$ 0.05 & 0.03 $\pm$ 0.02 & - \\ \midrule

        \multirow{3}{*}{Humanoid2d}
        & UT($\times 10^{2}$) & -0.05 $\pm$ 0.00\phantom{-} & 4.51 $\pm$ 0.03 & -0.04 $\pm$ 0.00\phantom{-} & 1.55 $\pm$ 0.04 & 2.71 $\pm$ 0.02 & -3.79 $\pm$ 0.72\phantom{-} & 0.45 $\pm$ 0.03 & 1.43 $\pm$ 0.12& 1.35 $\pm$ 0.00 & 2.44 $\pm$ 0.00 & \textbf{8.13 $\pm$ 0.01} \\
        & HV($\times 10^{7}$) & 0.06 $\pm$ 0.00 & 1.51 $\pm$ 0.04 & 0.01 $\pm$ 0.01 & 1.01 $\pm$ 0.02 & 0.62 $\pm$ 0.00 & 0.25 $\pm$ 0.01 & 0.75 $\pm$ 0.05 & 0.77 $\pm$ 0.21& 0.42 $\pm$ 0.03 & 1.19 $\pm$ 0.05 & \textbf{1.75 $\pm$ 0.02}\\
        & ED & 0.07 $\pm$ 0.03 & 0.40 $\pm$ 0.04 & 0.08 $\pm$ 0.00 & 0.07 $\pm$ 0.01 & 0.18 $\pm$ 0.01 & 0.00 $\pm$ 0.00 & 0.04 $\pm$ 0.00 & 0.15 $\pm$ 0.01& 0.14 $\pm$ 0.00 & 0.06 $\pm$ 0.00 & - \\ \midrule

        \multirow{3}{*}{Humanoid5d}
        & UT($\times 10^{3}$) & 0.38 $\pm$ 0.01 & 0.12 $\pm$ 0.00 & 0.08 $\pm$ 0.00 & 0.54 $\pm$ 0.00 & 0.58 $\pm$ 0.00 & -0.10 $\pm$ 0.01\phantom{-} & 0.54 $\pm$ 0.00 & 0.72 $\pm$ 0.02& 0.30 $\pm$ 0.00 & 0.55 $\pm$ 0.03 & \textbf{1.11 $\pm$ 0.00} \\
        & HV($\times 10^{16}$) & 3.91 $\pm$ 0.04 & 1.80 $\pm$ 0.02 & 0.00 $\pm$ 0.00 & 0.59 $\pm$ 0.00 & 0.31 $\pm$ 0.00 & 0.00 $\pm$ 0.00 & 2.15 $\pm$ 0.08 & 3.48 $\pm$ 0.03& 0.20 $\pm$ 0.01 & 4.69 $\pm$ 0.20 & \textbf{6.99 $\pm$ 0.06}\\
        & ED & 0.03 $\pm$ 0.00 & 0.00 $\pm$ 0.00 & 0.04 $\pm$ 0.00 & 0.10 $\pm$ 0.00 & 0.08 $\pm$ 0.00 & 0.00 $\pm$ 0.00 & 0.05 $\pm$ 0.00 & 0.09 $\pm$ 0.00& 0.03 $\pm$ 0.00 & 0.03 $\pm$ 0.00 & - \\

        \bottomrule
    \end{tabular}}
    % \end{adjustbox}
\end{table}
\end{landscape}

\begin{figure}[H]
    \centering
    \includegraphics[width=0.7\linewidth]{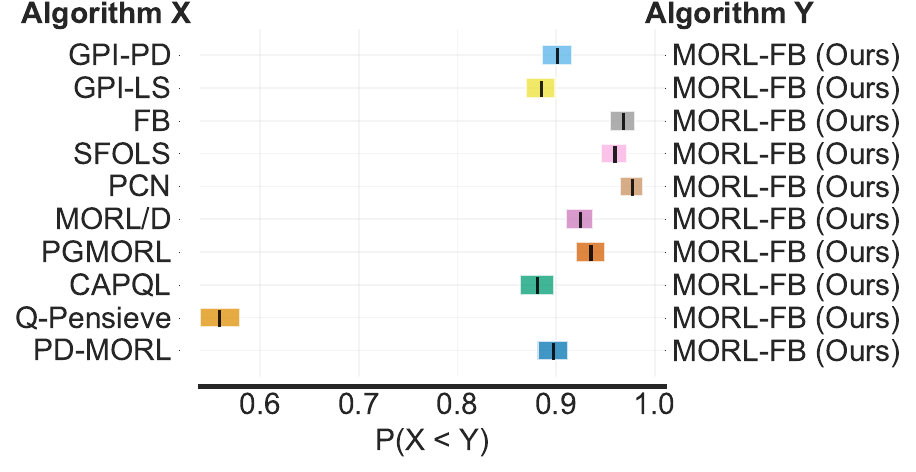}
    \caption{\textbf{Probability of Improvement (POI) of MORL-FB against benchmark algorithms.} This figure illustrates the POI of MORL-FB relative to various benchmark MORL algorithms.}
    \label{fig:Continuous_POI}
\end{figure}

We also evaluated the probability of improvement (POI) between MORL-FB and the benchmark algorithms suggested by \citet{agarwal2021deep} in \Cref{fig:Continuous_POI}.  POI quantifies the likelihood that one algorithm will outperform another.  The results demonstrate that MORL-FB consistently achieves superior performance compared to baselines.

\begin{figure}[H]
    \centering
    \includegraphics[width=1.0\linewidth]{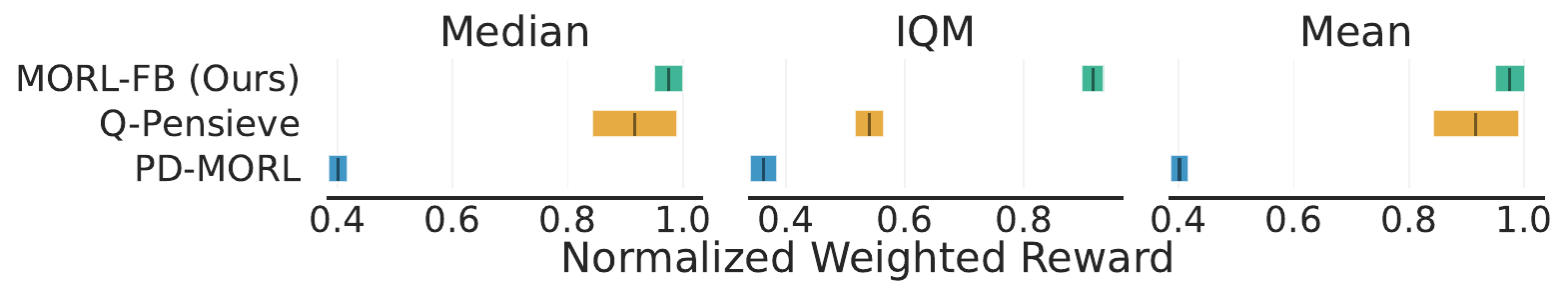}
    \caption{\textbf{Medium, IQM, and Mean performance of MORL-FB and benchmarks trained with small preference sets.} This figure displays the Median, Interquartile Mean (IQM), and Mean performance for MORL-FB and other benchmark algorithms, trained with only a small set of preference vectors.}
    \label{fig:constraint_IQM}
\end{figure}

\begin{figure}[H]
    \centering
    \includegraphics[width=0.7\linewidth]{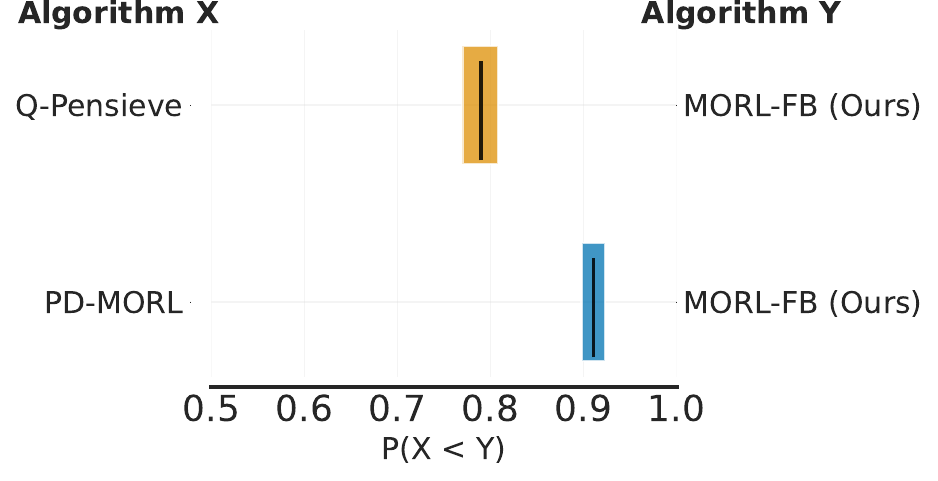}
    \caption{\textbf{Probability of Improvement (POI) of MORL-FB under constrained preference training.} This figure illustrates the POI of MORL-FB relative to other benchmark algorithms, all trained with specific constraint sets of preference vectors.}
    \label{fig:constraint_POI}
\end{figure}

Additionally, we assessed the performance of training with constraint preferences using the method from \citet{agarwal2021deep}, presented in \Cref{fig:constraint_IQM,fig:constraint_POI}.

\end{document}